\documentclass{article}

\usepackage[eandd, final]{neurips_2026}

\usepackage[utf8]{inputenc} %
\usepackage[T1]{fontenc}    %
\usepackage[hidelinks]{hyperref}       %
\usepackage{url}            %
\usepackage{booktabs}       %
\usepackage{amsfonts}       %
\usepackage{nicefrac}       %
\usepackage{microtype}      %
\usepackage{xcolor}         %
\usepackage{natbib}
\usepackage{pifont}
\usepackage{amsmath}
\usepackage{graphicx}
\usepackage{wrapfig}
\usepackage{bm}

\title{Can an LLM Reason Like a Lawyer? \\
\large{Benchmarking the ability of LLMs to map the facts of a case to the elements of the applicable legal rule}}

\author{%
  Shounak Paul \\
  Max Planck Institute for Software Systems\\
  67663 Kaiserslautern \\
  \texttt{shpaul@mpi-sws.org} \\
  \And
  Seungeon Lee\thanks{Corresponding Author} \\
  Max Planck Institute for Software Systems \\
  67663 Kaiserslautern \\
  \texttt{selee@mpi-sws.org} \\
  \And
  Christoph Engel \\
  Max Planck Institute for Behavioral Economics \\
  D 53113 Bonn \\
  \texttt{engel@econ.mpg.de} \\
  \And
  Krishna P. Gummadi \\
  Max Planck Institute for Software Systems \\
  67663 Kaiserslautern \\
  \texttt{gummadi@mpi-sws.org} \\
}

\begin{document}

\maketitle

\begin{abstract}
Legal judgment prediction has made substantial progress in recent years, yet existing work evaluates whether models reach correct conclusions rather than whether they reason correctly. The core of legal decision-making is the mapping of case facts to the elements of an applicable legal provision, a step that prior benchmarks neglect. We introduce \texttt{LegalReasoningBench}, the first benchmark that tests legal syllogism from the perspective of mapping facts of a case to the elements of doctrine. Our benchmark is built on 375 judgments of the European Court of Human Rights (ECtHR) discussing Art. 11 ECHR on freedom of assembly. We exploit that the court routinely splits the decision in a stage reporting the facts of the case, and an evaluation stage in which it uses these facts selectively for justification. We give the (preprocessed) list of facts to 10 downstream LLMs and evaluate their justifications. The pipeline is validated by three expert legal annotators. Downstream models are reasonably accurate in predicting whether Art. 11 has been violated, and which elements of doctrine are critical, but perform substantially worse at attributing specific facts to doctrinal elements. Essentially they make the right decision, but for the wrong reasons. This weakness is invisible under standard evaluation metrics.
Our pipeline can easily be adapted to any legal rule that follows the Civil law tradition, i.e. to legal reasoning that requires the mapping of facts to an abstract legal rule.
\end{abstract}

\section{Introduction} \label{sec:introduction}
Court judgments, particularly in Continental jurisdictions (like Germany, France or Italy), routinely read like quasi-mathematical exercises.
In these countries, \emph{abstract legal rules} have to be applied to the \emph{facts of the case}.
Take the definition of theft in \S\,242 German Criminal Code:
\begin{quote}
    ``Whoever takes movable property belonging to another away from another with the intent of unlawfully appropriating it for themselves or a third party incurs a penalty of imprisonment for a term not exceeding five years or a fine.''
\end{quote}
A judge must decide whether these conditions are fulfilled. This requires that (i)~there is movable property belonging to another person, (ii)~the defendant has taken this property away, and (iii)~the defendant had the intention of appropriating the property. To that end, the judge must \emph{map the facts of a case to the elements (conditions) of the applicable legal rules}.
This mapping is central to \textit{legal syllogism}, the core deductive structure required to justify a legal decision both in Common Law~\citep{maccormick1994legal,schauer2009thinking} and Civil Law~\citep{alexy1989theory,mollers2020legal} jurisdictions.
\begin{wrapfigure}{r}{0.42\textwidth}
    \centering
    \includegraphics[width=0.40\textwidth]{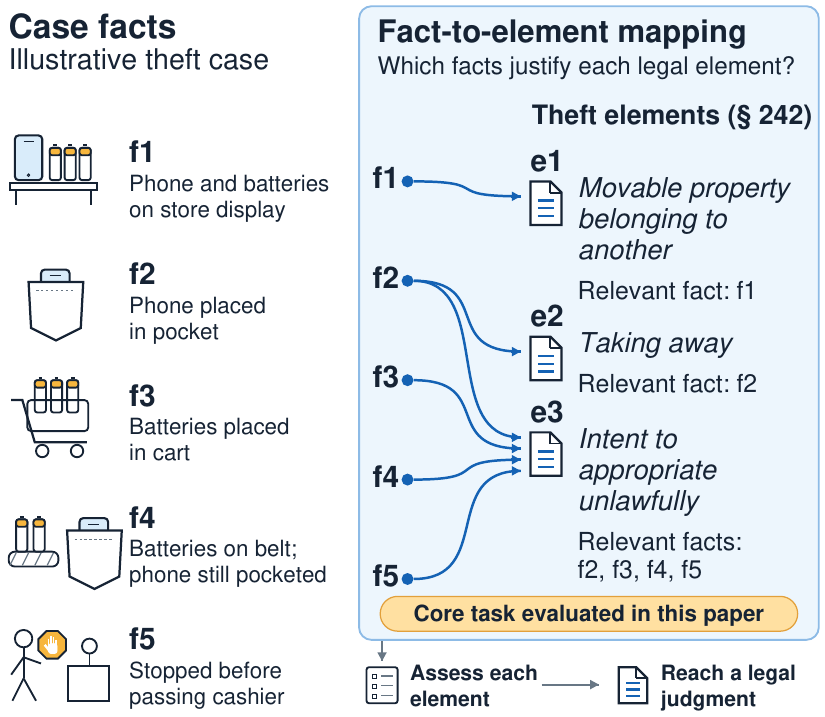}
    \caption{Illustration of fact-law mapping}
    \label{fig:illustration}
\end{wrapfigure}

Consider the theft case illustrated in Figure~1: in a department store, a customer C puts an iPhone on display in his pocket, but places cheap batteries in his shopping cart and then on the cashier’s conveyor belt. A shop detective intercepts C before he passes the cashier.
The first two conditions (law elements) are clearly met: the iPhone constitutes a movable property belonging to the other, and C putting the iPhone in his pocket constitutes the act of taking away. The tricky element is the third. Does putting other cheap items in the shopping cart and on the conveyor belt provide a sufficient signal that C had the intention to steal? Is it sufficiently likely that C had intended to empty his pockets before being intercepted by the shop detective?

In such a case, a trained judge tries out alternative interpretations of the law, and alternative readings of the facts, in an iterative exercise until they match (or no match can be established), distilling legally relevant facts from much richer circumstantial evidence.

First generation legal informatics has bracketed this complication with legal reasoning.
They predominantly focused on training models to predict legal outcomes --- whether a case is won or lost, which charge applies, or what sentence is appropriate~\citep{aletras2016predicting, chalkidis2019neural, zhong2018topjudge}.
However, with the advances in their reasoning capabilities, Large Language Models (LLMs) may enable disruptive innovations in legal reasoning.
Some recent work has begun evaluating step-by-step legal reasoning structured around legal syllogism~\citep{jiang2023legal, deng2023syllogistic, guha2023legalbench}, and a small number of papers have evaluated frontier LLMs specifically on cases from the European Court of Human Rights (ECtHR)~\citep{chlapanis2024lar, engel2024professor}.
Yet this literature does not address what arguably is the core of legal decision-making: identifying which subsets of case facts are relevant to the different elements of the applicable legal provision. Investigating LLMs ability to \emph{map atomic facts to law elements} is the focus of this paper.

To this end, {\bf we build \texttt{LegalReasoningBench}, a benchmark for the legal reasoning capability of large language models}. We argue that the European Convention on Human Rights (ECHR) is well suited for the task. First, the ECtHR has given the convention rights a clear, canonical structure, which enables us to test whether a downstream LLM evaluates each element of law the same way as the Court. Second, the jurisprudence of the Court is sufficiently rich so that we can generalize beyond idiosyncrasies of individual cases.
Third, this richness makes it possible to add a whole level of granularity in our benchmark. We can test whether a downstream LLM uses the same set of facts as ECtHR to substantiate its conclusion, separately for each element of doctrine.
As a proof-of-concept, we focus on \textit{freedom of assembly}, as protected by Article 11 ECHR.

{\bf Key challenge.} When constructing our benchmark, we faced
a significant complication.
While the primary purpose of the text is to show that the decision has been taken on sound doctrinal grounds,
the court also typically tries to convince the parties that it has taken their concerns seriously.
Often the court also speaks to a wider public. It for instance expects that the ruling would be read more widely, and would clarify the contested interpretation of a doctrinal element. The court may also speak to a political audience, hoping to stabilize a normatively desirable attitude in the population. For all these reasons, legal reasoning is often richer than would be absolutely necessary to justify the doctrinal conditions, which poses a challenge when building the benchmark.
We develop a methodology that distinguishes between mapping those facts that are indispensable for reaching the doctrinal conclusion, and the ones that are only contextual, and likely reported in the original ruling as the court intends to also serve additional audiences.

{\bf Critical design choices.}
To construct the benchmark, we develop an LLM-based pipeline and apply it to all published 375 cases in which the ECtHR discusses \textit{freedom of assembly}, and that fulfill the inclusion criteria defined in Section~\ref{sec:benchmark/sample}. We exploit the fact that the Court routinely organizes its rulings into two parts: a (\textit{fact) reporting} part and a \textit{(law) evaluation} part.
We enable the evaluation of mapping both parts by isolating ``atomic facts'' in the first part, and measuring which facts the court uses in the law part, for substantiating which element of law.
Atomic fact decomposition is our operationalization for computational tractability; we do \textit{not} assume that lawyers explicitly decompose cases into such units in practice.

Ultimately, we want to learn whether an LLM "reasons like a lawyer".
So, when creating the benchmark, we start from the evaluation part of the Court's ruling, and identify the atomic facts that the Court uses to justify its conclusion for each element of doctrine.
Three legal experts validated the resulting annotations, with \textit{all} of them finding at least $90\%$ annotations to be acceptable.
Then we give the downstream LLM the complete set of atomic facts and ask it to decide the case.
We evaluate three tasks: (i)~whether the LLM reaches the same conclusion as the Court on an Article 11 violation; (ii)~whether it reaches the same conclusion on each element of doctrine; and (iii)~whether it grounds these conclusions in the same facts as the Court.
Reconstructing the benchmark with a different LLM yields largely stable downstream model rankings, suggesting that the relative performance of the evaluated models is not specific to the LLM used for benchmark construction.
A preliminary application to $25$ Article 10 (\textit{freedom of expression}) cases further demonstrates that the pipeline can be extended to another legal provision with minimal changes.
\space We publicly release the benchmark\footnote{https://huggingface.co/datasets/archon159/LegalReasoningBench} and the code\footnote{https://github.com/archon159/LegalReasoningBench} for both the construction pipeline and the downstream evaluation.

{\bf Can LLMs reason like a lawyer?} We used our benchmark to evaluate legal reasoning abilities of 10 LLMs (three closed frontier models namely, {\tt Claude Sonnet 4.6}, {\tt GPT-5.4}, and {\tt Gemini 3 Flash}, and seven open models belonging to the {\tt GPT-OSS}, {\tt Qwen 3.5}, and {\tt Gemma 4} families).
Frontier models reach the same final conclusion as the ECtHR in $93.9-94.9\%$ cases. Yet agreement on the outcome, or even on individual elements of doctrine, does \textit{not} establish that a model can identify the facts supporting those conclusions.
When we ask frontier models to map the supporting facts, they achieve high scores on \textit{fewer} cases compared to element conclusions.
This difficulty becomes more pronounced when factual support must be attributed to individual elements of doctrine: Nine of ten models perform worse at mapping essential facts to particular elements than at identifying the decisive facts for the case overall.
We conclude that, {\bf while LLMs are helpful for predicting the outcome of a court case, as yet, their ability to reason like a lawyer is still quite limited}.

\paragraph{Related work.} In App.~\ref{app:relwork}, we provide an extended discussion of related work. At this point, we only succinctly define our intended contribution to the literature. First generation papers have focused on predicting court decisions \citep[e.g.][]{aletras2016predicting, chalkidis2019neural}. Later papers have shown that LLMs can be guided through the logical steps of doctrine \citep[e.g.][]{deng2023syllogistic} and have tested simple versions of the syllogism, i.e. mapping facts to elements of doctrine \citep[e.g.][]{guha2023legalbench}. But no paper has addressed the hermeneutical core of legal reasoning, i.e. the constructive exercise of selecting facts, selecting legal elements, and mapping them. This is our contribution. A critical input for our approach is the decomposition of the reporting section in a court ruling into atomic facts. For this, we build on the factual verification literature \citep[e.g.][]{min2023factscore}. We follow the tradition in the field that works with the case law of the ECtHR  \citep[e.g.][]{engel2024professor}.

\section{Benchmark Creation} \label{sec:benchmark}
We design a structured, LLM-based pipeline for comparing the ability of downstream LLMs to map facts to elements of legal doctrine, using actual court rulings. As a proof of concept, we apply this logic to one human right protected by the European Convention on Human Rights (ECHR). But the approach is meant to be generalizable to other human rights protected by the convention, and essentially to any legal rule that follows the tradition of the Civil Law jurisdictions, and reasons from abstract normative rules to the facts of a specific case.
Consequently, with minor adjustments, the same pipeline could generate a whole family of benchmarks for LLM legal reasoning.

\subsection{Sample Selection}
\label{sec:benchmark/sample}

We collected 1,161 case judgments from \url{hudoc.echr.coe.int} tagged as involving Art.\ 11 ECHR.
We apply four inclusion criteria: (i)~only cases concerning freedom of assembly rather than association, which has a significantly distinct doctrinal structure;
(ii)~cases where the Court reaches uniform conclusions across all applicants and complaints; (iii)~cases where the facts are undisputed between the parties; and
(iv) cases where the Court performs a genuine merits analysis of the Art.\ 11 elements rather than disposing on admissibility or procedural grounds.
After applying these criteria, 385 cases remain.
\footnote{The final benchmark consists of 375 cases since 10 cases got dropped due to timeout errors in the pipeline (see \S~\ref{sec:benchmark/architecture})}
Case counts per criterion are reported in Appendix~\ref{app:benchmark/sample}, prompt in App.~\ref{app:prompts/sample_sel}.

\subsection{Architecture of Pipeline} \label{sec:benchmark/architecture}

The central challenge in constructing this benchmark is to map atomic facts from the reporting section of a judgment to the elements of law in its evaluation section, in a way that is faithful to the Court’s own reasoning.
Our pipeline therefore does not read the evaluation section from the beginning and extract every fact it mentions.
Instead, it starts from the Court’s conclusion on each element and works backward to the facts the Court invoked to justify it.
This backward design ensures that the extracted facts reflect the Court’s stated reasoning.

\begin{figure}
\centering
\includegraphics[width=0.95\textwidth]{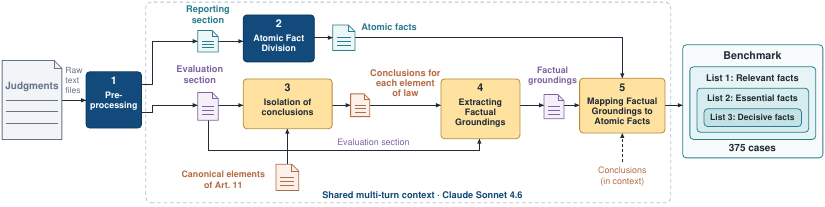}
\caption{Pipeline of Benchmark Creation}
\label{fig:pipeline}
\end{figure}

We organize the pipeline into five stages (Figure~\ref{fig:pipeline}):
(i)~pre-processing, (ii)~definition of atomic facts in the reporting section, (iii)~isolation of conclusions regarding each canonical element of the law, (iv)~extraction of facts invoked by the court in the evaluation section to justify its conclusion regarding each element of doctrine, and (v)~mapping of atomic facts (reporting section) to facts invoked for justification (evaluation section). We use \texttt{Claude Sonnet 4.6} in its default settings for all stages of the pipeline.

\textbf{Pre-processing.}
We use the LLM to remove formatting artifacts in the raw text files such as headers, footers and page numbers and return the reporting and evaluation sections \textit{separately} in a structured format, maintaining paragraph numbers and section headers (prompt in App.~\ref{app:prompts/preproc}).

\textbf{Atomic Fact Division.}
We prompt the LLM to divide the reporting section into atomic facts, and to resolve coreferences so that each atomic fact is intelligible on its own (prompt in App.~\ref{app:prompts/atomic_facts}).

\begin{wrapfigure}{r}{0.37\textwidth}
\centering
\includegraphics[width=0.35\textwidth]{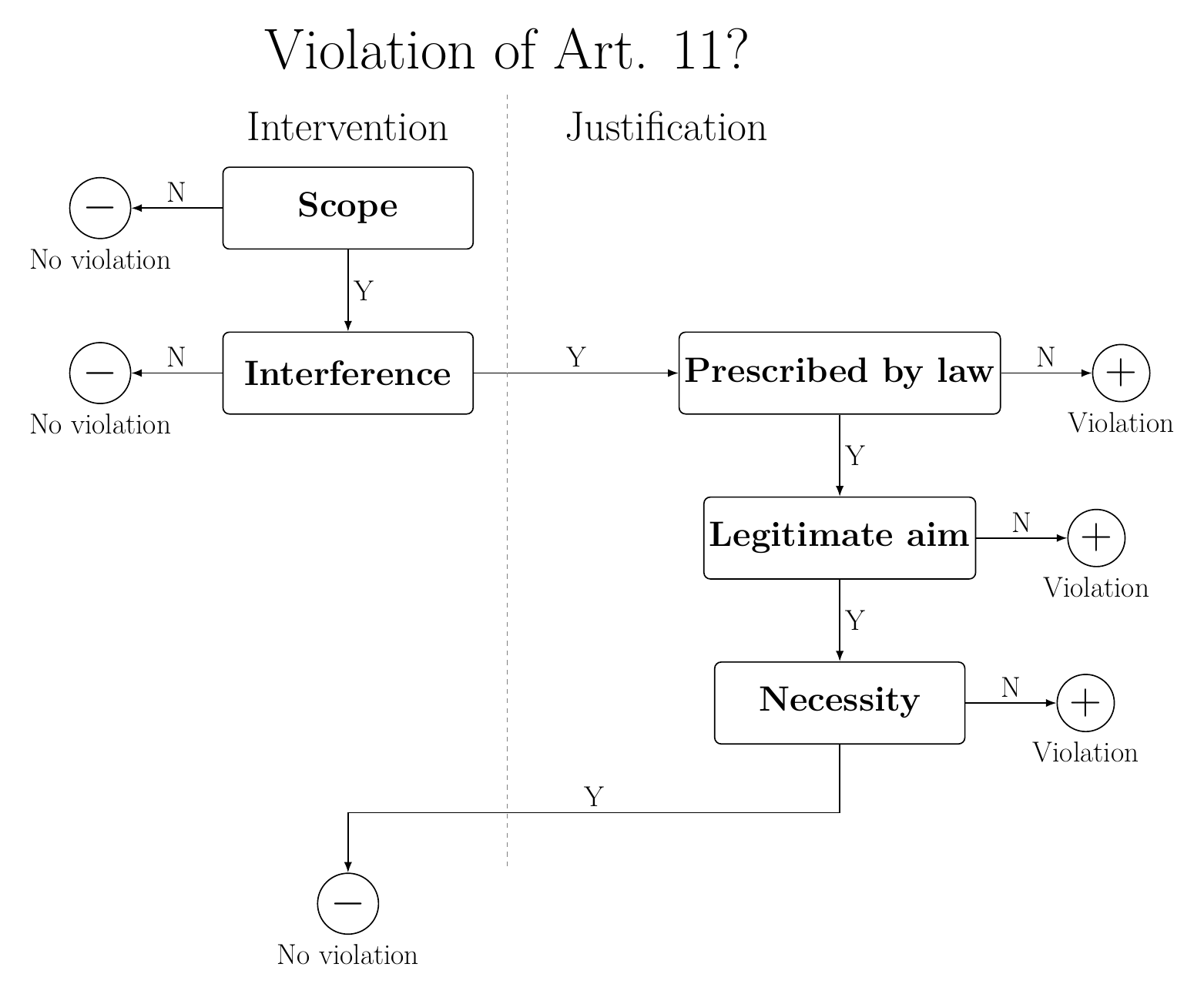}
\caption{Logic of Art.\ 11 ECHR}
\label{fig:flowchart}
\end{wrapfigure}

\textbf{Isolation of Conclusions Regarding Each Element of the Law.}
To determine whether Art.\ 11 has been violated, the Court proceeds through a strict sequence of legal tests (Figure~\ref{fig:flowchart}) across five canonical elements: \textbf{\textit{Scope}}, \textbf{\textit{Interference}}, \textbf{\textit{Prescribed by law}}, \textbf{\textit{Legitimate aim}} and \textbf{\textit{Necessity in a democratic society}}.
This stage anchors the backward modeling of the pipeline: once the Court’s conclusion on each element is determined, the subsequent stages only need to identify the facts that support those conclusions.
We pass the entire evaluation section to the LLM and ask it to extract the conclusion for each element (prompt in App.~\ref{app:prompts/conclusion}). The Court occasionally skips intermediate elements such as \textit{prescribed by law} or \textit{legitimate aim}, treating them as trivially established; a post-processing script fills in default values in such cases.
\textbf{\textit{Positive obligation is the sixth element we consider that represents an alternate route of reasoning.}} The Court asks whether the government should have protected an assembly rather than merely refrained from interfering with it. Since the Court’s jurisprudence does not break this route down into sub-elements in the first place, it cannot be evaluated at the same level of granularity as the other five elements, and we therefore treat it as a single, separate element.
For more details on the Art. 11 doctrine, see App.~\ref{app:doctrine}.

\textbf{Extracting Factual Groundings from the Evaluation Section of a Ruling.}
As we have explained, the Court's evaluation section is a rhetorical document that not only justifies the decision in terms of the canonical doctrine of Art. 11, but also aims at persuading the Court's multiple audiences. This is why directly matching atomic facts against the Court's evaluation section would produce too many mappings that do not support the conclusions. In creating the benchmark, separately for each element of doctrine, we extract the specific textual evidence the Court explicitly invoked to justify its conclusion on doctrinal grounds.
For terminological clarity, we call these the \textit{factual groundings} (prompt in App.~\ref{app:prompts/fact_grounding}; alternative approaches discussed in App.~\ref{app:benchmark/pipeline}).

\textbf{Mapping Factual Groundings to Atomic Facts.}
We ask the LLM to map the atomic facts to these factual groundings. This mapping is a non-trivial task since both the reporting and evaluation sections are interpretative acts in which the court essentially tells a story. In an attempt at discriminating between the persuasive overlay and the underlying legal argument, we create three alternative benchmarks against which we can evaluate the reasoning of downstream LLMs:

(i)~\textbf{List 1 (relevant facts):}
For every individual fact listed as \textit{factual grounding} (in the legal reasoning part of the decision), we ask the LLM to find all possible atomic facts that the
Court was pointing to.

(ii)~\textbf{List 2 (essential facts):}
We ask the LLM to focus on each element of doctrine that is
discussed by the Court, and find the atomic facts that are strictly necessary to support this
part of the reasoning of the court, grounded in the facts mentioned by the Court.
Where multiple atomic facts establish the same point, we include only the most critical
or representative facts.
By definition, List~2 $\subset$ List~1.

(iii)~\textbf{List 3 (decisive facts):}
We ask the LLM to focus on the final conclusion (Art. 11 ECHR violated or not) and find the atomic facts that are strictly necessary for the Court to come to this conclusion, assuming that the Court applies the doctrinal structure of Art. 11 ECHR. These are the facts the case actually turned on.
By design, List 3 $\subseteq$ List 2 $\subseteq$ List 1.

Each list targets a different point at which a model's legal reasoning can fail. List~2 tests whether, among relevant facts, the model can isolate those that are load-bearing for a specific legal element. List~3 tests whether the model can identify which facts determine the outcome of the case overall, independent of how those facts are distributed across elements. List~1 provides a valid recall upper bound for List~2.
The prompt is reported in App.~\ref{app:prompts/fact_mapping}.

\textbf{Implementation.}
Passing only the output of each stage to the next in a sequential pipeline did not produce reliable results. We therefore use a multi-turn chat setting, retaining prior inputs and outputs, with outputs represented as assistant turns to preserve LLM context across stages. Due to repeated connection timeouts on 10 cases, we excluded those, leaving a final benchmark of 375 cases. Pipeline design and implementation details are provided in App.~\ref{app:benchmark/pipeline} and App.~\ref{app:benchmark/implementation}, respectively.

\begin{table}[t]
\centering
\footnotesize
\setlength{\tabcolsep}{8pt}
\caption{Benchmark statistics across 375 cases.
Conclusion statistics use active cases \textit{only}; each element's avg. List~1 and 2 sizes exclude cases where the corresponding list is empty. Avg. List~1 and 2 sizes sum up list sizes across active elements, including empty ones, then average across cases.}
\label{tab:benchmark_stats}
\begin{tabular}{lrrrrr}
\toprule
& \multicolumn{3}{c}{\textbf{Conclusions}} & \multicolumn{2}{c}{\textbf{Avg facts}} \\
\cmidrule(lr){2-4} \cmidrule(lr){5-6}
\textbf{Element} & \textbf{Active} & \textbf{Yes} & \textbf{No} & \textbf{List 1} & \textbf{List 2} \\
\midrule
Scope               & $375$ & $375$ ($100\%$) &   $0$ ($0\%$)  & $6.82$ & $3.54$ \\
Interference        & $374$ & $374$ ($100\%$) &   $0$ ($0\%$)  & $5.43$ & $3.05$ \\
Prescribed by law   & $368$ & $309$ ($84\%$)  &  $59$ ($16\%$) & $8.10$ & $4.26$ \\
Legitimate aim      & $339$ & $332$ ($98\%$)  &   $7$ ($2\%$)  & $5.17$ & $2.90$ \\
Necessity           & $339$ &  $13$ ($4\%$)   & $326$ ($96\%$) & $9.45$ & $5.63$ \\
Positive obligation &  $28$ &  $19$ ($68\%$)  &   $9$ ($32\%$) & $11.54$ & $6.75$ \\
\midrule
\multicolumn{6}{l}{\textit{Case-level summary}} \\
\midrule
Violated            & \multicolumn{5}{l}{$361 / 375$ ($96.3\%$)} \\
Not violated        & \multicolumn{5}{l}{$14 / 375$ ($3.7\%$)} \\
Active elements/case & \multicolumn{5}{l}{Mean $4.86$ (min $2$, max $6$); $84.8\%$ have $5$ active elements} \\
Atomic facts/case   & \multicolumn{5}{l}{Mean $58.75$ (min $3$, max $574$)} \\
Avg List 1/case     & \multicolumn{5}{l}{$25.42$ relevant facts (summed across elements)} \\
Avg List 2/case     & \multicolumn{5}{l}{$14.10$ essential facts (summed across elements)} \\
Avg List 3/case     & \multicolumn{5}{l}{$6.21$ decisive facts} \\
\bottomrule
\end{tabular}
\end{table}

\subsection{Benchmark Statistics}
\label{sec:benchmark/stats}

Table~\ref{tab:benchmark_stats} summarizes the $375$-case benchmark. Violation outcomes are highly imbalanced ($96.3\%$), and most cases follow the standard five-element pathway. \textit{Scope} and \textit{Interference} always conclude affirmatively, while \textit{Necessity} concludes negatively in $96.2\%$ of active cases. \textit{Positive obligation} is active in only $28$ cases, representing its structurally distinct doctrinal route. The three-list funnel reduces an average of $58.75$ atomic facts per case to $25.42$ \textit{relevant} (List~1), $14.10$ \textit{essential} (List~2), and $6.21$ \textit{decisive} (List~3) facts. Further statistics are discussed in App.~\ref{app:benchmark/stats}.

\begin{table}[!t]
\centering
\footnotesize
\setlength{\tabcolsep}{8pt}
\caption{Pipeline validation by three legal experts across twenty cases. Overall reports the mean across experts with case-level bootstrap $95\%$ confidence intervals ($1{,}000$ resamples). Scores and intervals are percentages; Gwet's AC1 measures agreement between experts on a $0$--$1$ scale. %
}
\label{tab:validation}
\begin{tabular}{lrrrrr}
\toprule
\textbf{Metric} & \textbf{Expert 1} & \textbf{Expert 2} & \textbf{Expert 3} & \textbf{Overall [95\% CI]} & \textbf{AC1} \\
\midrule
Violation accuracy
& $100.0$ & $100.0$ & $100.0$
& $100.0\ [100.0, 100.0]$ & $1.000$ \\
Element conclusion accuracy
& $96.7$ & $94.1$ & $95.0$
& $95.2\ [91.8, 98.3]$ & $0.767$ \\
List 2 precision
& $93.0$ & $96.5$ & $91.3$
& $93.6\ [87.2, 97.2]$ & $0.936$ \\
List 3 precision
& $99.5$ & $98.2$ & $97.3$
& $98.3\ [96.9, 99.6]$ & $0.972$ \\
\bottomrule
\end{tabular}
\end{table}

\subsection{Validation of the Benchmark by Legal Experts} \label{sec:benchmark/validation}

Since the benchmark has been created using an LLM pipeline, it is important to validate its accuracy against human expert judgment. We employed three legal experts through Upwork, all holding master’s degrees in law from European institutions with significant ECtHR experience. Each expert was provided with the same twenty cases comprising the original judgment PDF and the pipeline outputs, along with a detailed annotation guide. Experts were instructed to evaluate pipeline outputs strictly against what the Court actually said. Five pilot cases were reviewed by the legal scholar on our team, and annotations were shared among experts to align their interpretation of the guidelines before they individually evaluated the remaining fifteen cases. Details are provided in Appendix~\ref{app:benchmark/validation}.

As Table~\ref{tab:validation} shows, the experts largely endorse the machine-generated conclusions and supporting facts.
Across \textit{all} items being rated, \textit{all} annotators found $\geq 90\%$ outputs to be acceptable (counting borderline ratings as acceptable).
Inter-expert agreement is also high (Gwet’s AC1~\citep{gwet2008computing}: $0.767-1.000$).
This validation assesses the appropriateness of proposed facts rather than their completeness; we did not measure recall, as exhaustively identifying missing facts would require a substantially more burdensome exercise.

\section{Downstream Evaluation}\label{sec:experiments}
\textbf{Experimental Setup.}
We evaluate ten LLMs across three experimental conditions that vary how much intermediate legal reasoning is required of the model, given the atomic facts of a case and instructions specifying the full Art. 11 doctrine (Figure~\ref{fig:flowchart}).
In the \textbf{direct violation (DV)} condition, the model only predicts whether Art. 11 has been violated with no intermediate reasoning (prompt in App.~\ref{app:prompts/ds_dv}). In this setup, we do not provide the doctrinal guidance.
In the \textbf{direct conclusions (DC)} condition, the model has to additionally make its conclusions on each legal element explicit (prompt in App.~\ref{app:prompts/ds_dc}).
In the \textbf{full reasoning (FR)} condition, we add one more step: the model must first map the atomic facts to every element of doctrine (prompt in App.~\ref{app:prompts/ds_vanilla}). All three conditions are evaluated in zero-shot mode with reasoning or extended thinking mode disabled. More details are in App.~\ref{app:implementation}.

\textbf{Models.}
We evaluate three closed-weight frontier models ({\tt Claude Sonnet 4.6}~\citep{anthropic2026sonnet46}, {\tt GPT-5.4}~\citep{openai2026gpt54}, {\tt Gemini 3 Flash}~\citep{google2025gemini3flash}) and seven open-weight models across three families ({\tt GPT-OSS} at 20B and 120B~\citep{openai2025gptoss120bgptoss20bmodel}; {\tt Qwen 3.5} at 9B, 35B, and 122B~\citep{qwen2026qwen3.5}; {\tt Gemma 4} at 4B and 31B~\citep{gemmateam2026gemma4}), across all 375 benchmark cases.
We also include two random baselines: a \textbf{uniform} baseline sampling from Bernoulli(0.5), and a \textbf{calibrated} baseline sampling from empirical label distributions (more details in App.~\ref{app/baselines}).

\textbf{Metrics.} We calculate the macro-F1 (\textit{m}F1) and accuracy (Acc) values for violation and element conclusions, which are binary classification tasks. Fact-mapping is a set retrieval task that is measured with F1.
For List~3, we take the union of the model's predicted facts across all six elements (details on the metrics in App.~\ref{app:fullresults/metrics}.
Table~\ref{tab:main_results} summarizes the results; more detailed results in App.~\ref{app:viol-conc}--\ref{app:elem-diff}.

\begin{table}[t]
  \caption{Performance on the 375 ECtHR Article 11 cases in terms of model predictions for the final \textbf{violation} (DV, DC, FR conditions), the \textbf{element conclusions} (DC, FR), and \textbf{fact-mapping} for FR (List~1: \textit{relevant}, List~2: \textit{essential}, List~3: \textit{decisive} facts). We report macro-F1 (\textit{m}F1) and accuracy (Acc) for violation and element conclusions, and F1 for fact mapping. Element conclusion, List~1 and List~2 scores are averaged over elements. \textbf{Bold} values show the best model per column.}
  \label{tab:main_results}
  \centering
  \footnotesize
  \setlength{\tabcolsep}{3.0pt}
  \begin{tabular}{lrr@{}p{8pt}@{}rr@{}p{8pt}@{}rr@{}p{15pt}@{}rr@{}p{8pt}@{}rr@{}p{15pt}@{}rrr}
    \toprule
     & \multicolumn{8}{c}{\textbf{Violation}} & & \multicolumn{5}{c}{\textbf{Element conclusions}} & & \multicolumn{3}{c}{\textbf{Fact mapping F1}} \\
    \cmidrule(l{3pt}r{3pt}){2-9}\cmidrule(l{3pt}r{3pt}){11-15}\cmidrule(l{3pt}r{3pt}){17-19}
     & \multicolumn{2}{c}{DV} & & \multicolumn{2}{c}{DC} & & \multicolumn{2}{c}{FR} & & \multicolumn{2}{c}{DC} & & \multicolumn{2}{c}{FR} \\
    \cmidrule(l{3pt}r{3pt}){2-3}\cmidrule(l{3pt}r{3pt}){5-6}\cmidrule(l{3pt}r{3pt}){8-9}\cmidrule(l{3pt}r{3pt}){11-12}\cmidrule(l{3pt}r{3pt}){14-15}
    Model & \textit{m}F1 & Acc &  & \textit{m}F1 & Acc &  & \textit{m}F1 & Acc &  & \textit{m}F1 & Acc &  & \textit{m}F1 & Acc &  & List~1 & List~2 & List~3 \\
    \midrule
    \multicolumn{19}{l}{\textit{Baselines}} \\
    Uniform & $36.3$ & $50.0$ &  & $36.3$ & $50.0$ &  & $36.3$ & $50.0$ &  & $38.0$ & $49.9$ &  & $38.0$ & $49.9$ &  & $29.8$ & $25.3$ & $29.6$ \\
    Calibrated & $50.0$ & $92.7$ &  & $50.0$ & $92.7$ &  & $50.0$ & $92.7$ &  & $50.2$ & $86.7$ &  & $50.2$ & $86.7$ &  & $34.7$ & $29.9$ & $33.2$ \\
    \midrule
    \multicolumn{19}{l}{\textit{Closed weights}} \\
    Claude Sonnet 4.6 & $53.5$ & $94.9$ &  & $53.5$ & $94.9$ &  & $48.4$ & $93.9$ &  & $52.3$ & $83.8$ &  & $50.0$ & $\mathbf{81.0}$ &  & $51.5$ & $44.4$ & $57.9$ \\
    GPT-5.4 & $65.9$ & $94.9$ &  & $52.9$ & $94.4$ &  & $52.9$ & $94.4$ &  & $50.5$ & $80.8$ &  & $41.8$ & $63.2$ &  & $\mathbf{54.3}$ & $48.8$ & $63.0$ \\
    Gemini 3 Flash & $\mathbf{79.4}$ & $\mathbf{96.5}$ &  & $\mathbf{73.9}$ & $\mathbf{95.7}$ &  & $\mathbf{63.5}$ & $\mathbf{94.9}$ &  & $\mathbf{59.2}$ & $\mathbf{86.5}$ &  & $\mathbf{51.8}$ & $78.9$ &  & $51.2$ & $47.2$ & $57.6$ \\
    \midrule
    \multicolumn{19}{l}{\textit{Open weights}} \\
    GPT OSS 20B & $30.3$ & $41.3$ &  & $33.4$ & $45.9$ &  & $31.3$ & $43.2$ &  & $41.6$ & $62.6$ &  & $39.6$ & $60.4$ &  & $27.2$ & $26.2$ & $46.9$ \\
    GPT OSS 120B & $39.3$ & $56.5$ &  & $48.0$ & $82.4$ &  & $39.3$ & $61.3$ &  & $47.1$ & $72.6$ &  & $43.0$ & $63.4$ &  & $32.0$ & $30.3$ & $50.1$ \\
    Qwen 3.5 9B & $33.2$ & $46.9$ &  & $36.3$ & $57.1$ &  & $43.9$ & $78.1$ &  & $45.7$ & $73.5$ &  & $45.3$ & $73.2$ &  & $44.7$ & $39.4$ & $50.2$ \\
    Qwen 3.5 35B & $36.4$ & $51.7$ &  & $43.9$ & $78.1$ &  & $35.2$ & $54.4$ &  & $42.7$ & $66.6$ &  & $41.6$ & $60.7$ &  & $31.6$ & $29.5$ & $29.3$ \\
    Qwen 3.5 122B & $36.8$ & $49.3$ &  & $55.5$ & $93.6$ &  & $34.7$ & $53.1$ &  & $45.9$ & $76.7$ &  & $41.8$ & $68.7$ &  & $44.6$ & $40.5$ & $63.0$ \\
    Gemma 4 4B & $30.2$ & $37.9$ &  & $48.4$ & $93.9$ &  & $32.9$ & $46.4$ &  & $44.8$ & $73.1$ &  & $43.9$ & $71.2$ &  & $39.7$ & $39.9$ & $62.8$ \\
    Gemma 4 31B & $37.4$ & $55.7$ &  & $54.4$ & $90.1$ &  & $59.2$ & $94.1$ &  & $45.4$ & $69.3$ &  & $44.8$ & $69.5$ &  & $51.3$ & $\mathbf{49.7}$ & $\mathbf{67.1}$ \\
    \bottomrule
  \end{tabular}
\end{table}

\subsection{Results} \label{sec:results}

\textbf{Additional intermediate reasoning does not consistently improve legal judgment.}
Predicting the element conclusions as an intermediate step (DV $\rightarrow$ DC) leads to an improvement in violation prediction for \textit{all} 7 open weight models (avg. $\textcolor{green!60!black}{\bm{\uparrow}} 31.3 \%$ \textit{m}F1), which have weak violation performance in DV.
However, the performance of the closed models actually decreases or remains unchanged (avg. $\textcolor{red!60!black}{\bm{\downarrow}} 8.9 \%$ \textit{m}F1).
With the additional burden of identifying the supporting facts (DC $\rightarrow$ FR), the performance decreases further for 7/10 models.
Further, \textit{all} models perform worse in element conclusion for FR compared to DC (avg. $\textcolor{red!60!black}{\bm{\downarrow}} 6.3 \%$ \textit{m}F1).
Thus, requiring the models to explicitly reason through the intermediate steps does \textit{not} necessarily improve legal judgment.

\begin{wrapfigure}{r}{0.57\textwidth}
\centering
\includegraphics[width=0.57\textwidth]{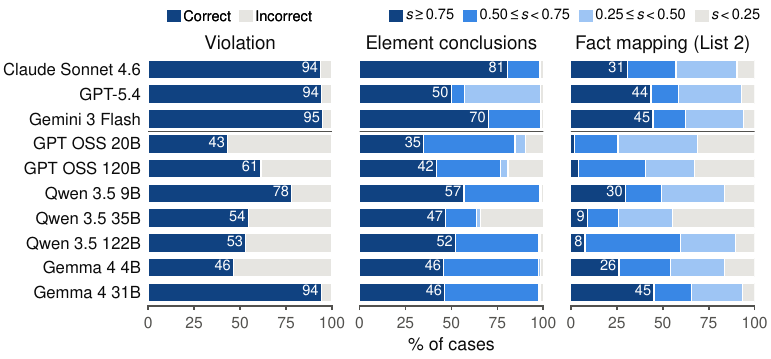}
\caption{Case-level performance in the FR condition. Bars show the $\%$ of cases with correct \textit{violation} predictions and the distribution of avg. accuracy scores for \textit{element conclusions} and avg. \textit{fact-mapping} F1 (List~2). Scores for the latter two tasks are grouped into $4$ intervals.}
\label{fig:task-breakdown}
\end{wrapfigure}

\textbf{Models struggle most in identifying supporting facts.}
To further establish the varying capabilities of the models for the three tasks, we ask, \textit{how often do models produce high-scoring outputs for each task}?
We analyze all model outputs in the FR condition in the following way:
(i)~\textit{violation}: the percentage of cases the model output matches the ground truth; (ii)~\textit{element conclusions} and \textit{fact-mapping}: the average accuracy and average List~2 (essential facts) F1 across the elements. We group the latter two scores into intervals and plot the percentage of cases in each interval (Figure~\ref{fig:task-breakdown}).
The closed weight models correctly predict the violation for at least $94\%$ cases, while having avg. element conclusion accuracy $\geq 0.75$ on $50-81\%$ cases.
However, they are able to cross the fact-mapping F1 $\geq 0.75$ threshold on only $31-45\%$ cases.
Across \textit{all} models, a greater proportion of cases reach a score $\geq 0.75$ for element conclusions than for fact mapping.
This gap persists even if the threshold of a high-scoring output is reduced to $\geq 0.50$.
Further, 9/10 models perform poorer on List~2 compared to List~3, which suggests that attributing the \textit{essential facts} to legal elements poses an additional challenge compared to identifying the overall \textit{decisive facts}.
Evaluating just the decisive facts would miss this weakness in element-wise factual support.
Across all models, precision is lower than recall for both Lists~2 and 3, indicating a tendency to include facts beyond those identified by the Court as supporting the corresponding conclusions (Figure~\ref{fig:precision-recall}, App.~\ref{app:fact-mapping}).
These results reveal a gap between models’ ability to predict legal conclusions and their ability to support those conclusions with the relevant facts, even when explicitly required to do so.

\textbf{Model rankings differ between conclusion prediction and fact mapping.}
Comparing {\tt Gemini 3 Flash} and {\tt Gemma 4 31B}, among the strongest closed and open weight models respectively, we observe differences in their relative performance across tasks.
The former beats the latter on both violation prediction ($\textcolor{green!60!black}{\bm{\uparrow}} 7.2 \%$ \textit{m}F1) and element conclusion ($\textcolor{green!60!black}{\bm{\uparrow}} 15.6 \%$ \textit{m}F1).
However, when it comes to fact-mapping, the performance is similar or even worse (avg. $\textcolor{red!60!black}{\bm{\downarrow}} 6.5 \%$ F1 across the three lists), with the largest difference for List~3 (decisive facts).
We also observe a similar trend within the closed weight models themselves: {\tt GPT-5.4} scores lower than {\tt Gemini 3 Flash} on both violation and element conclusions but higher on all 3 fact-mapping lists.
These comparisons show that models that are relatively better at predicting conclusions are not necessarily better at mapping facts.

\textbf{High accuracy does not necessarily guarantee improved performance over the minority class.}
In the FR condition, the closed-weight models achieve similar violation accuracy ($93.9-94.9\%$), despite substantial differences in \textit{m}F1 ($48.4-63.5\%$).
Across all models, majority-class F1 exceeds minority-class F1 by an average of $74.6$ percentage points for violation prediction and $62.5$ percentage points for element conclusions (Figure~\ref{fig:majority-minority-viol-conc}, App.~\ref{app:viol-conc}).
Even {\tt Gemini 3 Flash}, which achieves the highest minority-class F1 for both tasks, reaches only $29.6\%$ and $20.0\%$ respectively.
Thus, high overall accuracy can conceal a shared weakness across models in identifying minority-class outcomes.

\subsection{Element-wise analysis} \label{sec:elem-analysis}
\begin{wrapfigure}{r}{0.47\textwidth}
\centering
\includegraphics[width=0.47\textwidth]{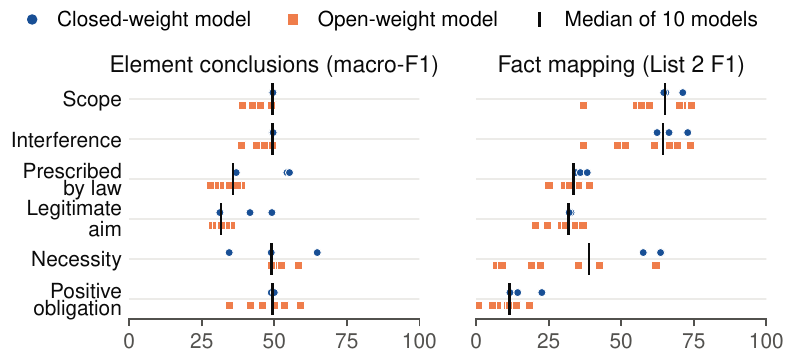}
\caption{Element-wise FR performance for element conclusions (\textit{m}F1) and essential-fact mapping (List~2 F1). Each marker represents a model. Scope and Interference contain only one ground-truth class, limiting their \textit{m}F1 to $50\%$.}
\label{fig:elem-diff}
\end{wrapfigure}
Figure~\ref{fig:elem-diff} shows that different elements of doctrine pose different challenges for LLMs in the FR condition. The models perform reasonably well on elements that allow for a rather straightforward yes/no response (\textit{Scope, Interference}). Performance is less consistent for \textit{Prescribed by law} and \textit{Legitimate aim}, which involve multiple subcategories: even models with stronger conclusion scores struggle with fact-mapping.
The openly judgmental elements reveal two distinct patterns. \textit{Necessity}, which requires complex balancing, exposes substantial differences between models, both in identifying minority-class conclusions (Table~\ref{tab:element_confusion_fr}, App.~\ref{app:elem-diff}) and in mapping essential facts (List~2 F1 ranges from $7\%$ to $63.7\%$). By contrast, \textit{Positive obligation}, which is open textured, poses a shared difficulty, with all models performing poorly on List~2.

\subsection{Generalizability of the Benchmark Creation Pipeline} \label{sec:generalizability}

\begin{figure}
    \centering
    \includegraphics[width=0.95\linewidth]{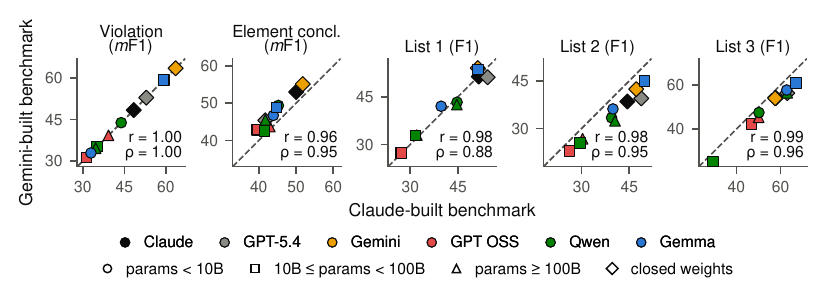}
    \caption{Model performance (\%) in FR condition for {\tt Claude}-built and {\tt Gemini}-built benchmarks on the same cases. Pearson’s $r$ and Spearman’s $\rho$ summarize score and rank agreement, respectively.}
    \label{fig:benchmark-alignment}
\end{figure}

\textbf{Model rankings remain largely stable when the benchmark-construction model changes.}
To assess the sensitivity of downstream evaluation to the model creating the benchmark, we construct alternative reference annotations using {\tt Gemini 3 Flash}.
Specifically, we repeat Stages 3--5 of the procedure described in Section~\ref{sec:benchmark} (Figure~\ref{fig:pipeline}) using {\tt Gemini} instead of {\tt Claude}.
We retain the atomic facts extracted by {\tt Claude} to isolate the effect of the construction model on conclusion extraction, factual grounding, and fact mapping.
The resulting reference annotations show high agreement with the original annotations --- Gwet's AC1: $1.00$ for violation, $0.85$ for element conclusions, and $0.92$--$0.97$ for fact-mapping Lists~1--3.
As Figure~\ref{fig:benchmark-alignment} shows, downstream model scores are strongly correlated across the two benchmarks (Pearson's $r$: $0.96$--$1.00$), and model rankings remain largely stable (Spearman's $\rho$: $0.88$--$1.00$).
Neither construction model gains a consistent advantage across tasks, although {\tt Gemini} rises from fourth to first on List~1 when evaluated against its own reference.
These analyses suggest that the comparative findings are largely preserved when the construction model changes, despite some changes in individual task rankings.
More details are in App.~\ref{app:general/gemini}.

\begin{wraptable}{r}{0.45\textwidth}
  \caption{Results on Art. 10 benchmark (FR condition). All values in percentage.}
  \vspace{0.5em}
  \label{tab:art10}
  \centering
  \footnotesize
  \setlength{\tabcolsep}{3pt}
  \begin{tabular}{lrrrrr}
    \toprule
     & \multicolumn{2}{c}{Macro-F1} & \multicolumn{3}{c}{Fact mapping F1} \\
    \cmidrule(lr){2-3}\cmidrule(lr){4-6}
    Model & Viol. & Elem. & L1 & L2 & L3 \\
    \midrule
    \multicolumn{6}{l}{\textit{Closed weights}} \\
    Claude Sonnet 4.6 & $\mathbf{87.5}$ & $62.2$ & $35.6$ & $26.4$ & $25.3$ \\
    GPT-5.4 & $75.0$ & $52.9$ & $37.4$ & $28.8$ & $28.7$ \\
    Gemini 3 Flash & $79.7$ & $56.0$ & $\mathbf{41.3}$ & $\mathbf{36.0}$ & $33.2$ \\
    \midrule
    \multicolumn{6}{l}{\textit{Open weights}} \\
    GPT OSS 20B & $59.9$ & $46.1$ & $20.8$ & $19.6$ & $20.9$ \\
    GPT OSS 120B & $82.5$ & $\mathbf{68.3}$ & $28.5$ & $24.8$ & $24.1$ \\
    Qwen 3.5 9B & $70.2$ & $56.4$ & $31.0$ & $24.6$ & $24.8$ \\
    Qwen 3.5 35B & $58.5$ & $48.7$ & $33.5$ & $29.3$ & $26.4$ \\
    Qwen 3.5 122B & $70.2$ & $52.7$ & $35.7$ & $29.4$ & $28.2$ \\
    Gemma 4 4B & $68.8$ & $52.2$ & $23.0$ & $21.2$ & $21.1$ \\
    Gemma 4 31B & $59.2$ & $49.5$ & $36.4$ & $35.9$ & $\mathbf{33.8}$ \\
    \bottomrule
  \end{tabular}
\end{wraptable}

\textbf{The pipeline can be transferred to other legal provisions with minimal changes.}
To assess transferability, we applied the full pipeline to 26 recent Article 10 (\textit{freedom of expression}) cases with {\tt Claude Sonnet 4.6}, of which 25 passed the selection criteria.
Art. 10 shares Art. 11’s broad legal-test structure but includes additional \textit{legitimate aims}, such as preventing disclosure of confidential information and maintaining judicial authority and impartiality (details in App.~\ref{app:general/art10}).
Adapting the pipeline required changes only to article-specific element definitions in the schemas and prompts, keeping its logic and structure unchanged.
Compared to Art. 11, this sample contains 2.6 times as many atomic facts per case ($152.52$ versus $58.75$), and a less pronounced violation imbalance ($76.0\%$ versus $96.3\%$).
We also repeated the benchmark construction procedure with {\tt Gemini 3 Flash} and observed a high agreement between the two across tasks (Gwet's AC1~\citep{gwet2008computing}: $0.91-1.00$).
Downstream evaluation under FR against the {\tt Claude}-built benchmark again shows the models struggle in terms of factual support (Table~\ref{tab:art10}).
List~2 F1 ranges from $19.6-36.0\%$ and is lower than on Art. 11 for every model.
The small sample and lack of expert validation make this a preliminary transfer study. Nevertheless, it demonstrates that the same pipeline can be extended for a second legal provision.

\subsection{Additional Experimental Setups}

We examine whether alternative prompting and reasoning setups alleviate the difficulty of fact mapping (App.~\ref{app:additional_expt}).
Our main evaluation supplies doctrinal questions and decision logic to avoid relying on models’ internal knowledge.
Although removing this guidance improves fact mapping for most models, List~2 F1 remains at or below $51.1\%$, showing that the difficulty persists without full scaffolding (App.~\ref{app:additional_expt/scaffolding}).
We also investigate staged prediction (App.~\ref{app:additional_expt/staging}), extended thinking (App.~\ref{app:additional_expt/thinking}), and few-shot prompting (App.~\ref{app:additional_expt/few-shot}).
Staging separates fact mapping from element-conclusion and violation prediction into two turns.
This substantially improves violation prediction for some models, with smaller gains in element conclusions, whereas extended thinking improves both for all six tested models.
However, neither intervention consistently improves fact mapping.
Five-shot prompting improves List~2 F1 for seven of ten models and violation prediction for all ten.
Nevertheless, no model exceeds $51.4\%$ essential-fact F1 across these additional setups.
These methods, despite yielding performance gains, struggle to attribute supporting facts in concluding the respective legal elements.

\section{Concluding Discussion} \label{sec:discussion}
\textbf{Summary.}
The machine learning community has originally focused on judgment prediction. This completely brackets legal reasoning, i.e. the way how the conclusion is reached. Later work has checked whether downstream LLMs assess individual elements of the applicable legal rule the same way as a benchmark ruling from a real court. But some toy examples notwithstanding, the literature has not addressed the core of legal decision-making: the selective mapping from a rich set of potentially relevant facts to the individual elements of a legal rule. This paper directly addresses this gap. We introduce the first gold-standard benchmark for the task, built on 375 ECtHR Article 11 judgments, and provide a multi-stage pipeline for constructing such benchmarks at scale.

\textbf{Contributions and insights.}
In the benchmark construction phase, directly matching atomic facts from the reporting section to the evaluation section leads to over-mapping, because the Court mentions additional facts in its reporting section for rhetorical purposes, apart from supporting individual element conclusions.
The backward modeling approach, which first isolates element-level conclusions and then extracts the specific factual grounding the Court invoked for each, addresses this problem and is potentially applicable to other jurisdictions whose judgments separate factual reporting from doctrinal evaluation.
The downstream evaluation reveals decreasing competence across three levels of legal reasoning: models are more successful at predicting violations than at assessing individual legal elements, and struggle most with fact-to-element attribution. The three-condition evaluation further shows that explicitly requiring intermediate reasoning does not uniformly improve conclusion prediction. Even models that correctly predict violations through approximately correct element conclusions often select the wrong supporting facts or attribute them to the wrong elements.
Reconstructing the benchmark with a different LLM yields similar performance trends, suggesting that the findings are not specific to the construction model.
Standard outcome-level evaluation cannot detect this weakness, and our benchmark uniquely makes it visible.

\textbf{Future work and broader impacts.}
Our preliminary extension to Art. 10 (\textit{freedom of expression}) motivates broader evaluation on other ECHR provisions, such as Art. 5 (\textit{right to liberty}).
With minimal modification, the pipeline could also be adapted to legal rules from other Civil Law jurisdictions , and more tentatively, to semi-official documents like the Restatements of Law, that try to distill more abstract principles from common law cases.
From a broader perspective, LLMs are increasingly used in the court room (e.g. in Shenzhen, in Slovenia, in Israel, for some case categories in Germany). Our paper raises a concern.
Legal doctrine is not merely a tool for a court for producing the outcome that the rule-giving institution (the Member States ratifying the ECHR in our case) would have wanted.
Doctrine is also meant to constrain the decision making power of the court, and to make transparent why the decision has been taken.
It is worrisome if this check gets lost.

\textbf{Limitations.}
Several limitations deserve attention. First, the rhetorical character of legal reasoning means that perfect replication of the court's fact mapping by a downstream LLM cannot be expected, introducing irreducible noise into the evaluation. Individual case results should therefore be interpreted cautiously; the benchmark is most meaningful as a corpus-level instrument, and confidence in a model's capability should be built across sufficiently large and homogeneous collections of cases rather than individual judgments.

Second, in areas of law where doctrine is contested, the element structure itself is unstable, and the pipeline's conclusions about which facts are necessary for which elements would be less reliable.

Third, we have constrained our pipeline to a single human right, albeit one with a rich and differentiated doctrinal substructure; the Article 10 extension is preliminary. A natural extension would be to cases that can be plausibly brought under the purview of more than one human right, evaluating how well downstream LLMs reason from atomic facts to the selection of the most pertinent legal rule.

Fourth, memorization remains a concern.
The judgments are publicly available in HUDOC and may have been included in models’ training data.
Two observations temper this concern: even with potential memorization, frontier models perform poorly on fact-mapping, and the fact-to-element attribution failures are not consistent with a model that has simply memorized the judgments.
Moreover, this paper serves as a proof-of-concept for the pipeline and evaluation methodology.
We also report a small exploratory evaluation on judgments published after the models' disclosed training cutoffs
(App.~\ref{app:additional_expt/memorization}), but the sample is too small to draw conclusions about memorization.
Future work could use synthetic cases based on real rulings or mock cases so designed that the mappings are known by construction, to reduce potential memorization advantages.

\begin{ack}
This project is funded in part by the AI, Computing and Society Initiative of the Max Planck Society.
\end{ack}

\bibliographystyle{plainnat}
\bibliography{references}

\clearpage
\newpage

\appendix

\section{Extended Discussion on Related Work}
\label{app:relwork}

Efforts in Legal AI research has transformed from predicting judicial outcomes to reason in a structured way, faithful to the practice in law.
This is motivated by a concern regarding how AI models being deployed for legal applications are evaluated.
A model may reproduce a court's conclusion without identifying the facts that the court used to justify it.
To assess legal reasoning, one must examine the outcome predicted by the model, the conclusion reached by the model on individual elements of the legal doctrine, and whether the model is able to ground those conclusions in the facts, similar to the Court.
In the literature survey below, we study four strands of research that aim to address different aspects of the problem.

\subsection{Legal Judgment Prediction and Decomposition of Legal Tasks}

Early works on Legal Judgment Prediction (LJP) focussed on predicting the verdict of the case directly given the facts~\citep{aletras2016predicting,chalkidis2019neural}.
These works established that automatic prediction of the verdict for ECtHR cases is feasible, with the focus of the evaluation being the final verdict.
Subsequently, \citet{medvedeva2020using} showed that in these scenarios, the models may base their predictions (often, correct) on the identity of the judges rather than the facts of the case.
Other works recognized the need to decompose the prediction setup into intermediate steps.
TopJudge~\citep{zhong2018topjudge} models the dependencies between the intermediate tasks such as applicable legal articles, charges and prison terms using directed acyclic graphs.
NeurJudge~\citep{yue2021neurjudge} focused on separating the facts into the different circumstances needed to deal with each different intermediate task.
\citet{liu2025lfp} focus their attention more upstream, by predicting what facts should at all be presented as inputs to LJP models based on pre-trial evidence.
However, none of these works focused on evaluating whether the models can ground their decisions on the correct facts.

\subsection{LLM Legal Reasoning: Syllogism and Step-by-Step Evaluation}

\textbf{Legal Competence of LLMs.}
The advent of LLMs shifted the attention from supervised prediction to leveraging the legal knowledge acquired by the models through their extensive pre-training procedures.
\citet{choi2022chatgpt} found that ChatGPT could obtain a passing grade in multiple choice questions and legal essay writing from law school courses.
While \citet{katz2024gpt4} reported that GPT-4 passed the Uniform Bar Exam at the 90th percentile compared to the repeat test takers, \citet{martinez2025} audited these results and clarified that the performance is only at 62nd percentile compared to the first-time test takers.

\textbf{Introducing Structure to Legal Reasoning.}
Another line of work attempted to structure legal reasoning by infusing syllogism, i.e., treating the legal rules as the major premise and the case facts as the minor premise and reasoning through them to arrive at the conclusion.
Prompting LLMs to explicitly follow such reasoning structures improves over generic chain-of-thought, as observed for legal entailment~\citep{yu2022legal} as well as LJP~\citep{jiang2023legal}.
\citet{deng2024adapt} create a framework for decomposing the facts into individual elements which are treated as the minor premise independently to better distinguish between confusing charges, and used fine-tuning over synthetic reasoning traces to improve performance.
On the other hand, \citet{zhang2025syler} combined fine-tuning with reinforcement learning to make use of relevant legal rules and precedents to derive the major premises to provide more legally grounded answers in QA tasks.

\textbf{Benchmarks for Legal Reasoning.}
The advent of LLMs has broadened the scope of evaluation beyond final outcomes to specific legal abilities and intermediate reasoning.
\citet{guha2023legalbench} introduced $162$ different tasks centred around the \textbf{IRAC} framework, which involves the ability to identify the \textbf{I}ssues, recall the relevant legal \textbf{R}ule, \textbf{A}pply the rule to given case facts and arrive at the legal \textbf{C}onclusion.
A series of works investigated model performance on different tasks in the Chinese legal domain that fall under different stages of legal syllogism, such as memorizing legal rules and principles, application of rules to cases through prediction of articles, charges and penalty terms, and advanced tasks such as analyzing and generating judicial reasoning and legal QA~\citep{fei2024lawbench,li2024lexeval,dai2025laiw}.
Focusing particularly on the four elements of Chinese criminal law (subject, object, subjective aspect and objective aspect), \citet{liu2025jurex} focus on the major premise by explicitly defining the four elements via expert annotations for $155$ criminal charges, while \citet{deng2023syllogistic} focus on the minor premise by using ChatGPT with further manual corrections to annotate the four elements for each specific case before downstream benchmarking.
\citet{yu2025mslr} make use of a collaborative human-LLM pipeline to annotate expert reasoning traces of existing cases with the IRAC framework to enable downstream benchmarking.
Separate from the above works, \citet{yang2026appgen} focus on a different stage in the litigation process by investigating the capability of LLMs to generate the appellate decisions given the initial decision and evidentiary updates.

The closest works examine different parts of the legal reasoning process. \citet{liu2025jurex} provide expert definitions of doctrinal elements, while \citet{deng2023syllogistic} evaluate case-specific analysis of the doctrinal elements.
\citet{yu2025mslr} evaluate intermediate reasoning through IRAC annotations.
By contrast, our benchmark makes the factual basis of each doctrinal conclusion an explicit evaluation target. It provides mappings from individual atomic case facts the Court used to conclude the corresponding elements of law.
We score these mappings separately from element-level conclusions and the final violation prediction.
This allows us to distinguish whether a model agrees with the Court's conclusions from whether it grounds those conclusions in the same facts.

\subsection{ECtHR cases as a Testbed for Structured Legal Reasoning}

Judgments from the European Court of Human Rights (ECtHR) have been widely used in legal AI research.
The judgments are publicly available in English, and the Court follows a recurring doctrinal framework.
The Court also addresses multiple Convention articles with varying degrees of complexity, making its judgments a useful testbed for evaluating and developing legal AI applications.
Early works established the prediction of whether a human right has been violated as the primary task~\citep{aletras2016predicting,chalkidis2019neural}.
\citet{chalkidis2021paragraph} further annotated paragraphs in the reporting section as rationales for the \textit{alleged} violations, based on whether those paragraph numbers are explicitly cited by the Court in its evaluation section.
\citet{santosh2022deconfounding} established that models can exploit legally irrelevant signals to make violation predictions, and improved the alignment with expert-annotated rationales through adversarial training.
\citet{xu2023dissonance} examined disagreements between legal experts over which facts support violation predictions, highlighting the difficulty of defining reference rationales for evaluation.
\citet{xu2023vechr} introduced a dataset in which experts identify the types of vulnerability affecting people in ECtHR cases and mark the text supporting those labels. They evaluate whether models can identify these vulnerability types and select the same supporting text as the experts.

Another line of work inspected the legal arguments used in the Court, by annotating these arguments and training models to identify them~\citep{habernal2024mining}, as well as testing if LLMs can select the correct next argument based on the case facts and preceding arguments~\citep{chlapanis2024lar}.
Specifically for Art. 11 ECHR, \citet{engel2024professor} investigated whether an LLM can write a legal commentary explaining how the Court applies Art. 11 based on prior judgments, organizing the discussion around the legal questions considered by the Court.

Closest to our work, \citet{chalkidis2021paragraph} identify paragraphs supporting alleged violations, while \citet{xu2023dissonance} show that experts can disagree about which facts matter for case outcomes.
Our benchmark has a narrower focus on the facts the Court states it relied on, rather than those it could or should have relied on.
We identify these factual grounds at the level of atomic facts and link them to individual doctrinal elements.
This allows us to evaluate whether models reproduce the Court's stated factual grounds for each doctrinal conclusion.

\subsection{Atomic Decomposition and Fact-Law Mapping}

Atomic decomposition makes it possible to evaluate individual claims within a longer text.
\citet{min2023factscore} introduced FActScore, which splits generated text into atomic claims and checks whether each claim is supported by a knowledge source.
VeriScore~\citep{song2024veriscore} focuses on extracting claims that can be verified, while SAFE~\citep{wei2024safe} uses an LLM to search for evidence and assess factual support.
In the legal domain, \citet{gray2024llm} used LLMs to discover legal factors and their definitions from court opinions, evaluating their usefulness for outcome prediction against expert-defined factors.
ARC~\citep{elaraby2026arc} uses atomic representations of arguments to assess how well legal summaries preserve the arguments in the original documents.
Our use of atomic facts serves a different purpose.
We decompose the reporting section of a judgment into individual facts and evaluate which of them a model assigns to each doctrinal element.

\subsection{Our Contribution}

The literature reviewed above provides methods for structuring legal reasoning, annotating factual rationales, and evaluating intermediate conclusions. Our contribution brings these concerns together by explicitly linking atomic case facts to doctrinal elements and evaluating those mappings along with element-level conclusions and the final violation decision.
This allows us to locate where agreement with the Court breaks down. A model may reach the same final decision while drawing different conclusions on individual elements, or reach the same doctrinal conclusions while grounding them in different facts.
We make these distinctions measurable against the Court's stated reasoning at all three levels.
We also provide an automated construction pipeline, with expert validation of the resulting annotations. Art. 11 ECHR serves as a proof of concept. The pipeline can be adapted to other Convention articles, as shown in our preliminary study of Art. 10, or legal provisions with an established doctrinal structure.

\section{Canonical Elements and Doctrinal Chain of Article 11 ECHR}
\label{app:doctrine}

This Appendix describes the six canonical elements used in our benchmark and the doctrinal chain connecting them. The framework is based on the Court Registry's Guide on Article 11~\footnote{\url{https://ks.echr.coe.int/documents/d/echr-ks/guide_art_11_eng-pdf}}, focusing on \textit{freedom of peaceful assembly}. We explain each canonical element and how it is represented in our benchmark.

\begin{description}

\item[\textbf{(a) Scope.}] The scope element asks three questions. First, the substantive question: has there been an assembly, meaning a gathering of individuals for a common expressive or communicative purpose? Second, the personal question: did the applicant participate or attempt to participate in the assembly? Third the question of peaceful assembly: was the assembly peaceful in nature, meaning that the participants did not themselves engage in violence or incitement to violence? All three conditions must be satisfied for the applicant's conduct to fall within the scope of protections of Article 11. A negative answer to any of them terminates the analysis without a finding of violation.

\item[\textbf{(b) Interference.}] If scope is established, the Court asks whether some act or omission by a state authority interfered with the applicant's exercise of the right to freedom of assembly. Interference can take many forms, including dispersal of an assembly, imposition of restrictions on its location, time or manner, arrest or detention of participants, or failure by the state to protect an assembly from disruption by hostile third parties. The threshold is quite low; any measure that restricts or penalises participation in an assembly qualifies as interference.

\item[\textbf{(c) Prescribed by law.}] Once interference is established, the Court examines whether it had a sufficient legal basis. This requires three cumulative conditions to be met. First, the legal provision empowering the government to interfere must be sufficiently foreseeable: a person must be able to regulate their conduct with reference to its wording, with the assistance of legal advice if necessary. Second, the provision must be adequately accessible to the public. Third, it must be accompanied by sufficient procedural safeguards against arbitrary or abusive application by the authorities. If any of these three conditions is not satisfied, the interference was not prescribed by law and Article 11 has been violated.

\item[\textbf{(d) Legitimate aim.}] If the interference was prescribed by law, the Court asks whether it pursued one of the legitimate aims exhaustively listed in Article 11(2) ECHR: national security, public safety, the prevention of disorder or crime, the protection of health or morals, and the protection of the rights and freedoms of others. The Court applies this test with a relatively low threshold; governments almost always succeed in identifying a plausible legitimate aim. A finding that no legitimate aim was pursued is rare but results in a violation.

\item[\textbf{(e) Necessity in a democratic society.}] This is the most substantive, and often, the element that determines the outcome of the analysis. The Court asks whether the interference was proportionate to the legitimate aim pursued, i.e., whether the government adequately balanced the intensity of the restriction against the weight of the interest it sought to protect. The assessment takes into account the severity of the measure, the importance of the aim pursued by the state, whether less restrictive alternatives were available, and the margin of appreciation afforded to the state. The Court finding the interference to be disproportionate, i.e., \textit{not} necessary in a democratic society, is the most common situation for Article 11 violations.

\item[\textbf{(f) Positive obligation.}] Positive obligation is an alternate route of reasoning on Art. 11 cases. Rather than asking whether the government actively interfered with freedom of assembly in an unjustified way, the Court asks whether the government was under an obligation to take positive steps to facilitate or protect the exercise of the right, and whether it failed to discharge that obligation. This route is engaged when the complaint concerns government inaction or failure to protect an assembly from disruption by hostile actors. However, this cannot be evaluated at the same level of granularity as the remaining elements (as this granularity does not exist in the Court's jurisprudence in the first place), and hence we consider this to be an independent element of doctrine.

\end{description}

\paragraph{Doctrinal chain.}
The analysis of the Court proceeds based on chain of logic defined by the doctrinal elements.
Scope and interference act as logic gates: if scope is not established, there is no right to protect, and if interference is not established, there is nothing to justify from the perspective of the state.
A negative answer to either terminates the analysis without a finding of violation. If both are established, the Court proceeds to the three justification elements: prescribed by law, legitimate aim, and necessity in a democratic society. A negative conclusion on any one of them is sufficient to establish that the interference was unlawful, and Article 11 has been violated.
Positive obligation is an alternate route that is assessed independently.
This conditional structure means that different cases discuss different subsets of the six elements. Our benchmark captures this phenomenon through the active case counts reported in Table~\ref{tab:benchmark_stats}.

\section{Additional Details on Benchmark Creation}
\label{app:benchmark}

\begin{wraptable}{r}{0.45\textwidth}
\caption{Number of cases satisfying each sample selection criterion individually and all criteria jointly.}
\label{tab:sample_selection}
\centering
\footnotesize
\begin{tabular}{lc}\toprule
Criterion & Satisfied cases \\\midrule
Freedom of assembly & 669 \\
Uniform conclusions & 585 \\
Undisputed facts    & 726 \\
Genuine merits discussion & 528 \\\midrule
All criteria & 385 \\\bottomrule
\end{tabular}
\end{wraptable}

\subsection{Sample Selection}
\label{app:benchmark/sample}

Table~\ref{tab:sample_selection} reports the number of cases satisfying each inclusion criterion individually. Note that the criteria are applied jointly, i.e., included cases must satisfy all four criteria. Out of 1161 cases, a total of 385 cases satisfy all four criteria, however 10 cases got dropped in the final benchmark due to connection timeout errors on the pipeline despite repeated retries.

\subsection{Pipeline Design Decisions}
\label{app:benchmark/pipeline}

The multi-stage architecture of our pipeline is necessitated by the structure of the task itself.
ECtHR judgments are long documents, typically spanning 20 to 60 paragraphs across two structurally distinct sections (reporting and evaluation sections).
The mapping task requires the model to simultaneously attend to both sections.
A naive single-stage prompt that takes the full judgment text as input and directly produces element-level fact mappings and conclusions faces the following difficulties.

First, asking a model to perform atomic fact extraction, conclusion isolation, factual grounding identification, and fact-to-element mapping in a single pass produces atomic facts with inconsistent levels of granularity. The mapping is also incomplete in many cases.
Second, the three-list structure of the benchmark is inherently sequential since List 3 (\textit{decisive facts}) $\subseteq$ List 2 (\textit{essential facts}) $\subseteq$ List 1 (\textit{relevant facts}). These subset dependencies cannot be reliably enforced in a single prompt.
This also makes the intermediate outputs inconsistent with each other.
Third, each stage of the pipeline produces an intermediate output that serves as a structuring constraint for the next stage. The element-level conclusions anchor the factual grounding extraction, ensuring that the facts identified as supporting each element are consistent with the Court's actual reasoning rather than the model's own legal judgment. Similarly, the factual groundings anchor the atomic fact mapping.
Compressing these stages into a single prompt removes these intermediate constraints and substantially degrades output quality, as the model must implicitly perform all operations simultaneously.

\subsection{Pipeline Implementation}
\label{app:benchmark/implementation}

Table~\ref{tab:timings} reports wall-clock timing statistics for each stage of the pipeline, measured across all cases processed. Sample selection was run on all 1,161 candidate cases; the remaining stages were run on the 375 benchmark cases only. All stages use Anthropic's \texttt{Claude Sonnet 4.6} via the official API and were run sequentially without parallelisation.

\begin{table}[h]
\centering
\small
\setlength{\tabcolsep}{4pt}
\caption{Wall-clock timing statistics per pipeline stage. Times are in seconds per case. Total is in minutes. Sample selection was run on 1,161 cases; all other stages on 375 cases.}
\label{tab:timings}
\begin{tabular}{l r rrrrrr}
\toprule
\textbf{Stage} & \textbf{N} & \textbf{Mean} & \textbf{Median} & \textbf{Std} & \textbf{Min} & \textbf{Max} & \textbf{Total (min)} \\
\midrule
Sample selection       & 1161 &   3.97 &   3.93 &   0.89 &   2.02 &   11.58 &   76.8 \\
Pre-processing         &  375 & 102.40 &  38.05 & 147.06 &  10.47 & 1617.25 &  640.0 \\
Atomic fact division   &  375 &  54.96 &  16.09 &  77.85 &   5.11 &  579.38 &  343.5 \\
Conclusion isolation   &  375 &  25.24 &  22.80 &   8.56 &  14.85 &   63.34 &  157.7 \\
Fact grounding         &  375 &  22.14 &  18.67 &  10.47 &  10.20 &   87.51 &  138.4 \\
Fact mapping           &  375 &  42.21 &  34.12 &  24.22 &  17.63 &  194.23 &  263.8 \\
\midrule
End-to-end (pipeline)  &  375 & 246.95 & 134.26 & 255.05 &  69.85 & 2281.12 & 1543.4 \\
\bottomrule
\end{tabular}
\end{table}

The complete pipeline for 375 cases ran in approximately 25.7 hours end-to-end, averaging 4.1 minutes per case with a median of 2.2 minutes. A small number of unusually long judgments drive up the average substantially, compared to the median.

Pre-processing is the most time-consuming stage, accounting for 41.5\% of total pipeline time (640 minutes). It also shows the highest variability, with a mean-to-median ratio of 2.7x and a maximum of 27 minutes for a single case. This reflects the wide variance in document lengths of ECtHR judgments. Atomic fact division is the second most expensive stage (343.5 minutes, 22.3\%), with similarly high variability across cases.

The three subsequent stages are considerably faster and more consistent across cases. Conclusion isolation (157.7 minutes) and fact grounding extraction (138.4 minutes) are the most stable stages. Fact mapping (263.8 minutes) varies to a somewhat greater extent.

Sample selection, which was applied to all 1,161 candidate cases, completed in 76.8 minutes total at approximately 4 seconds per case, making it a negligible overhead in the overall process.
The full pipeline executes within the range of about a day on a single machine.

\paragraph{API Costs.} The entire sample selection stage (for 1161 cases) incurred an API cost of \$34.53. The total API cost of the pipeline after sample selection, for all five stages and 375 cases, was \$120.40.

\paragraph{Temperature and Thinking Mode.} For \textit{all} stages of the pipeline, we disabled the `thinking' mode and used temperature 0.0.

\subsection{Benchmark Statistics}
\label{app:benchmark/stats}

The distribution of active elements per case reflects the doctrinal structure of Art.\ 11.
The majority of cases (84.8\%) involve five active elements.
This corresponds to the standard doctrinal pathway followed by the Court in most cases: through scope, interference, and all three justification elements.
Cases with fewer active elements typically arise when the Court finds scope or interference is not established and concludes the analysis, or when positive obligation is the only route that is legally valid.

We observe that the Court reasons through substantially more facts for \textit{necessity} than any other element on both List~1 (mean 9.45) and List~2 (mean 5.63), except \textit{positive obligation} which involves a distinct doctrinal route.
This is natural, since the Court conducts the proportionality assessment in this element, i.e., weighing the intensity of the interference against the importance of the legitimate aim.
Positive obligation also attracts many facts on both lists (11.54 and 6.75), which reflects the need to establish both the existence of a positive duty and the government's failure to discharge it.
The List 2:List 1 compression ratio is consistent across elements (0.52--0.60), suggesting that the pipeline can systematically reduce the set of \textit{relevant facts} to the \textit{necessary facts} absolutely needed to arrive at the conclusion for an element.

\subsection{Validation by Legal Experts}
\label{app:benchmark/validation}

The expert validation procedure was a thorough and time-consuming exercise. We posted an advertisement for the job on the freelancing platform Upwork~\footnote{\url{https://www.upwork.com}}. Interested applicants were asked to provide a cover letter detailing their understanding of the project, and how they were a good fit for hiring.
One of the authors, who is an esteemed legal scholar with over four decades of experience in European law, vetted all eight applications we received, and we proceeded with the three most promising candidates. As already reported in Section~\ref{sec:benchmark/validation}, the three chosen candidates all hold master’s degrees in law from European institutions with significant experience of ECtHR cases.

The three chosen experts were initially asked to validate five pilot cases, all of which were reviewed by the legal scholar in our team.
This was then used to hash out the details of the annotation guide, specifically with regard to the difference between List~2 (\textit{essential facts}) and List~3 (\textit{decisive facts}), and align the experts in terms of their interpretation of the fact-mapping. For this round, they were paid \$10 per case. Subsequently, they were provided with 15 more cases and the aligned guide, and were paid \$5 per case for the scaled-up exercise. Each case required around 45 minutes, and the whole validation process took close to 2.5 weeks.

The experts were asked to mark: (i) whether the final Art.\ 11 violation output is correct~\ding{51} or wrong~\ding{55}; (ii) whether the conclusion for each element of doctrine is correct~\ding{51} or wrong~\ding{55}; and (iii) for each atomic fact in Lists 2 and 3, whether the mapping is clearly correct~\ding{51}, clearly incorrect~\ding{55}, or borderline~$\sim$.

\begin{table}[t]
  \caption{Expert validation of the pipeline on 20 cases. Left: validation scores (\%), with 95\% case-level bootstrap confidence intervals (1{,}000 resamples) in brackets; overall is the mean across the three experts. Element-conclusion accuracy is averaged over the six elements. Right: inter-annotator agreement among the three experts. Element-conclusion agreement is averaged over 4 elements except \textit{scope} and \textit{interference}, for which some metrics are undefined due to the missing minority class. Lenient precision counts borderline ratings as correct; strict precision counts borderline ratings as incorrect.}
  \label{tab:validation_overall}
  \centering
  \footnotesize
  \setlength{\tabcolsep}{2pt}
  \begin{tabular*}{\textwidth}{@{\extracolsep{\fill}}lcccccccc}
    \toprule
     & \multicolumn{4}{c}{Validation scores (\%)} & \multicolumn{4}{c}{Inter-annotator agreement} \\
    \cmidrule(lr){2-5}\cmidrule(lr){6-9}
    Metric & Expert 1 & Expert 2 & Expert 3 & Overall & \shortstack{Raw\\(\%)} & \shortstack{Fleiss'\\$\kappa$} & \shortstack{Krippendorff's\\$\alpha$} & \shortstack{Gwet's\\AC1} \\
    \midrule
    \shortstack[l]{Violation accuracy} & \shortstack{$100.0$\\[-0.4ex]{\tiny$[100.0,\,100.0]$}} & \shortstack{$100.0$\\[-0.4ex]{\tiny$[100.0,\,100.0]$}} & \shortstack{$100.0$\\[-0.4ex]{\tiny$[100.0,\,100.0]$}} & \shortstack{$100.0$\\[-0.4ex]{\tiny$[100.0,\,100.0]$}} & $100.0$ & $1.000$ & $1.000$ & $1.000$ \\[0.6ex]
    \shortstack[l]{Element-conclusion\\accuracy} & \shortstack{$96.7$\\[-0.4ex]{\tiny$[93.3,\,99.2]$}} & \shortstack{$94.1$\\[-0.4ex]{\tiny$[88.9,\,98.3]$}} & \shortstack{$95.0$\\[-0.4ex]{\tiny$[90.8,\,98.3]$}} & \shortstack{$95.2$\\[-0.4ex]{\tiny$[91.8,\,98.3]$}} & $86.4$ & $0.357$ & $0.499$ & $0.767$ \\[0.6ex]
    \shortstack[l]{List 2 precision\\(lenient)} & \shortstack{$93.0$\\[-0.4ex]{\tiny$[82.6,\,98.7]$}} & \shortstack{$96.5$\\[-0.4ex]{\tiny$[91.5,\,99.1]$}} & \shortstack{$91.3$\\[-0.4ex]{\tiny$[79.7,\,97.1]$}} & \shortstack{$93.6$\\[-0.4ex]{\tiny$[87.2,\,97.2]$}} & $94.2$ & $0.194$ & $0.197$ & $0.936$ \\[0.6ex]
    \shortstack[l]{List 2 precision\\(strict)} & \shortstack{$90.6$\\[-0.4ex]{\tiny$[79.2,\,98.4]$}} & \shortstack{$79.1$\\[-0.4ex]{\tiny$[68.7,\,88.6]$}} & \shortstack{$84.6$\\[-0.4ex]{\tiny$[74.1,\,91.5]$}} & \shortstack{$84.8$\\[-0.4ex]{\tiny$[75.1,\,91.6]$}} & $83.6$ & $0.218$ & $0.220$ & $0.787$ \\[0.6ex]
    \shortstack[l]{List 3 precision\\(lenient)} & \shortstack{$99.5$\\[-0.4ex]{\tiny$[98.6,\,100.0]$}} & \shortstack{$98.2$\\[-0.4ex]{\tiny$[96.3,\,100.0]$}} & \shortstack{$97.3$\\[-0.4ex]{\tiny$[94.7,\,99.8]$}} & \shortstack{$98.3$\\[-0.4ex]{\tiny$[96.9,\,99.6]$}} & $97.3$ & $0.200$ & $0.201$ & $0.972$ \\[0.6ex]
    \shortstack[l]{List 3 precision\\(strict)} & \shortstack{$98.8$\\[-0.4ex]{\tiny$[97.1,\,100.0]$}} & \shortstack{$88.7$\\[-0.4ex]{\tiny$[84.7,\,92.9]$}} & \shortstack{$89.2$\\[-0.4ex]{\tiny$[84.0,\,93.7]$}} & \shortstack{$92.2$\\[-0.4ex]{\tiny$[90.0,\,94.7]$}} & $85.9$ & $0.162$ & $0.163$ & $0.831$ \\
    \bottomrule
  \end{tabular*}
\end{table}

\begin{table}[t]
  \caption{Per-element expert validation (\%). For each expert and for the mean across the three experts (Overall): accuracy of the element conclusion (Acc) and List~2 precision under lenient (Len.) and strict (Str.) rating of borderline facts.}
  \label{tab:validation_elements}
  \centering
  \footnotesize
  \setlength{\tabcolsep}{2.5pt}
  \begin{tabular*}{\textwidth}{@{\extracolsep{\fill}}lcccccccccccc}
    \toprule
     & \multicolumn{3}{c}{Expert 1} & \multicolumn{3}{c}{Expert 2} & \multicolumn{3}{c}{Expert 3} & \multicolumn{3}{c}{Overall} \\
    \cmidrule(lr){2-4}\cmidrule(lr){5-7}\cmidrule(lr){8-10}\cmidrule(lr){11-13}
    Element & Acc & Len. & Str. & Acc & Len. & Str. & Acc & Len. & Str. & Acc & Len. & Str. \\
    \midrule
    Scope & $100.0$ & $98.7$ & $98.7$ & $100.0$ & $96.4$ & $74.3$ & $100.0$ & $93.4$ & $88.4$ & $100.0$ & $96.1$ & $87.1$ \\
    Interference & $100.0$ & $95.6$ & $95.6$ & $100.0$ & $97.5$ & $92.2$ & $100.0$ & $95.0$ & $90.9$ & $100.0$ & $96.0$ & $92.9$ \\
    Prescribed by law & $90.0$ & $100.0$ & $93.1$ & $85.0$ & $98.8$ & $81.8$ & $80.0$ & $92.7$ & $78.6$ & $85.0$ & $97.2$ & $84.5$ \\
    Legitimate aim & $95.0$ & $90.8$ & $83.1$ & $89.5$ & $98.5$ & $85.2$ & $95.0$ & $94.5$ & $87.3$ & $93.2$ & $94.6$ & $85.2$ \\
    Necessity & $95.0$ & $98.3$ & $98.3$ & $100.0$ & $98.8$ & $91.1$ & $100.0$ & $97.2$ & $94.8$ & $98.3$ & $98.1$ & $94.7$ \\
    Positive obligation & $100.0$ & $75.0$ & $75.0$ & $90.0$ & $88.9$ & $50.0$ & $95.0$ & $75.0$ & $67.9$ & $95.0$ & $79.6$ & $64.3$ \\
    \bottomrule
  \end{tabular*}
\end{table}

Table~\ref{tab:validation_overall} reports the validation scores with 95\% case-level bootstrap confidence intervals (1{,}000 resamples).
All experts considered every violation outcome and a large majority of the element conclusions to be correct (avg. accuracy 95.2\%).
For fact-mapping, we compare lenient scoring, which treats borderline ratings as correct, in contrast to strict scoring, which treats them as incorrect.
List~2 precision falls from 93.6\% to 84.8\% under strict scoring, while List~3 precision falls from 98.3\% to 92.2\%.
This difference is largest for Expert~2 and smallest for Expert~1, demonstrating that experts have varying thresholds for what they consider to be borderline.
However, even when all borderline mappings are rejected, experts endorse 84.8\% of essential fact mappings and 92.2\% of decisive fact mappings.

The table also reports agreement between experts.
Raw agreement is $83.6-100\%$, and Gwet's AC1~\citep{gwet2008computing} is $0.767-1.000$.
Fleiss' $\kappa$~\citep{fleiss1971measuring} and Krippendorff's $\alpha$~\citep{krippendorff1970estimating} are much lower for fact-mapping. This reflects the skew towards accepting the proposed mappings, which can produce high raw agreement but low chance-corrected
agreement \citep{feinstein1990high}.
The high raw agreement and AC1 further show that the endorsement of our benchmark is not limited to one expert.

Table~\ref{tab:validation_elements} shows the validation results for each element.
Conclusion accuracy is $100\%$ for \textit{scope} and \textit{interference} but lowest for \textit{prescribed by law} (85.0\%).
List~2 precision is highest for necessity ($98.1\%$ lenient; $94.7\%$ strict) and lowest for positive obligation ($79.6\%$; $64.3\%$).
Even under strict scoring, List~2 precision remains above $84\%$ for five of the six elements, with positive obligation as the main exception.
This supports the quality of fact mapping across most elements.

\clearpage
\newpage

\section{Prompts}
\label{app:prompts}
\subsection{Sample Selection} \label{app:prompts/sample_sel}
{\small\ttfamily
\# SYSTEM PROMPT

You are an expert in European human rights law, specialising in the case law of the European Court of Human Rights (ECtHR). You have deep familiarity with the structure of ECtHR judgments and the legal framework under Article 11 of the Convention, including the standard scrutiny sequence: scope, interference, lawfulness, legitimate aim, necessity, and positive obligations.

---

\# USER PROMPT

You are given the full text of an ECtHR judgment concerning Article 11 of the European Convention on Human Rights. The text has been extracted directly from a PDF and may contain headers, footers, page numbers, and section separators — disregard these and focus solely on the substantive legal content.

Your task is to determine whether this case should be included in a gold-standard dataset for evaluating legal reasoning under Article 11. A case should be included only if all four of the following conditions are met:

1. **Right to assembly** — The Article 11 complaint must concern the right to peaceful assembly, not association rights such as trade union disputes, party registration, or organizational dissolution.

2. **Single coherent merits analysis** — The Article 11 merits reasoning must function as a single unified analysis — one that applies consistently across all applicants and complaints and reaches a common outcome.

3. **Facts treated as settled** — The Article 11 merits reasoning must proceed on facts the Court treats as established. Cases where the Court itself must determine what happened before it can evaluate the law are excluded.

4. **Genuine merits adjudication** — The case must conclude with a standalone Article 11 merits finding, i.e.,  either a violation or no violation of Article 11. Cases that do not reach a substantive Article 11 conclusion, whether due to inadmissibility, procedural disposal, or because the finding is expressed under another article read in conjunction with Article 11, are excluded.

\#\# Output Format

Respond in the following JSON format:

\{\\
  "criteria": \{\\
    "is\_assembly\_case":     true/false,\\
    "is\_coherent\_merits":   true/false,\\
    "facts\_are\_settled":    true/false,\\
    "is\_genuine\_merits":    true/false\\
  \},\\
  "include": true/false,\\
  "justification": "..."\\
\}

`include` should be `true` if and only if all criteria are `true`. The justification should be a single concise sentence explaining the decision.

\#\# Input

\{case\_document\}
}

\subsection{Pre-processing} \label{app:prompts/preproc}

{\small\ttfamily
\# SYSTEM PROMPT

You are a precise document structure parser for European Court of Human Rights judgments. You extract structural information from legal documents and return it as compact JSON. You never output prose, explanations, or markdown — only valid JSON exactly matching the requested schema.

\# USER PROMPT

Parse the following European Court of Human Rights judgment into structured paragraph lists for the Facts section and the Article 11 reasoning section only.

The document is raw text extracted from a PDF. Each page begins with `========`. Use page boundaries to detect and discard contamination — headers, footers, page numbers, footnotes — and to merge paragraph content that flows across page boundaries.

\#\# OUTPUT FORMAT

\{\\
  "facts": [...],\\
  "art11": [...]\\
\}

Each entry:

\{\\
  "para\_id":          int,\\
  "source\_paragraph": int | null,\\
  "section\_path":     [str, ...],\\
  "text":             str\\
\}

`para\_id` is sequential from 1, independent for each list. `source\_paragraph` is the printed ECtHR paragraph number, or null if the content has no printed paragraph number. `section\_path` is the heading hierarchy above the paragraph. `text` is copied verbatim from the document — do not paraphrase, summarise, or generate any text of your own. Every word in `text` must appear in the source document.

\#\# FACTS SECTION

The facts section includes everything under the top-level "THE FACTS" heading (or equivalent) until the paragraph immediately before the next top-level heading ("THE LAW" or equivalent). This includes all subsections: circumstances of the case, relevant domestic law, domestic court decisions, international materials, and anything else under "THE FACTS".

\#\# ART. 11 SECTION

The full section within "THE LAW" where the Court examines the Article 11 complaint on freedom of peaceful assembly, whether standalone or jointly with Article 10.

\#\# PARAGRAPHS AND HEADINGS

Real headings organise the document into sections and appear in `section\_path` only — never as paragraph entries and never in paragraph text. A real heading is one that introduces a named section, followed by either another heading or a numbered paragraph.

Numbered paragraphs (`NN.  `) are the primary paragraph entries. Any unnumbered content that follows a numbered paragraph — continuation prose, quoted text, article texts — is part of that paragraph and included in its text. Content that looks like a heading but is followed directly by unnumbered prose rather than a heading or numbered paragraph is not a real heading: it and the prose beneath it belong to the preceding numbered paragraph.

\#\# DOCUMENT

\{document\}
}

\subsection{Atomic Fact Division} \label{app:prompts/atomic_facts}

{\small\ttfamily
\# SYSTEM PROMPT

You are an expert legal information extractor specialising in European Court of Human Rights judgments. Your task is to decompose structured paragraph text into atomic facts — the smallest units of information that are independently meaningful. You return only valid JSON with no surrounding prose or markdown.

\# USER PROMPT

\#\# OBJECTIVE

You will receive the facts section of an ECtHR judgment as a list of paragraph objects. Each paragraph has been pre-annotated with a para\_id and section\_path. Your task is to decompose each paragraph into atomic facts and carry the para\_id through to each fact unchanged.

An atomic fact is the smallest unit of information that describes a single event, finding, action, or state, and can stand alone without reference to surrounding text.

---

\#\# OUTPUT FORMAT

Return a single JSON array. Each element represents one atomic fact:

\{\\
  "fact\_id": "F001",      // sequential across entire output, zero-padded to 3 digits\\
  "text\_resolved": "...", // self-contained fact text, see resolution rules below\\
  "para\_id": 5            // copied from input paragraph, unchanged\\
\}

---

\#\# TEXT RESOLUTION

Resolve every pronoun, abbreviation, partial reference, shortened law name, article number, role label, or court name to its most specific unambiguous form, so that each fact is fully self-contained. If an abbreviation is defined in the same paragraph, expand it in every fact that uses it. Strip the paragraph number prefix from all fact text.

Exception: preserve the following ECtHR terms-of-art exactly as written: "the applicant", "the Government" (meaning the respondent state), "the Convention", "the Court" when referring to the ECtHR itself.

---

\#\# ATOMIZATION

Decompose each paragraph into facts that are as granular as needed for independent legal analysis — distinct events, findings, actions, and states should each become their own fact. A direct speech act is a single fact unless its content spans distinct events, in which case split and prepend the speaker attribution to each resulting fact.

Every piece of information in every paragraph must appear in at least one fact.

---

\#\# INPUT

\{fact\_paras\}
}

\subsection{Isolation of Conclusions} \label{app:prompts/conclusion}

{\small\ttfamily
\# SYSTEM PROMPT

You are a legal expert specializing in European human rights law, with deep knowledge of Article 11 ECHR jurisprudence on freedom of assembly. Your task is to analyze ECtHR judgments to build a structured understanding of how courts reason about Article 11 cases. You will work through a judgment in stages, building on your prior analysis at each step. Your analysis should be thorough, precise, and faithful to what the Court actually said.

---

\# USER PROMPT

Read the following reasoning section from a decision by the European Court of Human Rights, and then analyze the questions below.

**Reasoning text:**
\{reasoning\_text\}

**1.** Does the Court come to the conclusion that Article 11 ECHR has been violated? If yes, from which paragraph(s) do you take this information? Quote the paragraph number and the wording.

**2.** Please also give us the decisions of the Court regarding the elements of the canonical doctrine of Article 11 ECHR. In the Court's established jurisprudence, Article 11 ECHR requires the following tests:

- **(a) Scope:** i) has there been an assembly? ii) did the applicant (try to) participate in the assembly? iii) was the assembly peaceful?

- **(b) Interference:** did some governmental act or omission interfere with the applicant's freedom of assembly?

- **(c) Prescribed by law:** i) is the intervention sufficiently foreseeable, given the wording of the law? ii) is the provision that empowers government sufficiently accessible? iii) does the provision come with sufficient safeguards against abuse?

- **(d) Legitimate aim:** did it serve one of the legitimate aims listed in Article 11 ECHR, namely: national security, public safety, the prevention of disorder, prevention of crime, the protection of health, the protection of morals, protection of the rights and freedoms of others?

- **(e) Necessity in a democratic society:** did government adequately balance the intensity of the intervention with the weight of the legitimate aim?

- **(f) Positive obligation:** has the Court spelled out a positive obligation for government? If so, has it come to the conclusion that the defendant government has violated this obligation?

Which of these canonical elements of the doctrine of Article 11 ECHR are explicitly discussed in the decision? Are elements of doctrine implied (e.g. by the fact that the Court explicitly discusses an element that logically requires an earlier element to be given)? Again, quote the paragraph number and the wording for each canonical element that the Court has explicitly discussed.

---

\#\# Output

Return a JSON object with two keys. In `analysis`, answer the questions above as fully as you would in a standalone response — for the final violation conclusion as well as each element and sub-element, quote the specific passages that ground each finding, treating each sub-element as a separate question requiring its own answer. In `results`, extract the structured conclusions as shown below.

\{\\
  "analysis": "Your full analysis here...",\\
  "results": \{\\
    "violated": "true/false/null",\\
    "elements": \{\\
      "scope":                         \{"conclusion": "yes/no/null", "discussed": "explicit/implied/not\_discussed"\},\\
      "scope\_assembly\_exists":         \{"conclusion": "yes/no/null", "discussed": "explicit/implied/not\_discussed"\},\\
      "scope\_participation":           \{"conclusion": "yes/no/null", "discussed": "explicit/implied/not\_discussed"\},\\
      "scope\_peaceful":                \{"conclusion": "yes/no/null", "discussed": "explicit/implied/not\_discussed"\},\\
      "interference":                  \{"conclusion": "yes/no/null", "discussed": "explicit/implied/not\_discussed"\},\\
      "prescribed\_by\_law":             \{"conclusion": "yes/no/null", "discussed": "explicit/implied/not\_discussed"\},\\
      "prescribed\_by\_law\_foreseeable": \{"conclusion": "yes/no/null", "discussed": "explicit/implied/not\_discussed"\},\\
      "prescribed\_by\_law\_accessible":  \{"conclusion": "yes/no/null", "discussed": "explicit/implied/not\_discussed"\},\\
      "prescribed\_by\_law\_safeguards":  \{"conclusion": "yes/no/null", "discussed": "explicit/implied/not\_discussed"\},\\
      "legitimate\_aim":                \{"conclusion": "yes/no/null", "discussed": "explicit/implied/not\_discussed", "aims\_found": []\},\\
      "necessity":                     \{"conclusion": "yes/no/null", "discussed": "explicit/implied/not\_discussed"\},\\
      "positive\_obligation":           \{"conclusion": "yes/no/null", "discussed": "explicit/implied/not\_discussed"\}\\
    \}\\
  \}\\
\}\\

- `analysis`: Remember to escape any quotes with \textbackslash

- `violated`: `"true"` if Article 11 has been violated, `"false"` if not, `"null"` if unclear

- `conclusion`: `"yes"` if the Court answers the element's question affirmatively, `"no"` if negatively, `"null"` if not discussed or unresolved

- `discussed`: `"explicit"` if directly addressed, `"implied"` if follows from other elements, `"not\_discussed"` if absent entirely

- `aims\_found`: list of applicable aims from: `"national\_security"`, `"public\_safety"`, `"prevention\_of\_disorder"`, `"prevention\_of\_crime"`, `"protection\_of\_health"`, `"protection\_of\_morals"`, `"rights\_of\_others"` — empty list `[]` if legitimate aim not discussed
}

\subsection{Factual Grounding} \label{app:prompts/fact_grounding}

{\small\ttfamily
\# SYSTEM PROMPT

You are a legal expert specializing in European human rights law, with deep knowledge of Article 11 ECHR jurisprudence on freedom of assembly. Your task is to analyze ECtHR judgments to build a structured understanding of how courts reason about Article 11 cases. You will work through a judgment in stages, building on your prior analysis at each step. Your analysis should be thorough, precise, and faithful to what the Court actually said.

---

\# USER PROMPT

On the elements of legal doctrine that you have found to be either implied or explicitly discussed: on which facts is the court's assessment grounded? Please quote these facts verbatim from the reasoning text. Where the Court draws on party submissions, quote them only if the Court explicitly adopted or accepted them as established, not merely because a party asserted them.

Do not force a mapping — if you cannot identify any factual grounding for a given element, just note this.
}

\subsection{Fact Mapping} \label{app:prompts/fact_mapping}

{\small\ttfamily
\# SYSTEM PROMPT

You are a legal expert specializing in European human rights law, with deep knowledge of Article 11 ECHR jurisprudence on freedom of assembly. Your task is to analyze ECtHR judgments to build a structured understanding of how courts reason about Article 11 cases. You will work through a judgment in stages, building on your prior analysis at each step. Your analysis should be thorough, precise, and faithful to what the Court actually said.

---

\# USER PROMPT

In the previous steps, we gave you only the reasoning section of the judgment and asked you to identify the Court's conclusions and the factual grounding for each element. We now ask you to relate your analysis to the earlier part of the judgment in which the Court reports the circumstances of the case — the submissions of the parties, the applicable national law, the decisions of national authorities and courts, and any other material the Court considered relevant. We have formalized this material into a set of atomic facts, provided below.

**Atomic facts:**
\{atomic\_facts\}

We ask you to write three lists.

**List 1:** For every individual fact that you listed as "factual grounding" (in the legal reasoning part of the decision), find all possible "atomic facts" that the Court was pointing to. Make sure that these atomic facts actually correspond to the exact factual grounding quote.

**List 2:** In the next step, focus on each element of law that you have found is discussed by the Court: which "atomic facts" are strictly necessary to support this part of the reasoning of the court, grounded in the facts mentioned by the Court. Where multiple atomic facts establish the same point, include only the most critical or representative facts.

**List 3:** In the final step, focus on the final conclusion (Art. 11 ECHR has been violated, or not): which "atomic facts" are strictly necessary for the Court to come to this conclusion, assuming that the Court will apply the doctrinal structure of Art. 11 ECHR. Focus on what the case actually turned on.

This procedure implies that any fact in List 3 must feature in List 2, and any fact listed in List 2 must feature in List 1. Put differently, what we want you to do is to narrow down on the atomic facts that are absolutely necessary to back up the legal reasoning of the Court, as grounded in its assessment of the facts (List 2). By contrast, in List 3, we do not want you to replicate the Court's reasoning. Rather in the light of the facts of the case: which atomic facts are essential for reaching the conclusion (Art. 11 violated or not) that the Court reached.

---

\#\# Output Format

Return a JSON object with two keys. In `analysis`, write the three lists in full as described above. In `results`, extract the structured conclusions.

\{\\
  "analysis": "Your three lists here...",\\
  "results": \{\\
    "list1": \{\\
      "scope":                ["FXXX", "FXXX"],\\
      "interference":         ["FXXX"],\\
      "prescribed\_by\_law":    ["FXXX"],\\
      "legitimate\_aim":       ["FXXX"],\\
      "necessity":            ["FXXX"],\\
      "positive\_obligation":  ["FXXX"]\\
    \},\\
    "list2": \{\\
      "scope":                ["FXXX"],\\
      "interference":         ["FXXX"],\\
      "prescribed\_by\_law":    ["FXXX"],\\
      "legitimate\_aim":       ["FXXX"],\\
      "necessity":            ["FXXX"],\\
      "positive\_obligation":  ["FXXX"]\\
    \},\\
    "list3": ["FXXX", "FXXX"]\\
  \}\\
\}

- `analysis`: Remember to escape any quotes with \textbackslash

- `list1` and `list2` are keyed by the six canonical elements; include only fact IDs that appear in the atomic facts provided

- `list3` is a flat list of fact IDs representing the irreducible factual core for the final conclusion

- Any fact ID in `list2` must appear in `list1` for the same element; any fact ID in `list3` must appear in `list2` for at least one element

- An element's list may be empty `[]` if no atomic facts correspond to it
}

\subsection{Full Reasoning (FR) Downstream Evaluation Setup} \label{app:prompts/ds_vanilla}
{\small\ttfamily
\# SYSTEM PROMPT

You are a legal expert specializing in European human rights law, with deep knowledge of Article 11 ECHR jurisprudence on freedom of assembly. Your task is to analyze the facts of an ECtHR case and reason about whether Article 11 has been violated.

---

\# USER PROMPT

You are analyzing the facts of an Article 11 case from the European Court of Human Rights. The facts of a case usually contain the circumstances of the case, including the submissions by the applicant, the defendant government and (potentially) other actors, the applicable national law, the decisions taken by national authorities and courts, and any other material the court deems important to understand and decide the case.

The Court's established jurisprudence on Article 11 ECHR requires the following canonical elements to be assessed:

- **(a) Scope:** i) Has there been an assembly? ii) Did the applicant (try to) participate in the assembly? iii) Was the assembly peaceful?

- **(b) Interference:** Did some governmental act or omission interfere with the applicant's freedom of assembly?

- **(c) Prescribed by law:** i) Is the intervention sufficiently foreseeable given the wording of the law? ii) Is the provision that empowers government sufficiently accessible? iii) Does the provision come with sufficient safeguards against abuse?

- **(d) Legitimate aim:** Did the interference serve one of the legitimate aims listed in Article 11(2) ECHR, namely: national security, public safety, prevention of disorder, prevention of crime, protection of health, protection of morals, or protection of the rights and freedoms of others?

- **(e) Necessity in a democratic society:** Did the government adequately balance the intensity of the intervention with the weight of the legitimate aim?

- **(f) Positive obligation:** Has the Court spelled out a positive obligation for the government? If so, has it concluded that the defendant government violated this obligation?

These elements follow a conditional doctrinal chain. Scope and interference are threshold questions: if either is not established, no violation can arise. If interference is established, the Court examines whether it was prescribed by law, pursued a legitimate aim, and was necessary in a democratic society. A "no" conclusion on any of these three justification elements means the interference was unlawful and Article 11 has been violated. Positive obligation is a parallel reasoning route applicable when the government failed to take positive steps to facilitate the right rather than actively interfering with it.

Now, your task is to identify the facts needed to reason about the case. To facilitate this task, we have already formalized this material into a set of "atomic facts", provided below.

**Atomic facts:**
\{atomic\_facts\}

For each canonical element, identify which atomic facts are needed to answer its question, and provide a conclusion. Then provide a final conclusion on whether Article 11 has been violated.

---

\#\# Output Format

Return a single JSON object. Do not include any text outside the JSON.

\{
  "elements": \{\\
    "scope": \{\\
      "conclusion": "yes/no",\\
      "facts": ["FXXX", "FXXX"]\\
    \},\\
    "interference": \{\\
      "conclusion": "yes/no",\\
      "facts": ["FXXX", "FXXX"]\\
    \},\\
    "prescribed\_by\_law": \{\\
      "conclusion": "yes/no",\\
      "facts": ["FXXX"]\\
    \},\\
    "legitimate\_aim": \{\\
      "conclusion": "yes/no",\\
      "facts": ["FXXX"]\\
    \},
    "necessity": \{\\
      "conclusion": "yes/no",\\
      "facts": ["FXXX", "FXXX"]\\
    \},\\
    "positive\_obligation": \{\\
      "conclusion": "yes/no",\\
      "facts": ["FXXX"]\\
    \}\\
  \},\\
  "violated": true/false\\
\}

- For each element, `conclusion` should be `yes` if the answer to that element's question is affirmative, and `no` if it is negative.

- `violated` is a boolean — `true` if Article 11 has been violated, `false` otherwise

- Only include fact IDs that appear in the atomic facts provided above

- Return an empty list `[]` for `facts` if no atomic facts are relevant to an element
}

\subsection{Direct Violation (DV) Downstream Evaluation Setup} \label{app:prompts/ds_dv}

{\small\ttfamily
\# SYSTEM PROMPT

You are a legal expert specializing in European human rights law, with deep knowledge of Article 11 ECHR jurisprudence on freedom of assembly. Your task is to analyze the facts of an ECtHR case and reason about whether Article 11 has been violated.

---

\# USER PROMPT

You are analyzing the facts of an Article 11 case from the European Court of Human Rights. The facts of a case usually contain the circumstances of the case, including the submissions by the applicant, the defendant government and (potentially) other actors, the applicable national law, the decisions taken by national authorities and courts, and any other material the court deems important to understand and decide the case.

Now, your task is to reason about the case. To facilitate this task, we have already formalized this material into a set of "atomic facts", provided below.

**Atomic facts:**
\{atomic\_facts\}

Based on the atomic facts provided, determine whether Article 11 has been violated.

---

\#\# Output Format

Return a single JSON object. Do not include any text outside the JSON.

\{\\
  "violated": true/false\\
\}

- `violated` is a boolean — `true` if Article 11 has been violated, `false` otherwise
}

\subsection{Direct Conclusion (DC) Downstream Evaluation Setup} \label{app:prompts/ds_dc}

{\small\ttfamily
\# SYSTEM PROMPT

You are a legal expert specializing in European human rights law, with deep knowledge of Article 11 ECHR jurisprudence on freedom of assembly. Your task is to analyze the facts of an ECtHR case and reason about whether Article 11 has been violated.

---

\# USER PROMPT

You are analyzing the facts of an Article 11 case from the European Court of Human Rights. The facts of a case usually contain the circumstances of the case, including the submissions by the applicant, the defendant government and (potentially) other actors, the applicable national law, the decisions taken by national authorities and courts, and any other material the court deems important to understand and decide the case.

The Court's established jurisprudence on Article 11 ECHR requires the following canonical elements to be assessed:

- **(a) Scope:** i) Has there been an assembly? ii) Did the applicant (try to) participate in the assembly? iii) Was the assembly peaceful?

- **(b) Interference:** Did some governmental act or omission interfere with the applicant's freedom of assembly?

- **(c) Prescribed by law:** i) Is the intervention sufficiently foreseeable given the wording of the law? ii) Is the provision that empowers government sufficiently accessible? iii) Does the provision come with sufficient safeguards against abuse?

- **(d) Legitimate aim:** Did the interference serve one of the legitimate aims listed in Article 11(2) ECHR, namely: national security, public safety, prevention of disorder, prevention of crime, protection of health, protection of morals, or protection of the rights and freedoms of others?

- **(e) Necessity in a democratic society:** Did the government adequately balance the intensity of the intervention with the weight of the legitimate aim?

- **(f) Positive obligation:** Has the Court spelled out a positive obligation for the government? If so, has it concluded that the defendant government violated this obligation?

These elements follow a conditional doctrinal chain. Scope and interference are threshold questions: if either is not established, no violation can arise. If interference is established, the Court examines whether it was prescribed by law, pursued a legitimate aim, and was necessary in a democratic society. A "no" conclusion on any of these three justification elements means the interference was unlawful and Article 11 has been violated. Positive obligation is a parallel reasoning route applicable when the government failed to take positive steps to facilitate the right rather than actively interfering with it.

Now, your task is to reason about the case. To facilitate this task, we have already formalized this material into a set of "atomic facts", provided below.

**Atomic facts:**
\{atomic\_facts\}

For each canonical element, provide a conclusion. Then provide a final conclusion on whether Article 11 has been violated.

---

\#\# Output Format

Return a single JSON object. Do not include any text outside the JSON.

\{
  "elements": \{\\
    "scope": \{\\
      "conclusion": "yes/no",\\
    \},\\
    "interference": \{\\
      "conclusion": "yes/no",\\
    \},\\
    "prescribed\_by\_law": \{\\
      "conclusion": "yes/no",\\
    \},\\
    "legitimate\_aim": \{\\
      "conclusion": "yes/no",\\
    \},
    "necessity": \{\\
      "conclusion": "yes/no",\\
    \},\\
    "positive\_obligation": \{\\
      "conclusion": "yes/no",\\
    \}\\
  \},\\
  "violated": true/false\\
\}

- For each element, `conclusion` should be `yes` if the answer to that element's question is affirmative, and `no` if it is negative.

- `violated` is a boolean — `true` if Article 11 has been violated, `false` otherwise
}

\subsection{Minimal Prompt Scaffolding} \label{app:prompts/ds_minimal}

{\small\ttfamily
\# SYSTEM PROMPT

You are a legal expert specializing in European human rights law, with deep knowledge of Article 11 ECHR jurisprudence on freedom of assembly. Your task is to analyze the facts of an ECtHR case and reason about whether Article 11 has been violated.

---

\# USER PROMPT

You are analyzing the facts of an Article 11 case from the European Court of Human Rights. The facts of a case usually contain the circumstances of the case, including the submissions by the applicant, the defendant government and (potentially) other actors, the applicable national law, the decisions taken by national authorities and courts, and any other material the court deems important to understand and decide the case.

The Court's established jurisprudence on Article 11 ECHR requires the following canonical elements to be assessed:

- **(a) Scope:**

- **(b) Interference:**

- **(c) Prescribed by law:**

- **(d) Legitimate aim:**

- **(e) Necessity in a democratic society:**

- **(f) Positive obligation:**

Now, your task is to identify the facts needed to reason about the case. To facilitate this task, we have already formalized this material into a set of "atomic facts", provided below.

**Atomic facts:**
\{atomic\_facts\}

For each canonical element, identify which atomic facts are needed to answer its question, and provide a conclusion. Then provide a final conclusion on whether Article 11 has been violated.

---

\#\# Output Format

Return a single JSON object. Do not include any text outside the JSON.

\{
  "elements": \{\\
    "scope": \{\\
      "conclusion": "yes/no",\\
      "facts": ["FXXX", "FXXX"]\\
    \},\\
    "interference": \{\\
      "conclusion": "yes/no",\\
      "facts": ["FXXX", "FXXX"]\\
    \},\\
    "prescribed\_by\_law": \{\\
      "conclusion": "yes/no",\\
      "facts": ["FXXX"]\\
    \},\\
    "legitimate\_aim": \{\\
      "conclusion": "yes/no",\\
      "facts": ["FXXX"]\\
    \},
    "necessity": \{\\
      "conclusion": "yes/no",\\
      "facts": ["FXXX", "FXXX"]\\
    \},\\
    "positive\_obligation": \{\\
      "conclusion": "yes/no",\\
      "facts": ["FXXX"]\\
    \}\\
  \},\\
  "violated": true/false\\
\}

- For each element, `conclusion` should be `yes` if the answer to that element's question is affirmative, and `no` if it is negative.

- `violated` is a boolean — `true` if Article 11 has been violated, `false` otherwise

- Only include fact IDs that appear in the atomic facts provided above

- Return an empty list `[]` for `facts` if no atomic facts are relevant to an element
}

\subsection{Partial Prompt Scaffolding} \label{app:prompts/ds_partial}

{\small\ttfamily
\# SYSTEM PROMPT

You are a legal expert specializing in European human rights law, with deep knowledge of Article 11 ECHR jurisprudence on freedom of assembly. Your task is to analyze the facts of an ECtHR case and reason about whether Article 11 has been violated.

---

\# USER PROMPT

You are analyzing the facts of an Article 11 case from the European Court of Human Rights. The facts of a case usually contain the circumstances of the case, including the submissions by the applicant, the defendant government and (potentially) other actors, the applicable national law, the decisions taken by national authorities and courts, and any other material the court deems important to understand and decide the case.

The Court's established jurisprudence on Article 11 ECHR requires the following canonical elements to be assessed:

- **(a) Scope:** i) Has there been an assembly? ii) Did the applicant (try to) participate in the assembly? iii) Was the assembly peaceful?

- **(b) Interference:** Did some governmental act or omission interfere with the applicant's freedom of assembly?

- **(c) Prescribed by law:** i) Is the intervention sufficiently foreseeable given the wording of the law? ii) Is the provision that empowers government sufficiently accessible? iii) Does the provision come with sufficient safeguards against abuse?

- **(d) Legitimate aim:** Did the interference serve one of the legitimate aims listed in Article 11(2) ECHR, namely: national security, public safety, prevention of disorder, prevention of crime, protection of health, protection of morals, or protection of the rights and freedoms of others?

- **(e) Necessity in a democratic society:** Did the government adequately balance the intensity of the intervention with the weight of the legitimate aim?

- **(f) Positive obligation:** Has the Court spelled out a positive obligation for the government? If so, has it concluded that the defendant government violated this obligation?

Now, your task is to identify the facts needed to reason about the case. To facilitate this task, we have already formalized this material into a set of "atomic facts", provided below.

**Atomic facts:**
\{atomic\_facts\}

For each canonical element, identify which atomic facts are needed to answer its question, and provide a conclusion. Then provide a final conclusion on whether Article 11 has been violated.

---

\#\# Output Format

Return a single JSON object. Do not include any text outside the JSON.

\{
  "elements": \{\\
    "scope": \{\\
      "conclusion": "yes/no",\\
      "facts": ["FXXX", "FXXX"]\\
    \},\\
    "interference": \{\\
      "conclusion": "yes/no",\\
      "facts": ["FXXX", "FXXX"]\\
    \},\\
    "prescribed\_by\_law": \{\\
      "conclusion": "yes/no",\\
      "facts": ["FXXX"]\\
    \},\\
    "legitimate\_aim": \{\\
      "conclusion": "yes/no",\\
      "facts": ["FXXX"]\\
    \},
    "necessity": \{\\
      "conclusion": "yes/no",\\
      "facts": ["FXXX", "FXXX"]\\
    \},\\
    "positive\_obligation": \{\\
      "conclusion": "yes/no",\\
      "facts": ["FXXX"]\\
    \}\\
  \},\\
  "violated": true/false\\
\}

- For each element, `conclusion` should be `yes` if the answer to that element's question is affirmative, and `no` if it is negative.

- `violated` is a boolean — `true` if Article 11 has been violated, `false` otherwise

- Only include fact IDs that appear in the atomic facts provided above

- Return an empty list `[]` for `facts` if no atomic facts are relevant to an element
}

\subsection{Staging for Downstream Evaluation} \label{app:prompts/ds_staging}

\paragraph{Fact Mapping.} This is the first stage of setup, where we just ask the LLM to map atomic facts to the legal elements.

{\small\ttfamily
\# SYSTEM PROMPT

You are a legal expert specializing in European human rights law, with deep knowledge of Article 11 ECHR jurisprudence on freedom of assembly. Your task is to analyze the facts of an ECtHR case and reason about whether Article 11 has been violated.

---

\# USER PROMPT

You are analyzing the facts of an Article 11 case from the European Court of Human Rights. The facts of a case usually contain the circumstances of the case, including the submissions by the applicant, the defendant government and (potentially) other actors, the applicable national law, the decisions taken by national authorities and courts, and any other material the court deems important to understand and decide the case.

The Court's established jurisprudence on Article 11 ECHR requires the following canonical elements to be assessed:

- **(a) Scope:** i) Has there been an assembly? ii) Did the applicant (try to) participate in the assembly? iii) Was the assembly peaceful?

- **(b) Interference:** Did some governmental act or omission interfere with the applicant's freedom of assembly?

- **(c) Prescribed by law:** i) Is the intervention sufficiently foreseeable given the wording of the law? ii) Is the provision that empowers government sufficiently accessible? iii) Does the provision come with sufficient safeguards against abuse?

- **(d) Legitimate aim:** Did the interference serve one of the legitimate aims listed in Article 11(2) ECHR, namely: national security, public safety, prevention of disorder, prevention of crime, protection of health, protection of morals, or protection of the rights and freedoms of others?

- **(e) Necessity in a democratic society:** Did the government adequately balance the intensity of the intervention with the weight of the legitimate aim?

- **(f) Positive obligation:** Has the Court spelled out a positive obligation for the government? If so, has it concluded that the defendant government violated this obligation?

Now, your task is to identify the facts needed to reason about each element. To facilitate this task, we have already formalized this material into a set of "atomic facts", provided below.

**Atomic facts:**
\{atomic\_facts\}

For each canonical element, identify which atomic facts are needed to reason about that element.

---

\#\# Output Format

Return a single JSON object. Do not include any text outside the JSON.

\{
  "elements": \{\\
    "scope": \{\\
      "facts": ["FXXX", "FXXX"]\\
    \},\\
    "interference": \{\\
      "facts": ["FXXX", "FXXX"]\\
    \},\\
    "prescribed\_by\_law": \{\\
      "facts": ["FXXX"]\\
    \},\\
    "legitimate\_aim": \{\\
      "facts": ["FXXX"]\\
    \},
    "necessity": \{\\
      "facts": ["FXXX", "FXXX"]\\
    \},\\
    "positive\_obligation": \{\\
      "facts": ["FXXX"]\\
    \}\\
  \}\\
\}

- Only include fact IDs that appear in the atomic facts provided above

- Return an empty list `[]` for `facts` if no atomic facts are relevant to an element
}

\paragraph{Conclusions and Violation.} This is the second stage, where we ask the LLM to derive the conclusions for each element and the final violation decision based on the facts it mapped in the prior stage.

{\small\ttfamily
\# SYSTEM PROMPT

You are a legal expert specializing in European human rights law, with deep knowledge of Article 11 ECHR jurisprudence on freedom of assembly. Your task is to analyze the facts of an ECtHR case and reason about whether Article 11 has been violated.

---

\# USER PROMPT

In the previous step, you identified the atomic facts relevant to each element of the Article 11 analysis. Now, using only the facts you mapped to each element above, reason about whether each element is satisfied and provide a conclusion for each. Then give a final conclusion on whether Article 11 has been violated.

These elements follow a conditional doctrinal chain. Scope and interference are threshold questions: if either is not established, no violation can arise. If interference is established, the Court examines whether it was prescribed by law, pursued a legitimate aim, and was necessary in a democratic society. A "no" conclusion on any of these three justification elements means the interference was unlawful and Article 11 has been violated. Positive obligation is a parallel reasoning route applicable when the government failed to take positive steps to facilitate the right rather than actively interfering with it.

---

\#\# Output Format

Return a single JSON object. Do not include any text outside the JSON.

\{
  "elements": \{\\
    "scope": \{\\
      "conclusion": "yes/no",\\
    \},\\
    "interference": \{\\
      "conclusion": "yes/no",\\
    \},\\
    "prescribed\_by\_law": \{\\
      "conclusion": "yes/no",\\
    \},\\
    "legitimate\_aim": \{\\
      "conclusion": "yes/no",\\
    \},
    "necessity": \{\\
      "conclusion": "yes/no",\\
    \},\\
    "positive\_obligation": \{\\
      "conclusion": "yes/no",\\
    \}\\
  \},\\
  "violated": true/false\\
\}

- For each element, `conclusion` should be `yes` if the answer to that element's question is affirmative, and `no` if it is negative.

- `violated` is a boolean — `true` if Article 11 has been violated, `false` otherwise
}

\subsection{Few-shot Prompting} \label{app:prompts/ds_fewshot}

{\small\ttfamily
\# SYSTEM PROMPT

You are a legal expert specializing in European human rights law, with deep knowledge of Article 11 ECHR jurisprudence on freedom of assembly. Your task is to analyze the facts of an ECtHR case and reason about whether Article 11 has been violated.

---

\# USER PROMPT

Below are 5 worked examples of this exact task. Each example shows a set of atomic facts from a different Article 11 case, followed by the correct answer as a JSON object in the required output format. Study them, then solve the new case that follows.

=== EXAMPLE 1 ===

Atomic facts:

F001: \{text of atomic fact\}

F002: \{text of atomic fact\}

...

Answer:

\{\\
  "elements": \{\\
    "scope": \{\\
      "conclusion": "yes/no",\\
      "facts": ["FXXX", "FXXX"]\\
    \},\\
    "interference": \{\\
      "conclusion": "yes/no",\\
      "facts": ["FXXX", "FXXX"]\\
    \},\\
    "prescribed\_by\_law": \{\\
      "conclusion": "yes/no",\\
      "facts": ["FXXX"]\\
    \},\\
    "legitimate\_aim": \{\\
      "conclusion": "yes/no",\\
      "facts": ["FXXX"]\\
    \},\\
    "necessity": \{\\
      "conclusion": "yes/no",\\
      "facts": ["FXXX", "FXXX"]\\
    \},\\
    "positive\_obligation": \{\\
      "conclusion": "yes/no",\\
      "facts": ["FXXX"]\\
    \}\\
  \},\\
  "violated": true/false\\
\}

=== EXAMPLE 2 ===

[Examples 2 to 5 follow in the same format.]

=== END OF EXAMPLES ===

Now solve the following new case, producing an answer in the same JSON format as the examples above.

You are analyzing the facts of an Article 11 case from the European Court of Human Rights. The facts of a case usually contain the circumstances of the case, including the submissions by the applicant, the defendant government and (potentially) other actors, the applicable national law, the decisions taken by national authorities and courts, and any other material the court deems important to understand and decide the case.

The Court's established jurisprudence on Article 11 ECHR requires the following canonical elements to be assessed:

- **(a) Scope:** i) Has there been an assembly? ii) Did the applicant (try to) participate in the assembly? iii) Was the assembly peaceful?

- **(b) Interference:** Did some governmental act or omission interfere with the applicant's freedom of assembly?

- **(c) Prescribed by law:** i) Is the intervention sufficiently foreseeable given the wording of the law? ii) Is the provision that empowers government sufficiently accessible? iii) Does the provision come with sufficient safeguards against abuse?

- **(d) Legitimate aim:** Did the interference serve one of the legitimate aims listed in Article 11(2) ECHR, namely: national security, public safety, prevention of disorder, prevention of crime, protection of health, protection of morals, or protection of the rights and freedoms of others?

- **(e) Necessity in a democratic society:** Did the government adequately balance the intensity of the intervention with the weight of the legitimate aim?

- **(f) Positive obligation:** Has the Court spelled out a positive obligation for the government? If so, has it concluded that the defendant government violated this obligation?

These elements follow a conditional doctrinal chain. Scope and interference are threshold questions: if either is not established, no violation can arise. If interference is established, the Court examines whether it was prescribed by law, pursued a legitimate aim, and was necessary in a democratic society. A "no" conclusion on any of these three justification elements means the interference was unlawful and Article 11 has been violated. Positive obligation is a parallel reasoning route applicable when the government failed to take positive steps to facilitate the right rather than actively interfering with it.

Now, your task is to identify the facts needed to reason about the case. To facilitate this task, we have already formalized this material into a set of "atomic facts", provided below.

**Atomic facts:**
\{atomic\_facts\}

For each canonical element, identify which atomic facts are needed to answer its question, and provide a conclusion. Then provide a final conclusion on whether Article 11 has been violated.

---

\#\# Output Format

Return a single JSON object. Do not include any text outside the JSON.

\{
  "elements": \{\\
    "scope": \{\\
      "conclusion": "yes/no",\\
      "facts": ["FXXX", "FXXX"]\\
    \},\\
    "interference": \{\\
      "conclusion": "yes/no",\\
      "facts": ["FXXX", "FXXX"]\\
    \},\\
    "prescribed\_by\_law": \{\\
      "conclusion": "yes/no",\\
      "facts": ["FXXX"]\\
    \},\\
    "legitimate\_aim": \{\\
      "conclusion": "yes/no",\\
      "facts": ["FXXX"]\\
    \},
    "necessity": \{\\
      "conclusion": "yes/no",\\
      "facts": ["FXXX", "FXXX"]\\
    \},\\
    "positive\_obligation": \{\\
      "conclusion": "yes/no",\\
      "facts": ["FXXX"]\\
    \}\\
  \},\\
  "violated": true/false\\
\}

- For each element, `conclusion` should be `yes` if the answer to that element's question is affirmative, and `no` if it is negative.

- `violated` is a boolean — `true` if Article 11 has been violated, `false` otherwise

- Only include fact IDs that appear in the atomic facts provided above

- Return an empty list `[]` for `facts` if no atomic facts are relevant to an element
}

\subsection{Article 10 Experiments} \label{app:prompts/art_10}

We construct the Article 10 (\textit{freedom of expression}) extension with the same pipeline as for Article 11 and adapt only what is article-specific. The six top-level elements (\textit{scope}, \textit{interference}, \textit{prescribed by law}, \textit{legitimate aim}, \textit{necessity}, \textit{positive obligation}), the doctrinal logic chain between them, and the three fact lists are unchanged.
We provide below, for each stage, what changed relative to Article 11 and the new prompt
text; where only the article wording in the system prompt differs, we give the new system prompt and refer to the
corresponding Article 11 user prompt.

\paragraph{Sample selection.} The article references change to Article 10, and the assembly-specific criterion
(peaceful assembly rather than association) is dropped, leaving three criteria.

{\small\ttfamily
\# SYSTEM PROMPT

You are an expert in European human rights law, specialising in the case law of the European Court of Human Rights (ECtHR). You have deep familiarity with the structure of ECtHR judgments and the legal framework under Article 10 of the Convention, including the standard scrutiny sequence: scope, interference, lawfulness, legitimate aim, necessity, and positive obligations.

---

\# USER PROMPT

You are given the full text of an ECtHR judgment concerning Article 10 of the European Convention on Human Rights. The text has been extracted directly from a PDF and may contain headers, footers, page numbers, and section separators — disregard these and focus solely on the substantive legal content.

Your task is to determine whether this case should be included in a gold-standard dataset for evaluating legal reasoning under Article 10. A case should be included only if all three of the following conditions are met:

1. **Single coherent merits analysis** — The Article 10 merits reasoning must function as a single unified analysis — one that applies consistently across all applicants and complaints and reaches a common outcome.

2. **Facts treated as settled** — The Article 10 merits reasoning must proceed on facts the Court treats as established. Cases where the Court itself must determine what happened before it can evaluate the law are excluded.

3. **Genuine merits adjudication** — The case must conclude with a standalone Article 10 merits finding, i.e.,  either a violation or no violation of Article 10. Cases that do not reach a substantive Article 10 conclusion, whether due to inadmissibility, procedural disposal, or because the finding is expressed under another article read in conjunction with Article 10, are excluded.

\#\# Output Format

Respond in the following JSON format:

\{\\
  "criteria": \{\\
    "is\_coherent\_merits":   true/false,\\
    "facts\_are\_settled":    true/false,\\
    "is\_genuine\_merits":    true/false\\
  \},\\
  "include": true/false,\\
  "justification": "..."\\
\}

`include` should be `true` if and only if all criteria are `true`. The justification should be a single concise sentence explaining the decision.

\#\# Input

\{case\_document\}
}

\paragraph{Pre-processing.} The system prompt is identical to the one for Article 11 (Section~\ref{app:prompts/preproc}).
In the user prompt, the target section becomes the Article 10 reasoning section (examined standalone or jointly with Article 11, reversing the Article 11 wording).

{\small\ttfamily
\# USER PROMPT (changed lines only)

Parse the following European Court of Human Rights judgment into structured paragraph lists for the Facts section and the Article 10 reasoning section only.

\#\# OUTPUT FORMAT

\{\\
  "facts": [...],\\
  "art10": [...]\\
\}

\#\# ART. 10 SECTION

The full section within "THE LAW" where the Court examines the Article 10 complaint on freedom of expression, whether standalone or jointly with Article 11.
}

\paragraph{Atomic fact division.} This stage does not mention the article or its elements, so the system and user prompts are identical to those for Article 11 (Section~\ref{app:prompts/atomic_facts}).

\paragraph{Isolation of conclusions.} This is the main doctrinal change. The scope test is reformulated around the three guarantees and the applicant's involvement (the Article 11 questions on assembly, participation and peacefulness no longer apply), the interference test gains a second question on whether the measure is a formality, condition, restriction or penalty,
and the legitimate aims follow the longer list in Article 10(2). The other elements and the output format are unchanged.

{\small\ttfamily
\# SYSTEM PROMPT

You are a legal expert specializing in European human rights law, with deep knowledge of Article 10 ECHR jurisprudence on freedom of expression. Your task is to analyze ECtHR judgments to build a structured understanding of how courts reason about Article 10 cases. You will work through a judgment in stages, building on your prior analysis at each step. Your analysis should be thorough, precise, and faithful to what the Court actually said.

---

\# USER PROMPT

Read the following reasoning section from a decision by the European Court of Human Rights, and then analyze the questions below.

**Reasoning text:**
\{reasoning\_text\}

**1.** Does the Court come to the conclusion that Article 10 ECHR has been violated? If yes, from which paragraph(s) do you take this information? Quote the paragraph number and the wording.

**2.** Please also give us the decisions of the Court regarding the elements of the canonical doctrine of Article 10 ECHR. In the Court's established jurisprudence, Article 10 ECHR requires the following tests:

- **(a) Scope:** i) does the case deal with an expression, with access to information, or with the activities of the media? ii) has the applicant been actively involved in the contested activity?

- **(b) Interference:** i) did some governmental act or omission interfere with the applicant's freedom of expression? ii) can the contested intervention be qualified as a formality, condition, restriction or penalty?

- **(c) Prescribed by law:** i) is the intervention sufficiently foreseeable, given the wording of the law? ii) is the provision that empowers government sufficiently accessible? iii) does the provision come with sufficient safeguards against abuse?

- **(d) Legitimate aim:** did it serve one of the legitimate aims listed in Article 10 ECHR, namely: national security, territorial integrity, public safety, the prevention of disorder, prevention of crime, the protection of health, the protection of morals, protection of the rights of others, the protection of the reputation of others, the prevention of the disclosure of information received in confidence, maintaining the authority and impartiality of the judiciary?

- **(e) Necessity in a democratic society:** did government adequately balance the intensity of the intervention with the weight of the legitimate aim?

- **(f) Positive obligation:** has the Court spelled out a positive obligation for government? If so, has it come to the conclusion that the defendant government has violated this obligation?

Which of these canonical elements of the doctrine of Article 10 ECHR are explicitly discussed in the decision? Are elements of doctrine implied (e.g. by the fact that the Court explicitly discusses an element that logically requires an earlier element to be given)? Again, quote the paragraph number and the wording for each canonical element that the Court has explicitly discussed.

---

\#\# Output

Return a JSON object with two keys. In `analysis`, answer the questions above as fully as you would in a standalone response — for the final violation conclusion as well as each element and sub-element, quote the specific passages that ground each finding, treating each sub-element as a separate question requiring its own answer. In `results`, extract the structured conclusions as shown below.

\{\\
  "analysis": "Your full analysis here...",\\
  "results": \{\\
    "violated": "true/false/null",\\
    "elements": \{\\
      "scope":                         \{"conclusion": "yes/no/null", "discussed": "explicit/implied/not\_discussed"\},\\
      "scope\_expression":              \{"conclusion": "yes/no/null", "discussed": "explicit/implied/not\_discussed"\},\\
      "scope\_information":             \{"conclusion": "yes/no/null", "discussed": "explicit/implied/not\_discussed"\},\\
      "scope\_media":                   \{"conclusion": "yes/no/null", "discussed": "explicit/implied/not\_discussed"\},\\
      "scope\_involvement":             \{"conclusion": "yes/no/null", "discussed": "explicit/implied/not\_discussed"\},\\
      "interference":                  \{"conclusion": "yes/no/null", "discussed": "explicit/implied/not\_discussed"\},\\
      "prescribed\_by\_law":             \{"conclusion": "yes/no/null", "discussed": "explicit/implied/not\_discussed"\},\\
      "prescribed\_by\_law\_foreseeable": \{"conclusion": "yes/no/null", "discussed": "explicit/implied/not\_discussed"\},\\
      "prescribed\_by\_law\_accessible":  \{"conclusion": "yes/no/null", "discussed": "explicit/implied/not\_discussed"\},\\
      "prescribed\_by\_law\_safeguards":  \{"conclusion": "yes/no/null", "discussed": "explicit/implied/not\_discussed"\},\\
      "legitimate\_aim":                \{"conclusion": "yes/no/null", "discussed": "explicit/implied/not\_discussed", "aims\_found": []\},\\
      "necessity":                     \{"conclusion": "yes/no/null", "discussed": "explicit/implied/not\_discussed"\},\\
      "positive\_obligation":           \{"conclusion": "yes/no/null", "discussed": "explicit/implied/not\_discussed"\}\\
    \}\\
  \}\\
\}\\

- `analysis`: Remember to escape any quotes with \textbackslash

- `violated`: `"true"` if Article 10 has been violated, `"false"` if not, `"null"` if unclear

- `conclusion`: `"yes"` if the Court answers the element's question affirmatively, `"no"` if negatively, `"null"` if not discussed or unresolved

- `discussed`: `"explicit"` if directly addressed, `"implied"` if follows from other elements, `"not\_discussed"` if absent entirely

- `aims\_found`: list of applicable aims from: `"national\_security"`, `"territorial\_integrity"`, `"public\_safety"`, `"prevention\_of\_disorder"`, `"prevention\_of\_crime"`, `"protection\_of\_health"`, `"protection\_of\_morals"`, `"rights\_of\_others"`, `"reputation\_of\_others"`, `"breach\_of\_confidence"`, `"judicial\_authority"` — empty list `[]` if legitimate aim not discussed
}

\paragraph{Factual grounding.} Only the system prompt changes, from Article 11 and freedom of assembly to Article 10 and freedom of expression.
The user prompt is the same as for Article 11 (Section~\ref{app:prompts/fact_grounding}).

{\small\ttfamily
\# SYSTEM PROMPT

You are a legal expert specializing in European human rights law, with deep knowledge of Article 10 ECHR jurisprudence on freedom of expression. Your task is to analyze ECtHR judgments to build a structured understanding of how courts reason about Article 10 cases. You will work through a judgment in stages, building on your prior analysis at each step. Your analysis should be thorough, precise, and faithful to what the Court actually said.
}

\paragraph{Fact mapping.} The system prompt changes in the same way as the factual grounding. In the user prompt, the only change is that the three references to Art.~11 in the List~3 instructions refer to Art.~10; the user prompt is otherwise the same as for
Article 11 (Section~\ref{app:prompts/fact_mapping}).

{\small\ttfamily
\# SYSTEM PROMPT

You are a legal expert specializing in European human rights law, with deep knowledge of Article 10 ECHR jurisprudence on freedom of expression. Your task is to analyze ECtHR judgments to build a structured understanding of how courts reason about Article 10 cases. You will work through a judgment in stages, building on your prior analysis at each step. Your analysis should be thorough, precise, and faithful to what the Court actually said.

\# USER PROMPT (changed lines only)

**List 3:** In the final step, focus on the final conclusion (Art. 10 ECHR has been violated, or not): which "atomic facts" are strictly necessary for the Court to come to this conclusion, assuming that the Court will apply the doctrinal structure of Art. 10 ECHR. Focus on what the case actually turned on.

This procedure implies that any fact in List 3 must feature in List 2, and any fact listed in List 2 must feature in List 1. Put differently, what we want you to do is to narrow down on the atomic facts that are absolutely necessary to back up the legal reasoning of the Court, as grounded in its assessment of the facts (List 2). By contrast, in List 3, we do not want you to replicate the Court's reasoning. Rather in the light of the facts of the case: which atomic facts are essential for reaching the conclusion (Art. 10 violated or not) that the Court reached.
}

\paragraph{Downstream evaluation (full reasoning, FR).} The downstream prompt is analogous to Article 11 (Section~\ref{app:prompts/ds_vanilla}) with the article wording changed to Article 10 and the element questions replaced by their Article 10 versions.

{\small\ttfamily
\# SYSTEM PROMPT

You are a legal expert specializing in European human rights law, with deep knowledge of Article 10 ECHR jurisprudence on freedom of expression. Your task is to analyze the facts of an ECtHR case and reason about whether Article 10 has been violated.

---

\# USER PROMPT

You are analyzing the facts of an Article 10 case from the European Court of Human Rights. The facts of a case usually contain the circumstances of the case, including the submissions by the applicant, the defendant government and (potentially) other actors, the applicable national law, the decisions taken by national authorities and courts, and any other material the court deems important to understand and decide the case.

The Court's established jurisprudence on Article 10 ECHR requires the following canonical elements to be assessed:

- **(a) Scope:** i) Does the case deal with an expression, with access to information, or with the activities of the media? ii) Has the applicant been actively involved in the contested activity?

- **(b) Interference:** i) Did some governmental act or omission interfere with the applicant's freedom of expression? ii) Can the contested intervention be qualified as a formality, condition, restriction or penalty?

- **(c) Prescribed by law:** i) Is the intervention sufficiently foreseeable given the wording of the law? ii) Is the provision that empowers government sufficiently accessible? iii) Does the provision come with sufficient safeguards against abuse?

- **(d) Legitimate aim:** Did the interference serve one of the legitimate aims listed in Article 10(2) ECHR, namely: national security, territorial integrity, public safety, prevention of disorder, prevention of crime, protection of health, protection of morals, protection of the rights of others, protection of the reputation of others, prevention of the disclosure of information received in confidence, or maintaining the authority and impartiality of the judiciary?

- **(e) Necessity in a democratic society:** Did the government adequately balance the intensity of the intervention with the weight of the legitimate aim?

- **(f) Positive obligation:** Has the Court spelled out a positive obligation for the government? If so, has it concluded that the defendant government violated this obligation?

These elements follow a conditional doctrinal chain. Scope and interference are threshold questions: if either is not established, no violation can arise. If interference is established, the Court examines whether it was prescribed by law, pursued a legitimate aim, and was necessary in a democratic society. A "no" conclusion on any of these three justification elements means the interference was unlawful and Article 10 has been violated. Positive obligation is a parallel reasoning route applicable when the government failed to take positive steps to facilitate the right rather than actively interfering with it.

Now, your task is to identify the facts needed to reason about the case. To facilitate this task, we have already formalized this material into a set of "atomic facts", provided below.

**Atomic facts:**
\{atomic\_facts\}

For each canonical element, identify which atomic facts are needed to answer its question, and provide a conclusion. Then provide a final conclusion on whether Article 10 has been violated.

---

\#\# Output Format

Return a single JSON object. Do not include any text outside the JSON.

\{
  "elements": \{\\
    "scope": \{\\
      "conclusion": "yes/no",\\
      "facts": ["FXXX", "FXXX"]\\
    \},\\
    "interference": \{\\
      "conclusion": "yes/no",\\
      "facts": ["FXXX", "FXXX"]\\
    \},\\
    "prescribed\_by\_law": \{\\
      "conclusion": "yes/no",\\
      "facts": ["FXXX"]\\
    \},\\
    "legitimate\_aim": \{\\
      "conclusion": "yes/no",\\
      "facts": ["FXXX"]\\
    \},
    "necessity": \{\\
      "conclusion": "yes/no",\\
      "facts": ["FXXX", "FXXX"]\\
    \},\\
    "positive\_obligation": \{\\
      "conclusion": "yes/no",\\
      "facts": ["FXXX"]\\
    \}\\
  \},\\
  "violated": true/false\\
\}

- For each element, `conclusion` should be `yes` if the answer to that element's question is affirmative, and `no` if it is negative.

- `violated` is a boolean — `true` if Article 10 has been violated, `false` otherwise

- Only include fact IDs that appear in the atomic facts provided above

- Return an empty list `[]` for `facts` if no atomic facts are relevant to an element
}

\clearpage
\newpage

\section{Downstream Evaluation: Detailed Results}
\label{app:fullresults}

This Appendix provides more detailed analysis for the results reported in the main paper for violation, element conclusions and fact-mapping.
We also analyze the differences between the elements.

\subsection{Implementation Details}
\label{app:implementation}

Open-weight models were run on a single H200 GPU, except {\tt GPT-OSS 120B} and {\tt Qwen 3.5 122B}, which required four H200 GPUs each.
We used the respective APIs for the closed-weight frontier models.
We used temperature $0.0$ across conditions, switching to $1.0$ for retries when a model failed to produce the required output format.
For the extended-thinking experiments in App.~\ref{app:additional_expt/thinking}, we used temperature $1.0$ for {\tt Claude Sonnet 4.6} and {\tt GPT-5.4}.

\subsection{Baseline Computation}
\label{app/baselines}

We evaluate two random baselines, averaging results over $100$ independent trials.

\paragraph{Uniform baseline.}
For violation prediction and element conclusions, which are binary prediction tasks, we independently sample \textit{yes}/\textit{no} with probability $0.5$.
For fact mapping, each atomic fact is selected with probability $0.5$, independently for each legal element.

\paragraph{Calibrated baseline.}
This baseline samples from the observed class distribution for the binary tasks.
For violation prediction, we sample an affirmative answer with probability equal to the benchmark's violation rate ($361/375$).
For element conclusions, we use the proportion of affirmative gold answers among active cases for that element.

For fact-mapping, the calibrated baseline selects facts at random, but uses the gold lists to determine how many to select.
For each element in both Lists~1 and~2, we first calculate the mean number of gold facts across active cases for the respective element and list combination.
Then, for each case, we draw a number from a Poisson distribution with the aforementioned mean, and randomly select that many atomic facts.
We sample without replacement and cap the number selected at the number of facts available.
We apply the same procedure to every element and list.
For List~3, we select facts once per case, using its average gold-list size of $6.21$.

\subsection{Metrics and Aggregation}
\label{app:fullresults/metrics}

We use $i$ to refer to an individual case and $e$ to refer to one of the six legal elements. An element is active in a case if the benchmark assigns it a \textit{yes} or \textit{no} conclusion, rather than leaving it unassessed.
For each element $e$, let $C_e$ be the set of cases in which it is active. For each case $i$, let $E_i$ be the set of elements active in that case.

\textbf{Violations.}
For violation prediction, we compute F1 separately for the violation and no-violation classes across all cases, and report their mean as \textit{m}F1.

\textbf{Element Conclusions.}
For each element $e$, we compute \textit{m}F1 over the cases in $C_e$, by averaging the F1 for \textit{yes} and \textit{no} classes.
If a class is absent from both the gold labels and predictions (e.g., \textit{no} for \textit{scope} and \textit{interference}), its F1 is set to zero.
The overall element-conclusion score is the average across the six elements:
\[
\mathrm{mF1}_{\mathrm{elements}}
= \frac{1}{6}\sum_{e=1}^{6}\mathrm{mF1}_e.
\]

\textbf{Fact Mapping.}
For Lists~1 and~2, let $S_{ie}$ be the set of facts predicted for element $e$ in case $i$, and $G_{ie}^{(\ell)}$ the corresponding
gold set for List~$\ell$, where $\ell \in \{1,2\}$.
We compute precision, recall and F1 separately for each case $i$ and element $e$, comparing $S_{ie}$ with $G_{ie}^{(\ell)}$.
For each list, we include only pairs where the element is active and the gold set is non-empty:
\[
C_e^{(\ell)}=\{i\in C_e:|G_{ie}^{(\ell)}|>0\},
\qquad \ell\in\{1,2\}.
\]
Each metric is first averaged across included cases within an element, then across the six elements:
\[
M^{(\ell)}
=\frac{1}{6}\sum_{e=1}^{6}
\frac{1}{|C_e^{(\ell)}|}
\sum_{i\in C_e^{(\ell)}}M(S_{ie},G_{ie}^{(\ell)}),
\qquad M\in\{P,R,F_1\}.
\]
An empty prediction receives zero for all three metrics.
Pairs with empty gold sets are excluded regardless of the prediction.

For List~3, we combine the predicted facts across all active elements in case $i$:
\[
S_i^{(3)}=\bigcup_{e\in E_i}S_{ie}.
\]
We compute precision, recall and F1 for each case by comparing $S_i^{(3)}$ with its gold List~3, denoted by $G_i^{(3)}$.
Each metric is then averaged over cases with a non-empty gold list:
\[
M^{(3)}=\frac{1}{|C^{(3)}|}
\sum_{i\in C^{(3)}}M(S_i^{(3)},G_i^{(3)}),
\qquad
C^{(3)}=\{i:|G_i^{(3)}|>0\}.
\]
As with Lists~1 and 2 above, an empty prediction receives zero for all three metrics.

\subsection{Predicting violation and element-wise conclusions}\label{app:viol-conc}

\begin{figure}[t]
    \centering
    \includegraphics[width=0.95\linewidth]{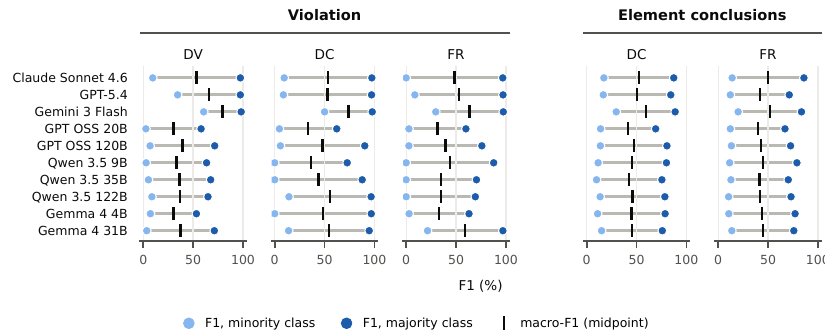}
    \caption{Majority and minority class F1 scores (\%) for violation and element conclusions. The black marker represents macro-F1. The gap represents the difference in percentage points.}
    \label{fig:majority-minority-viol-conc}
\end{figure}

Table~\ref{tab:main_results} reports violation and element-conclusion performance across DV, DC and FR.
While accuracy is high, especially for frontier models, \textit{m}F1 is substantially lower.
Figure~\ref{fig:majority-minority-viol-conc} shows majority- and minority-class F1 for both tasks in all applicable conditions.
For element conclusions, majority and minority classes are identified separately for each element, and their F1 scores are averaged across all six elements.
For \textit{scope} and \textit{interference}, the absent minority class is assigned F1 score of zero.

Majority-class F1 exceeds minority-class F1 for all models in both tasks.
In DV, frontier models achieve $94.9-96.5\%$ violation accuracy, but {\tt Claude}, {\tt GPT} and {\tt Gemini} correctly identify only $1$, $5$ and $10$ of the $14$ non-violated cases, respectively (TN in Table~\ref{tab:violation_confusion}).
The element-conclusion results show a similar class gap in both DC and FR.
Thus, high accuracy does not guarantee strong performance on minority-class outcomes.

Requiring intermediate element conclusions (DV $\rightarrow$ DC) improves violation \textit{m}F1 for all open-weight models, but these gains do not consistently extend to the minority class.
{\tt Qwen 3.5 122B} improves in terms of overall ($\textcolor{green!60!black}{\bm{\uparrow}} 50.8\%$ \textit{m}F1), but its TN count drops $9\rightarrow2$.
Similarly, {\tt Gemma 4 4B} shows improvement of ($\textcolor{green!60!black}{\bm{\uparrow}} 60.3\%$ \textit{m}F1), while TN drops $9\rightarrow0$.
Frontier models also lose minority-class performance: {\tt GPT}'s TN drops $5\rightarrow1$, and {\tt Gemini}'s drops $10\rightarrow8$.

Additionally requiring the models to map facts (DC $\rightarrow$ FR) reduces violation \textit{m}F1 for 7/10 models.
For {\tt Gemini}, minority-class performance drops ($\textcolor{red!60!black}{\bm{\downarrow}} 40.8\%$ F1; TN: $8\rightarrow4$), while majority-class F1 remains nearly unchanged.
However, {\tt Qwen 3.5 9B} and {\tt Gemma 4 31B} improve in violation prediction ($\textcolor{green!60!black}{\bm{\uparrow}} 20.9\%$ and $\textcolor{green!60!black}{\bm{\uparrow}} 8.8\%$ \textit{m}F1, respectively).
For element conclusions, all models perform worse, with larger drops for {\tt GPT} ($\textcolor{red!60!black}{\bm{\downarrow}} 17.2\%$ \textit{m}F1) and {\tt Gemini} ($\textcolor{red!60!black}{\bm{\downarrow}} 12.5\%$ \textit{m}F1).
Thus, requiring the models to explicitly map the supporting facts does not consistently improve violation prediction, and reduces element-conclusion performance across all tested models.

\begin{table}[t]
  \caption{Violation confusion matrices under the DV, DC and FR conditions (gold: $361$ violated, $14$ not violated). Violated is the positive class. Accuracy and \textit{m}F1 values are in percentage.}
  \label{tab:violation_confusion}
  \centering
  \footnotesize
  \setlength{\tabcolsep}{2.5pt}
  \begin{tabular}{lrrrrrr@{}p{12pt}@{}rrrrrr@{}p{12pt}@{}rrrrrr}
    \toprule
     & \multicolumn{6}{c}{DV} & & \multicolumn{6}{c}{DC} & & \multicolumn{6}{c}{FR} \\
    \cmidrule(lr){2-7}\cmidrule(lr){9-14}\cmidrule(lr){16-21}
    Model & TP & TN & FP & FN & Acc & \textit{m}F1 & & TP & TN & FP & FN & Acc & \textit{m}F1 & & TP & TN & FP & FN & Acc & \textit{m}F1 \\
    \midrule
    \multicolumn{21}{l}{\textit{Closed weights}} \\
    Claude Sonnet 4.6 & $355$ & $1$ & $13$ & $6$ & $94.9$ & $53.5$ & & $355$ & $1$ & $13$ & $6$ & $94.9$ & $53.5$ & & $352$ & $0$ & $14$ & $9$ & $93.9$ & $48.4$ \\
    GPT-5.4 & $351$ & $5$ & $9$ & $10$ & $94.9$ & $65.9$ & & $353$ & $1$ & $13$ & $8$ & $94.4$ & $52.9$ & & $353$ & $1$ & $13$ & $8$ & $94.4$ & $52.9$ \\
    Gemini 3 Flash & $352$ & $10$ & $4$ & $9$ & $96.5$ & $79.4$ & & $351$ & $8$ & $6$ & $10$ & $95.7$ & $73.9$ & & $352$ & $4$ & $10$ & $9$ & $94.9$ & $63.5$ \\
    \midrule
    \multicolumn{21}{l}{\textit{Open weights}} \\
    GPT OSS 20B & $152$ & $3$ & $11$ & $209$ & $41.3$ & $30.3$ & & $167$ & $5$ & $9$ & $194$ & $45.9$ & $33.4$ & & $159$ & $3$ & $11$ & $202$ & $43.2$ & $31.3$ \\
    GPT OSS 120B & $206$ & $6$ & $8$ & $155$ & $56.5$ & $39.3$ & & $307$ & $2$ & $12$ & $54$ & $82.4$ & $48.0$ & & $228$ & $2$ & $12$ & $133$ & $61.3$ & $39.3$ \\
    Qwen 3.5 9B & $173$ & $3$ & $11$ & $188$ & $46.9$ & $33.2$ & & $214$ & $0$ & $14$ & $147$ & $57.1$ & $36.3$ & & $293$ & $0$ & $14$ & $68$ & $78.1$ & $43.9$ \\
    Qwen 3.5 35B & $189$ & $5$ & $9$ & $172$ & $51.7$ & $36.4$ & & $293$ & $0$ & $14$ & $68$ & $78.1$ & $43.9$ & & $204$ & $0$ & $14$ & $157$ & $54.4$ & $35.2$ \\
    Qwen 3.5 122B & $176$ & $9$ & $5$ & $185$ & $49.3$ & $36.8$ & & $349$ & $2$ & $12$ & $12$ & $93.6$ & $55.5$ & & $199$ & $0$ & $14$ & $162$ & $53.1$ & $34.7$ \\
    Gemma 4 4B & $133$ & $9$ & $5$ & $228$ & $37.9$ & $30.2$ & & $352$ & $0$ & $14$ & $9$ & $93.9$ & $48.4$ & & $171$ & $3$ & $11$ & $190$ & $46.4$ & $32.9$ \\
    Gemma 4 31B & $206$ & $3$ & $11$ & $155$ & $55.7$ & $37.4$ & & $335$ & $3$ & $11$ & $26$ & $90.1$ & $54.4$ & & $350$ & $3$ & $11$ & $11$ & $94.1$ & $59.2$ \\
    \bottomrule
  \end{tabular}
\end{table}

\begin{wrapfigure}{r}{0.52\textwidth}
    \centering
    \includegraphics[width=0.5\textwidth]{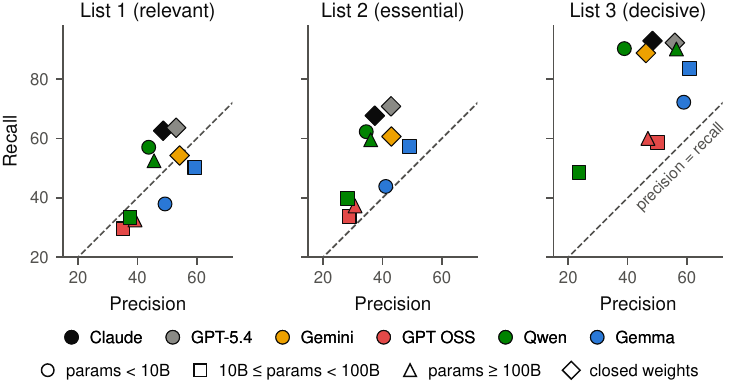}
    \caption{Precision vs. Recall (in \%) for Lists~1,2,3. Each point represents an evaluated model. Lists~1 and 2 values are averaged across elements.}
    \label{fig:precision-recall}
\end{wrapfigure}

\subsection{Mapping facts to legal elements} \label{app:fact-mapping}
Each model produces one fact mapping per element, which we score against Lists~1 and~2. For List~3, we pool these predictions into a single set of facts and compare it with the case-level gold list.
These lists progressively narrow down the Court's factual support to the core of legal syllogism: from \textit{relevant facts} (List~1) to \textit{essential facts} (List~2) and \textit{decisive facts} (List~3).
This lets us examine both whether the mapping covers the Court's supporting facts and whether it selects facts beyond those needed for its conclusions.
Figure~\ref{fig:precision-recall} shows the tradeoff between precision and recall for these comparisons.
Table~\ref{tab:fact-mapping-summ} reports the overall scores for all three lists, while Table~\ref{tab:fact-mapping-elems} provides the element-wise breakdown for Lists~1 and 2.

Models tend to select more facts than the Court uses to support its conclusions.
Against Lists~2 and~3, recall exceeds precision for every model.
For example, frontier models are able to map $88.8-92.9\%$ of the Court's \textit{decisive facts}, but achieve only $46.2-55.9\%$ precision.
For the key task of mapping \textit{essential facts} to individual elements (List~2), both recall ($60.7-70.9\%$) and precision ($37.4-43.0\%$) are lower than the other lists.
The low precision does not just reflect the inclusion of relevant but non-essential facts: against the broader List~1, frontier-model precision is only $48.7-54.2\%$.
Thus, the models' mappings do not merely include a broader set of relevant facts; many mapped facts are not part of the set of facts the Court relies on to conclude an element.

\begin{table}[t]
  \caption{Precision (P), recall (R) and F1 (\%) of the fact mapping in the FR condition. List~1 (\textit{relevant facts}) and List~2 (\textit{essential facts}) are averaged over the six elements; the per-element scores are in Table~\ref{tab:fact-mapping-elems}. List~3 (\textit{decisive facts}) is scored per case. Best model per column in \textbf{bold}.}
  \label{tab:fact-mapping-summ}
  \centering
  \footnotesize
  \setlength{\tabcolsep}{3.0pt}
  \begin{tabular}{lrrr@{}p{8pt}@{}rrr@{}p{8pt}@{}rrr}
    \toprule
     & \multicolumn{3}{c}{List~1 (\textit{relevant})} & & \multicolumn{3}{c}{List~2 (\textit{essential})} & & \multicolumn{3}{c}{List~3 (\textit{decisive})} \\
    \cmidrule(lr){2-4}\cmidrule(lr){6-8}\cmidrule(lr){10-12}
    Model & P & R & F1 & & P & R & F1 & & P & R & F1 \\
    \midrule
    \multicolumn{12}{l}{\textit{Closed weights}} \\
    Claude & $48.7$ & $62.6$ & $51.5$ & & $37.4$ & $67.7$ & $44.4$ & & $48.4$ & $\mathbf{92.9}$ & $57.9$ \\
    GPT-5.4 & $53.1$ & $\mathbf{63.6}$ & $\mathbf{54.3}$ & & $42.9$ & $\mathbf{70.9}$ & $48.8$ & & $55.9$ & $92.3$ & $63.0$ \\
    Gemini & $54.2$ & $54.3$ & $51.2$ & & $43.0$ & $60.7$ & $47.2$ & & $46.2$ & $88.8$ & $57.6$ \\
    \midrule
    \multicolumn{12}{l}{\textit{Open weights}} \\
    OSS 20B & $35.2$ & $29.5$ & $27.2$ & & $28.7$ & $33.6$ & $26.2$ & & $50.0$ & $58.7$ & $46.9$ \\
    OSS 120B & $39.2$ & $32.6$ & $32.0$ & & $30.8$ & $37.3$ & $30.3$ & & $46.8$ & $60.1$ & $50.1$ \\
    Qwen 9B & $43.8$ & $57.1$ & $44.7$ & & $34.5$ & $62.3$ & $39.4$ & & $38.9$ & $90.3$ & $50.2$ \\
    Qwen 35B & $37.5$ & $33.4$ & $31.6$ & & $28.2$ & $39.7$ & $29.5$ & & $23.6$ & $48.5$ & $29.3$ \\
    Qwen 122B & $45.6$ & $52.5$ & $44.6$ & & $36.0$ & $59.6$ & $40.5$ & & $56.5$ & $90.2$ & $63.0$ \\
    Gemma 4B & $49.3$ & $38.0$ & $39.7$ & & $41.1$ & $43.9$ & $39.9$ & & $58.9$ & $72.2$ & $62.8$ \\
    Gemma 31B & $\mathbf{59.4}$ & $50.2$ & $51.3$ & & $\mathbf{49.0}$ & $57.4$ & $\mathbf{49.7}$ & & $\mathbf{60.8}$ & $83.6$ & $\mathbf{67.1}$ \\
    \bottomrule
  \end{tabular}
\end{table}

\begin{table}[p]
  \caption{Element-conclusion confusion matrices for the six elements of Article 11 (panels (a)--(f)) in the \textbf{DC condition}. Each panel header gives the number of cases with gold answer \textit{yes} and \textit{no}. \textit{Yes} is the positive class. Acc is accuracy and \textit{m}F1 is macro-F1 over the two classes, both in percent. Scope and Interference have zero gold \textit{no} cases, so their \textit{m}F1 is at most $50\%$.}
  \label{tab:element_confusion_dc}
  \centering
  \footnotesize
  \setlength{\tabcolsep}{2.0pt}
  \begin{tabular*}{\textwidth}{@{\extracolsep{\fill}}lrrrrrr@{}p{8pt}@{}lrrrrrr}
    \toprule
    \multicolumn{7}{c}{\textbf{(a)} Scope (gold: $375$ yes, $0$ no)} & & \multicolumn{7}{c}{\textbf{(b)} Interference (gold: $374$ yes, $0$ no)} \\
    \cmidrule(lr){1-7}\cmidrule(lr){9-15}
    Model & TP & TN & FP & FN & Acc & \textit{m}F1 & & Model & TP & TN & FP & FN & Acc & \textit{m}F1 \\
    \cmidrule(lr){1-7}\cmidrule(lr){9-15}
    Claude & $369$ & $0$ & $0$ & $6$ & $98.4$ & $49.6$ & & Claude & $368$ & $0$ & $0$ & $6$ & $98.4$ & $49.6$ \\
    GPT-5.4 & $368$ & $0$ & $0$ & $7$ & $98.1$ & $49.5$ & & GPT-5.4 & $366$ & $0$ & $0$ & $8$ & $97.9$ & $49.5$ \\
    Gemini & $368$ & $0$ & $0$ & $7$ & $98.1$ & $49.5$ & & Gemini & $367$ & $0$ & $0$ & $7$ & $98.1$ & $49.5$ \\
    \cmidrule(lr){1-7}\cmidrule(lr){9-15}
    OSS 20B & $323$ & $0$ & $0$ & $52$ & $86.1$ & $46.3$ & & OSS 20B & $320$ & $0$ & $0$ & $54$ & $85.6$ & $46.1$ \\
    OSS 120B & $326$ & $0$ & $0$ & $49$ & $86.9$ & $46.5$ & & OSS 120B & $331$ & $0$ & $0$ & $43$ & $88.5$ & $47.0$ \\
    Qwen 9B & $363$ & $0$ & $0$ & $12$ & $96.8$ & $49.2$ & & Qwen 9B & $361$ & $0$ & $0$ & $13$ & $96.5$ & $49.1$ \\
    Qwen 35B & $308$ & $0$ & $0$ & $67$ & $82.1$ & $45.1$ & & Qwen 35B & $305$ & $0$ & $0$ & $69$ & $81.6$ & $44.9$ \\
    Qwen 122B & $367$ & $0$ & $0$ & $8$ & $97.9$ & $49.5$ & & Qwen 122B & $365$ & $0$ & $0$ & $9$ & $97.6$ & $49.4$ \\
    Gemma 4B & $368$ & $0$ & $0$ & $7$ & $98.1$ & $49.5$ & & Gemma 4B & $367$ & $0$ & $0$ & $7$ & $98.1$ & $49.5$ \\
    Gemma 31B & $347$ & $0$ & $0$ & $28$ & $92.5$ & $48.1$ & & Gemma 31B & $352$ & $0$ & $0$ & $22$ & $94.1$ & $48.5$ \\
    \midrule
    \multicolumn{7}{c}{\textbf{(c)} Prescribed by law (gold: $309$ yes, $59$ no)} & & \multicolumn{7}{c}{\textbf{(d)} Legitimate aim (gold: $332$ yes, $7$ no)} \\
    \cmidrule(lr){1-7}\cmidrule(lr){9-15}
    Model & TP & TN & FP & FN & Acc & \textit{m}F1 & & Model & TP & TN & FP & FN & Acc & \textit{m}F1 \\
    \cmidrule(lr){1-7}\cmidrule(lr){9-15}
    Claude & $211$ & $36$ & $23$ & $98$ & $67.1$ & $57.5$ & & Claude & $312$ & $3$ & $4$ & $20$ & $92.9$ & $58.1$ \\
    GPT-5.4 & $139$ & $37$ & $22$ & $170$ & $47.8$ & $43.5$ & & GPT-5.4 & $319$ & $3$ & $4$ & $13$ & $95.0$ & $61.7$ \\
    Gemini & $267$ & $39$ & $20$ & $42$ & $83.2$ & $72.7$ & & Gemini & $312$ & $2$ & $5$ & $20$ & $92.6$ & $55.0$ \\
    \cmidrule(lr){1-7}\cmidrule(lr){9-15}
    OSS 20B & $95$ & $38$ & $21$ & $214$ & $36.1$ & $34.6$ & & OSS 20B & $161$ & $3$ & $4$ & $171$ & $48.4$ & $34.1$ \\
    OSS 120B & $96$ & $55$ & $4$ & $213$ & $41.0$ & $40.3$ & & OSS 120B & $248$ & $4$ & $3$ & $84$ & $74.3$ & $46.7$ \\
    Qwen 9B & $138$ & $27$ & $32$ & $171$ & $44.8$ & $39.3$ & & Qwen 9B & $181$ & $0$ & $7$ & $151$ & $53.4$ & $34.8$ \\
    Qwen 35B & $129$ & $34$ & $25$ & $180$ & $44.3$ & $40.3$ & & Qwen 35B & $177$ & $1$ & $6$ & $155$ & $52.5$ & $35.0$ \\
    Qwen 122B & $46$ & $55$ & $4$ & $263$ & $27.4$ & $27.4$ & & Qwen 122B & $324$ & $0$ & $7$ & $8$ & $95.6$ & $48.9$ \\
    Gemma 4B & $137$ & $13$ & $46$ & $172$ & $40.8$ & $33.2$ & & Gemma 4B & $176$ & $0$ & $7$ & $156$ & $51.9$ & $34.2$ \\
    Gemma 31B & $108$ & $33$ & $26$ & $201$ & $38.3$ & $35.6$ & & Gemma 31B & $153$ & $2$ & $5$ & $179$ & $45.7$ & $32.3$ \\
    \midrule
    \multicolumn{7}{c}{\textbf{(e)} Necessity (gold: $13$ yes, $326$ no)} & & \multicolumn{7}{c}{\textbf{(f)} Positive obligation (gold: $19$ yes, $9$ no)} \\
    \cmidrule(lr){1-7}\cmidrule(lr){9-15}
    Model & TP & TN & FP & FN & Acc & \textit{m}F1 & & Model & TP & TN & FP & FN & Acc & \textit{m}F1 \\
    \cmidrule(lr){1-7}\cmidrule(lr){9-15}
    Claude & $0$ & $326$ & $0$ & $13$ & $96.2$ & $49.0$ & & Claude & $8$ & $6$ & $3$ & $11$ & $50.0$ & $49.7$ \\
    GPT-5.4 & $0$ & $326$ & $0$ & $13$ & $96.2$ & $49.0$ & & GPT-5.4 & $8$ & $6$ & $3$ & $11$ & $50.0$ & $49.7$ \\
    Gemini & $7$ & $322$ & $4$ & $6$ & $97.1$ & $78.4$ & & Gemini & $7$ & $7$ & $2$ & $12$ & $50.0$ & $50.0$ \\
    \cmidrule(lr){1-7}\cmidrule(lr){9-15}
    OSS 20B & $4$ & $268$ & $58$ & $9$ & $80.2$ & $49.8$ & & OSS 20B & $4$ & $7$ & $2$ & $15$ & $39.3$ & $38.6$ \\
    OSS 120B & $1$ & $271$ & $55$ & $12$ & $80.2$ & $45.9$ & & OSS 120B & $15$ & $3$ & $6$ & $4$ & $64.3$ & $56.2$ \\
    Qwen 9B & $0$ & $326$ & $0$ & $13$ & $96.2$ & $49.0$ & & Qwen 9B & $9$ & $6$ & $3$ & $10$ & $53.6$ & $53.0$ \\
    Qwen 35B & $0$ & $326$ & $0$ & $13$ & $96.2$ & $49.0$ & & Qwen 35B & $8$ & $4$ & $5$ & $11$ & $42.9$ & $41.7$ \\
    Qwen 122B & $1$ & $323$ & $3$ & $12$ & $95.6$ & $54.7$ & & Qwen 122B & $8$ & $5$ & $4$ & $11$ & $46.4$ & $45.8$ \\
    Gemma 4B & $0$ & $325$ & $1$ & $13$ & $95.9$ & $48.9$ & & Gemma 4B & $7$ & $8$ & $1$ & $12$ & $53.6$ & $53.5$ \\
    Gemma 31B & $2$ & $320$ & $6$ & $11$ & $95.0$ & $58.2$ & & Gemma 31B & $8$ & $6$ & $3$ & $11$ & $50.0$ & $49.7$ \\
    \bottomrule
  \end{tabular*}
\end{table}

\begin{table}[p]
  \caption{Element-conclusion confusion matrices for the six elements of Article 11 (panels (a)--(f)) in the \textbf{FR condition}. Each panel header gives the number of cases with gold answer \textit{yes} and \textit{no}. \textit{Yes} is the positive class. Acc is accuracy and \textit{m}F1 is macro-F1 over the two classes, both in percent. Scope and Interference have zero gold \textit{no} cases, so their \textit{m}F1 is at most $50\%$.}
  \label{tab:element_confusion_fr}
  \centering
  \footnotesize
  \setlength{\tabcolsep}{2.0pt}
  \begin{tabular*}{\textwidth}{@{\extracolsep{\fill}}lrrrrrr@{}p{8pt}@{}lrrrrrr}
    \toprule
    \multicolumn{7}{c}{\textbf{(a)} Scope (gold: $375$ yes, $0$ no)} & & \multicolumn{7}{c}{\textbf{(b)} Interference (gold: $374$ yes, $0$ no)} \\
    \cmidrule(lr){1-7}\cmidrule(lr){9-15}
    Model & TP & TN & FP & FN & Acc & \textit{m}F1 & & Model & TP & TN & FP & FN & Acc & \textit{m}F1 \\
    \cmidrule(lr){1-7}\cmidrule(lr){9-15}
    Claude & $367$ & $0$ & $0$ & $8$ & $97.9$ & $49.5$ & & Claude & $365$ & $0$ & $0$ & $9$ & $97.6$ & $49.4$ \\
    GPT-5.4 & $368$ & $0$ & $0$ & $7$ & $98.1$ & $49.5$ & & GPT-5.4 & $366$ & $0$ & $0$ & $8$ & $97.9$ & $49.5$ \\
    Gemini & $368$ & $0$ & $0$ & $7$ & $98.1$ & $49.5$ & & Gemini & $368$ & $0$ & $0$ & $6$ & $98.4$ & $49.6$ \\
    \cmidrule(lr){1-7}\cmidrule(lr){9-15}
    OSS 20B & $311$ & $0$ & $0$ & $64$ & $82.9$ & $45.3$ & & OSS 20B & $325$ & $0$ & $0$ & $49$ & $86.9$ & $46.5$ \\
    OSS 120B & $278$ & $0$ & $0$ & $97$ & $74.1$ & $42.6$ & & OSS 120B & $293$ & $0$ & $0$ & $81$ & $78.3$ & $43.9$ \\
    Qwen 9B & $369$ & $0$ & $0$ & $6$ & $98.4$ & $49.6$ & & Qwen 9B & $367$ & $0$ & $0$ & $7$ & $98.1$ & $49.5$ \\
    Qwen 35B & $240$ & $0$ & $0$ & $135$ & $64.0$ & $39.0$ & & Qwen 35B & $237$ & $0$ & $0$ & $137$ & $63.4$ & $38.8$ \\
    Qwen 122B & $367$ & $0$ & $0$ & $8$ & $97.9$ & $49.5$ & & Qwen 122B & $365$ & $0$ & $0$ & $9$ & $97.6$ & $49.4$ \\
    Gemma 4B & $366$ & $0$ & $0$ & $9$ & $97.6$ & $49.4$ & & Gemma 4B & $365$ & $0$ & $0$ & $9$ & $97.6$ & $49.4$ \\
    Gemma 31B & $362$ & $0$ & $0$ & $13$ & $96.5$ & $49.1$ & & Gemma 31B & $364$ & $0$ & $0$ & $10$ & $97.3$ & $49.3$ \\
    \midrule
    \multicolumn{7}{c}{\textbf{(c)} Prescribed by law (gold: $309$ yes, $59$ no)} & & \multicolumn{7}{c}{\textbf{(d)} Legitimate aim (gold: $332$ yes, $7$ no)} \\
    \cmidrule(lr){1-7}\cmidrule(lr){9-15}
    Model & TP & TN & FP & FN & Acc & \textit{m}F1 & & Model & TP & TN & FP & FN & Acc & \textit{m}F1 \\
    \cmidrule(lr){1-7}\cmidrule(lr){9-15}
    Claude & $198$ & $35$ & $24$ & $111$ & $63.3$ & $54.4$ & & Claude & $273$ & $3$ & $4$ & $59$ & $81.4$ & $49.2$ \\
    GPT-5.4 & $95$ & $45$ & $14$ & $214$ & $38.0$ & $36.9$ & & GPT-5.4 & $146$ & $2$ & $5$ & $186$ & $43.7$ & $31.3$ \\
    Gemini & $200$ & $36$ & $23$ & $109$ & $64.1$ & $55.2$ & & Gemini & $224$ & $2$ & $5$ & $108$ & $66.7$ & $41.6$ \\
    \cmidrule(lr){1-7}\cmidrule(lr){9-15}
    OSS 20B & $107$ & $22$ & $37$ & $202$ & $35.1$ & $31.4$ & & OSS 20B & $130$ & $2$ & $5$ & $202$ & $38.9$ & $28.8$ \\
    OSS 120B & $72$ & $39$ & $20$ & $237$ & $30.2$ & $29.6$ & & OSS 120B & $132$ & $4$ & $3$ & $200$ & $40.1$ & $30.2$ \\
    Qwen 9B & $124$ & $32$ & $27$ & $185$ & $42.4$ & $38.6$ & & Qwen 9B & $184$ & $0$ & $7$ & $148$ & $54.3$ & $35.2$ \\
    Qwen 35B & $126$ & $28$ & $31$ & $183$ & $41.8$ & $37.4$ & & Qwen 35B & $153$ & $1$ & $6$ & $179$ & $45.4$ & $31.7$ \\
    Qwen 122B & $47$ & $56$ & $3$ & $262$ & $28.0$ & $27.9$ & & Qwen 122B & $170$ & $0$ & $7$ & $162$ & $50.1$ & $33.4$ \\
    Gemma 4B & $136$ & $16$ & $43$ & $173$ & $41.3$ & $34.3$ & & Gemma 4B & $151$ & $1$ & $6$ & $181$ & $44.8$ & $31.4$ \\
    Gemma 31B & $106$ & $31$ & $28$ & $203$ & $37.2$ & $34.5$ & & Gemma 31B & $149$ & $2$ & $5$ & $183$ & $44.5$ & $31.7$ \\
    \midrule
    \multicolumn{7}{c}{\textbf{(e)} Necessity (gold: $13$ yes, $326$ no)} & & \multicolumn{7}{c}{\textbf{(f)} Positive obligation (gold: $19$ yes, $9$ no)} \\
    \cmidrule(lr){1-7}\cmidrule(lr){9-15}
    Model & TP & TN & FP & FN & Acc & \textit{m}F1 & & Model & TP & TN & FP & FN & Acc & \textit{m}F1 \\
    \cmidrule(lr){1-7}\cmidrule(lr){9-15}
    Claude & $0$ & $325$ & $1$ & $13$ & $95.9$ & $48.9$ & & Claude & $9$ & $5$ & $4$ & $10$ & $50.0$ & $49.0$ \\
    GPT-5.4 & $1$ & $173$ & $153$ & $12$ & $51.3$ & $34.5$ & & GPT-5.4 & $9$ & $5$ & $4$ & $10$ & $50.0$ & $49.0$ \\
    Gemini & $3$ & $323$ & $3$ & $10$ & $96.2$ & $64.8$ & & Gemini & $7$ & $7$ & $2$ & $12$ & $50.0$ & $50.0$ \\
    \cmidrule(lr){1-7}\cmidrule(lr){9-15}
    OSS 20B & $4$ & $277$ & $49$ & $9$ & $82.9$ & $51.3$ & & OSS 20B & $3$ & $7$ & $2$ & $16$ & $35.7$ & $34.4$ \\
    OSS 120B & $1$ & $316$ & $10$ & $12$ & $93.5$ & $52.5$ & & OSS 120B & $14$ & $4$ & $5$ & $5$ & $64.3$ & $59.1$ \\
    Qwen 9B & $0$ & $326$ & $0$ & $13$ & $96.2$ & $49.0$ & & Qwen 9B & $8$ & $6$ & $3$ & $11$ & $50.0$ & $49.7$ \\
    Qwen 35B & $0$ & $326$ & $0$ & $13$ & $96.2$ & $49.0$ & & Qwen 35B & $7$ & $8$ & $1$ & $12$ & $53.6$ & $53.5$ \\
    Qwen 122B & $0$ & $324$ & $2$ & $13$ & $95.6$ & $48.9$ & & Qwen 122B & $8$ & $4$ & $5$ & $11$ & $42.9$ & $41.7$ \\
    Gemma 4B & $0$ & $325$ & $1$ & $13$ & $95.9$ & $48.9$ & & Gemma 4B & $7$ & $7$ & $2$ & $12$ & $50.0$ & $50.0$ \\
    Gemma 31B & $2$ & $320$ & $6$ & $11$ & $95.0$ & $58.2$ & & Gemma 31B & $8$ & $5$ & $4$ & $11$ & $46.4$ & $45.8$ \\
    \bottomrule
  \end{tabular*}
\end{table}

\begin{table}[p]
  \caption{Precision (P), recall (R) and F1 (\%) of the fact mapping in the FR condition for each element (panels (a)--(f)). List~1 contains the \textit{relevant facts} and List~2 the \textit{essential facts} (a subset of List~1). \textbf{Bold} values show the best model per column within a panel. Averages over the six elements for List~1,2 and List~3 values are in Table~\ref{tab:fact-mapping-summ}.}
  \label{tab:fact-mapping-elems}
  \centering
  \footnotesize
  \setlength{\tabcolsep}{2.0pt}
  \begin{tabular*}{\textwidth}{@{\extracolsep{\fill}}lrrr@{}p{4pt}@{}rrr@{}p{8pt}@{}lrrr@{}p{4pt}@{}rrr}
    \toprule
    \multicolumn{8}{c}{\textbf{(a)} Scope} & & \multicolumn{8}{c}{\textbf{(b)} Interference} \\
    \cmidrule(lr){1-8}\cmidrule(lr){10-17}
    & \multicolumn{3}{c}{List 1} & & \multicolumn{3}{c}{List 2} & & & \multicolumn{3}{c}{List 1} & & \multicolumn{3}{c}{List 2} \\
    \cmidrule(lr){2-4}\cmidrule(lr){6-8}\cmidrule(lr){11-13}\cmidrule(lr){15-17}
    Model & P & R & F1 & & P & R & F1 & & Model & P & R & F1 & & P & R & F1 \\
    \cmidrule(lr){1-8}\cmidrule(lr){10-17}
    Claude & $70.9$ & $\mathbf{75.7}$ & $70.4$ & & $57.5$ & $84.9$ & $65.4$ & & Claude & $63.1$ & $\mathbf{85.9}$ & $69.2$ & & $53.3$ & $\mathbf{91.1}$ & $62.4$ \\
    GPT-5.4 & $76.1$ & $72.7$ & $\mathbf{71.8}$ & & $66.1$ & $\mathbf{85.6}$ & $71.2$ & & GPT-5.4 & $69.5$ & $85.0$ & $73.0$ & & $59.9$ & $90.0$ & $66.5$ \\
    Gemini & $69.6$ & $71.0$ & $66.5$ & & $57.1$ & $81.7$ & $64.7$ & & Gemini & $77.2$ & $78.4$ & $\mathbf{75.5}$ & & $67.7$ & $86.8$ & $72.9$ \\
    \cmidrule(lr){1-8}\cmidrule(lr){10-17}
    OSS 20B & $74.4$ & $42.3$ & $49.8$ & & $67.5$ & $52.6$ & $55.1$ & & OSS 20B & $61.3$ & $53.5$ & $50.8$ & & $54.3$ & $57.5$ & $48.8$ \\
    OSS 120B & $68.9$ & $49.6$ & $54.5$ & & $59.4$ & $59.3$ & $57.1$ & & OSS 120B & $57.9$ & $53.2$ & $52.5$ & & $50.8$ & $58.1$ & $51.3$ \\
    Qwen 9B & $63.2$ & $73.2$ & $63.2$ & & $50.8$ & $81.5$ & $59.6$ & & Qwen 9B & $63.3$ & $83.2$ & $67.2$ & & $55.5$ & $87.3$ & $61.5$ \\
    Qwen 35B & $45.9$ & $35.4$ & $37.1$ & & $35.8$ & $44.6$ & $37.1$ & & Qwen 35B & $42.1$ & $41.4$ & $39.3$ & & $33.0$ & $48.0$ & $36.8$ \\
    Qwen 122B & $79.4$ & $67.7$ & $70.6$ & & $68.5$ & $79.4$ & $71.1$ & & Qwen 122B & $73.3$ & $80.2$ & $73.8$ & & $63.6$ & $86.7$ & $69.3$ \\
    Gemma 4B & $80.3$ & $61.4$ & $66.3$ & & $71.2$ & $73.2$ & $70.0$ & & Gemma 4B & $75.2$ & $66.0$ & $67.1$ & & $67.0$ & $71.6$ & $66.7$ \\
    Gemma 31B & $\mathbf{83.7}$ & $66.3$ & $71.1$ & & $\mathbf{73.6}$ & $79.2$ & $\mathbf{74.2}$ & & Gemma 31B & $\mathbf{80.7}$ & $74.5$ & $75.2$ & & $\mathbf{71.3}$ & $82.8$ & $\mathbf{74.0}$ \\
    \midrule
    \multicolumn{8}{c}{\textbf{(c)} Prescribed by law} & & \multicolumn{8}{c}{\textbf{(d)} Legitimate aim} \\
    \cmidrule(lr){1-8}\cmidrule(lr){10-17}
    & \multicolumn{3}{c}{List 1} & & \multicolumn{3}{c}{List 2} & & & \multicolumn{3}{c}{List 1} & & \multicolumn{3}{c}{List 2} \\
    \cmidrule(lr){2-4}\cmidrule(lr){6-8}\cmidrule(lr){11-13}\cmidrule(lr){15-17}
    Model & P & R & F1 & & P & R & F1 & & Model & P & R & F1 & & P & R & F1 \\
    \cmidrule(lr){1-8}\cmidrule(lr){10-17}
    Claude & $42.3$ & $57.6$ & $\mathbf{44.0}$ & & $27.1$ & $62.0$ & $34.3$ & & Claude & $38.8$ & $55.7$ & $41.9$ & & $25.0$ & $61.4$ & $32.2$ \\
    GPT-5.4 & $40.5$ & $\mathbf{60.3}$ & $43.8$ & & $27.6$ & $\mathbf{67.8}$ & $35.9$ & & GPT-5.4 & $37.7$ & $\mathbf{56.4}$ & $41.4$ & & $25.1$ & $\mathbf{64.9}$ & $32.7$ \\
    Gemini & $47.0$ & $51.4$ & $44.0$ & & $32.6$ & $58.6$ & $38.3$ & & Gemini & $44.0$ & $47.2$ & $41.3$ & & $26.9$ & $53.3$ & $32.0$ \\
    \cmidrule(lr){1-8}\cmidrule(lr){10-17}
    OSS 20B & $32.2$ & $42.3$ & $30.2$ & & $22.2$ & $46.1$ & $25.1$ & & OSS 20B & $26.1$ & $28.4$ & $22.4$ & & $18.0$ & $34.5$ & $20.3$ \\
    OSS 120B & $42.2$ & $39.2$ & $35.3$ & & $31.6$ & $46.5$ & $32.8$ & & OSS 120B & $32.4$ & $35.6$ & $30.8$ & & $21.3$ & $40.2$ & $24.5$ \\
    Qwen 9B & $36.4$ & $54.6$ & $37.7$ & & $25.1$ & $59.9$ & $30.4$ & & Qwen 9B & $39.0$ & $52.8$ & $38.5$ & & $26.5$ & $61.3$ & $31.7$ \\
    Qwen 35B & $44.7$ & $40.1$ & $37.3$ & & $33.5$ & $47.8$ & $35.3$ & & Qwen 35B & $\mathbf{47.2}$ & $40.3$ & $39.7$ & & $\mathbf{32.1}$ & $47.7$ & $34.3$ \\
    Qwen 122B & $38.1$ & $50.3$ & $38.3$ & & $26.8$ & $58.5$ & $32.9$ & & Qwen 122B & $43.3$ & $51.2$ & $\mathbf{42.5}$ & & $30.9$ & $60.4$ & $\mathbf{36.8}$ \\
    Gemma 4B & $45.4$ & $34.4$ & $34.1$ & & $32.8$ & $40.2$ & $32.0$ & & Gemma 4B & $40.3$ & $28.9$ & $30.2$ & & $28.9$ & $38.1$ & $29.5$ \\
    Gemma 31B & $\mathbf{52.8}$ & $42.5$ & $42.0$ & & $\mathbf{38.4}$ & $49.7$ & $\mathbf{38.9}$ & & Gemma 31B & $42.5$ & $38.9$ & $37.0$ & & $28.2$ & $44.9$ & $30.9$ \\
    \midrule
    \multicolumn{8}{c}{\textbf{(e)} Necessity} & & \multicolumn{8}{c}{\textbf{(f)} Positive obligation} \\
    \cmidrule(lr){1-8}\cmidrule(lr){10-17}
    & \multicolumn{3}{c}{List 1} & & \multicolumn{3}{c}{List 2} & & & \multicolumn{3}{c}{List 1} & & \multicolumn{3}{c}{List 2} \\
    \cmidrule(lr){2-4}\cmidrule(lr){6-8}\cmidrule(lr){11-13}\cmidrule(lr){15-17}
    Model & P & R & F1 & & P & R & F1 & & Model & P & R & F1 & & P & R & F1 \\
    \cmidrule(lr){1-8}\cmidrule(lr){10-17}
    Claude & $59.1$ & $\mathbf{76.0}$ & $63.7$ & & $50.4$ & $79.0$ & $57.6$ & & Claude & $17.9$ & $25.0$ & $19.9$ & & $11.0$ & $27.7$ & $14.3$ \\
    GPT-5.4 & $67.0$ & $74.9$ & $\mathbf{68.1}$ & & $59.2$ & $\mathbf{79.7}$ & $\mathbf{63.7}$ & & GPT-5.4 & $\mathbf{27.6}$ & $32.5$ & $\mathbf{27.8}$ & & $\mathbf{19.4}$ & $37.2$ & $\mathbf{22.6}$ \\
    Gemini & $68.5$ & $65.3$ & $65.2$ & & $61.6$ & $70.1$ & $63.5$ & & Gemini & $19.0$ & $12.3$ & $14.7$ & & $12.0$ & $13.6$ & $11.6$ \\
    \cmidrule(lr){1-8}\cmidrule(lr){10-17}
    OSS 20B & $15.5$ & $8.8$ & $8.2$ & & $9.9$ & $9.2$ & $7.0$ & & OSS 20B & $1.4$ & $2.1$ & $1.5$ & & $0.4$ & $1.9$ & $0.7$ \\
    OSS 120B & $18.2$ & $7.6$ & $9.8$ & & $12.2$ & $8.4$ & $8.9$ & & OSS 120B & $15.8$ & $10.1$ & $9.4$ & & $9.4$ & $11.3$ & $7.1$ \\
    Qwen 9B & $45.9$ & $59.9$ & $47.8$ & & $38.9$ & $61.5$ & $42.6$ & & Qwen 9B & $15.2$ & $18.8$ & $13.8$ & & $10.5$ & $22.4$ & $10.7$ \\
    Qwen 35B & $28.0$ & $24.2$ & $23.0$ & & $23.1$ & $27.3$ & $22.1$ & & Qwen 35B & $17.0$ & $19.3$ & $13.7$ & & $11.8$ & $22.7$ & $11.4$ \\
    Qwen 122B & $21.9$ & $29.8$ & $22.7$ & & $15.7$ & $33.4$ & $19.0$ & & Qwen 122B & $17.6$ & $\mathbf{35.8}$ & $20.0$ & & $10.6$ & $\mathbf{39.1}$ & $13.9$ \\
    Gemma 4B & $44.9$ & $32.4$ & $35.0$ & & $39.1$ & $34.2$ & $35.2$ & & Gemma 4B & $9.7$ & $4.6$ & $5.5$ & & $7.6$ & $5.7$ & $5.6$ \\
    Gemma 31B & $\mathbf{71.4}$ & $59.3$ & $62.0$ & & $\mathbf{63.9}$ & $63.7$ & $62.0$ & & Gemma 31B & $25.2$ & $19.9$ & $20.5$ & & $18.5$ & $24.1$ & $18.4$ \\
    \bottomrule
  \end{tabular*}
\end{table}

\subsection{Differences across legal elements} \label{app:elem-diff}

We now examine how the models' decisions vary across elements.
Tables~\ref{tab:element_confusion_dc} and~\ref{tab:element_confusion_fr} report conclusion results in DC and FR, while Table~\ref{tab:fact-mapping-elems} reports fact-mapping performance.

\paragraph{Scope and Interference.}
Frontier models perform similarly on these elements, achieving $49.4-49.6\%$ conclusion \textit{m}F1 in FR.
These scores are close to the ceiling score of $50\%$, since neither element contains negative gold cases.
Their List~2 F1 ranges from $62.4$ to $72.9\%$.
Thus, frontier models largely agree with the Court's affirmative conclusions, and their mappings overlap with the Court to a reasonable extent.

\paragraph{Prescribed by law and Legitimate aim.}
Conclusion performance varies more across models on these elements.
For \textit{Prescribed by law}, only {\tt Claude} and {\tt Gemini} exceed the always-yes baseline of $45.6\%$ \textit{m}F1 in both DC and FR.
For \textit{Legitimate aim}, frontier-model \textit{m}F1 ranges from $55.0-61.7\%$ in DC, but falls to $31.3-49.2\%$ in FR.
Fact mapping is also weaker than for \textit{Scope} and \textit{Interference}: frontier models' List~2 F1 ranges from $32.0-38.3\%$ across these two elements.

\paragraph{Necessity.}
For \textit{Necessity}, an always-\textit{no} predictor achieves $49.0\%$ \textit{m}F1 because the Court concludes \textit{no} in $326/339$ cases.
In FR, five models fail to identify any of the $13$ affirmative cases and score only $48.9-49.0\%$ \textit{m}F1.
{\tt Gemini} achieves the highest \textit{m}F1 in both conditions, but its score drops from $78.4\%$ in DC to $64.8\%$ in FR as correctly identified affirmative cases fall from seven to three.
Fact-mapping performance varies widely, with List~2 F1 ranging from $7.0$ to $63.7\%$.
{\tt GPT-5.4} achieves the highest fact-mapping score, yet its conclusion \textit{m}F1 is only $34.5\%$ in FR.
\textit{Necessity} therefore reveals particularly large differences between models, while also showing that the models strongest at predicting conclusions are not necessarily the strongest at mapping essential facts.

\paragraph{Positive obligation.}
While models differ substantially in fact-mapping for \textit{Necessity}, all of them struggle with \textit{Positive obligation}.
List~2 F1 remains below $23\%$ for every model, and recall never exceeds $39.1\%$.
Models therefore miss many of the facts the Court uses to support its conclusions for this element.
Performance on element conclusion is also quite poor: frontier models' \textit{m}F1 ranges from $49.0-50.0\%$ in FR, and no model exceeds $59.1\%$.

\clearpage
\newpage

\section{Additional Experiments}
\label{app:additional_expt}

We test whether changes to prompt scaffolding, prediction order, thinking mode, and in-context examples improve performance.
Each experiment varies one aspect of the full reasoning (FR) setup used in the main evaluation.

\subsection{Prompt Scaffolding}
\label{app:additional_expt/scaffolding}
Our main evaluation supplies the doctrinal questions and decision logic so that models need not reconstruct this framework themselves.
We test whether this guidance contributes to the difficulty of fact mapping by comparing three setups: element names only (minimal), element questions or definitions (partial), and questions with the logic of the decision chain (full).
The full setup corresponds to the FR condition.
The three prompts appear in App.~\ref{app:prompts/ds_minimal}, \ref{app:prompts/ds_partial}, and \ref{app:prompts/ds_vanilla}, respectively.

Table~\ref{tab:scaffolding} shows that additional guidance does not consistently help models reach the correct conclusions or identify the supporting facts.
In fact, minimal scaffolding yields higher List~1 and List~2 F1 than full scaffolding for 8/10 models.
This suggests that supplying the doctrinal framework does not necessarily help models apply it to the facts.
However, removing this guidance does not resolve the difficulty: List~2 F1 remains at or below $51.1\%$ across all three settings.
Thus, the fact-mapping limitations observed in our main evaluation are not confined to the fully scaffolded prompt.

\begin{table}[t]
  \caption{Effect of prompt scaffolding: minimal (element names), partial (element questions) and full (questions and chain; the FR condition). Metrics: violation macro-F1 (V), element-conclusion macro-F1 averaged over the six elements (E), and F1 for List~1 (L1), List~2 (L2) and List~3 (L3).}
  \label{tab:scaffolding}
  \centering
  \footnotesize
  \setlength{\tabcolsep}{1.6pt}
  \begin{tabular}{lrrrrr@{\hspace{8pt}}rrrrr@{\hspace{8pt}}rrrrr}
    \toprule
     & \multicolumn{5}{c}{Minimal} & \multicolumn{5}{c}{Partial} & \multicolumn{5}{c}{Full} \\
    \cmidrule(lr){2-6}\cmidrule(lr){7-11}\cmidrule(lr){12-16}
    Model & V & E & L1 & L2 & L3 & V & E & L1 & L2 & L3 & V & E & L1 & L2 & L3 \\
    \midrule
    \multicolumn{16}{l}{\textit{Closed weights}} \\
    Claude Sonnet 4.6 & $48.6$ & $49.5$ & $55.3$ & $47.9$ & $55.7$ & $48.6$ & $49.8$ & $52.8$ & $45.6$ & $57.3$ & $48.4$ & $50.0$ & $51.5$ & $44.4$ & $57.9$ \\
    GPT-5.4 & $53.2$ & $43.4$ & $\mathbf{55.4}$ & $49.1$ & $62.5$ & $56.6$ & $41.8$ & $\mathbf{54.2}$ & $48.7$ & $62.9$ & $52.9$ & $41.8$ & $\mathbf{54.3}$ & $48.8$ & $63.0$ \\
    Gemini 3 Flash & $\mathbf{69.5}$ & $\mathbf{52.0}$ & $54.3$ & $50.6$ & $64.1$ & $\mathbf{68.1}$ & $\mathbf{50.5}$ & $52.3$ & $48.2$ & $60.4$ & $\mathbf{63.5}$ & $\mathbf{51.8}$ & $51.2$ & $47.2$ & $57.6$ \\
    \midrule
    \multicolumn{16}{l}{\textit{Open weights}} \\
    GPT OSS 20B & $28.6$ & $44.1$ & $26.5$ & $23.9$ & $51.1$ & $26.1$ & $40.6$ & $27.5$ & $25.6$ & $44.8$ & $31.3$ & $39.6$ & $27.2$ & $26.2$ & $46.9$ \\
    GPT OSS 120B & $34.2$ & $43.7$ & $36.2$ & $33.6$ & $56.9$ & $40.1$ & $43.8$ & $31.4$ & $28.7$ & $43.8$ & $39.3$ & $43.0$ & $32.0$ & $30.3$ & $50.1$ \\
    Qwen 3.5 9B & $33.3$ & $44.4$ & $41.5$ & $36.1$ & $49.1$ & $35.9$ & $44.9$ & $43.7$ & $39.1$ & $50.7$ & $43.9$ & $45.3$ & $44.7$ & $39.4$ & $50.2$ \\
    Qwen 3.5 35B & $37.7$ & $44.4$ & $46.4$ & $44.5$ & $63.4$ & $36.7$ & $44.5$ & $44.2$ & $41.9$ & $59.3$ & $35.2$ & $41.6$ & $31.6$ & $29.5$ & $29.3$ \\
    Qwen 3.5 122B & $35.5$ & $45.2$ & $51.3$ & $47.7$ & $62.5$ & $33.0$ & $43.7$ & $47.9$ & $44.3$ & $62.9$ & $34.7$ & $41.8$ & $44.6$ & $40.5$ & $63.0$ \\
    Gemma 4 4B & $32.6$ & $43.1$ & $41.7$ & $41.2$ & $62.4$ & $24.7$ & $43.4$ & $43.0$ & $42.3$ & $62.9$ & $32.9$ & $43.9$ & $39.7$ & $39.9$ & $62.8$ \\
    Gemma 4 31B & $57.0$ & $43.2$ & $52.7$ & $\mathbf{51.1}$ & $\mathbf{66.3}$ & $39.4$ & $45.0$ & $51.5$ & $\mathbf{49.7}$ & $\mathbf{66.8}$ & $59.2$ & $44.8$ & $51.3$ & $\mathbf{49.7}$ & $\mathbf{67.1}$ \\
    \bottomrule
  \end{tabular}
\end{table}

\subsection{Staged Prediction}
\label{app:additional_expt/staging}
In the FR condition, models map facts and predict conclusions in a single turn.
We test whether separating these tasks helps models use the relevant facts to reach their conclusions.
Models first map facts to elements, then predict element conclusions and violation in a second turn (prompt in App.~\ref{app:prompts/ds_staging}).

Table~\ref{tab:staging} shows that staging can improve violation prediction substantially, particularly for models such as {\tt Qwen 3.5 122B} ($\textcolor{green!60!black}{\bm{\uparrow}} 59.9 \%$ \textit{m}F1) and {\tt Gemma 4 4B} ($\textcolor{green!60!black}{\bm{\uparrow}} 58.4 \%$ \textit{m}F1).
However, these gains do not extend consistently to the element conclusions or the supporting facts.
Element-conclusion \textit{m}F1 improves for only five models, with a maximum relative gain of $\textcolor{green!60!black}{\bm{\uparrow}} 7.2 \%$.
List~2 F1 actually falls for six models, and remains at or below $48.6\%$.
These results suggest that the difficulty of fact mapping is not simply due to performing it alongside element conclusion and violation prediction, since giving it a separate turn does not consistently improve performance.

\begin{table}[t]
  \caption{Full single-turn reasoning (FR) against staged two-turn reasoning (fact mapping first, then conclusions). Metrics: violation macro-F1 (V), element-conclusion macro-F1 averaged over the six elements (E), and F1 for List~1 (L1), List~2 (L2) and List~3 (L3).}
  \label{tab:staging}
  \centering
  \footnotesize
  \setlength{\tabcolsep}{2.5pt}
  \begin{tabular}{lrrrrr@{\hspace{8pt}}rrrrr}
    \toprule
     & \multicolumn{5}{c}{Full (FR)} & \multicolumn{5}{c}{Staged} \\
    \cmidrule(lr){2-6}\cmidrule(lr){7-11}
    Model & V & E & L1 & L2 & L3 & V & E & L1 & L2 & L3 \\
    \midrule
    \multicolumn{11}{l}{\textit{Closed weights}} \\
    Claude Sonnet 4.6 & $48.4$ & $50.0$ & $51.5$ & $44.4$ & $57.9$ & $48.5$ & $43.1$ & $48.5$ & $39.3$ & $49.8$ \\
    GPT-5.4 & $52.9$ & $41.8$ & $\mathbf{54.3}$ & $48.8$ & $63.0$ & $52.7$ & $43.4$ & $50.8$ & $43.9$ & $59.6$ \\
    Gemini 3 Flash & $\mathbf{63.5}$ & $\mathbf{51.8}$ & $51.2$ & $47.2$ & $57.6$ & $\mathbf{69.0}$ & $\mathbf{49.2}$ & $52.7$ & $46.0$ & $54.0$ \\
    \midrule
    \multicolumn{11}{l}{\textit{Open weights}} \\
    GPT OSS 20B & $31.3$ & $39.6$ & $27.2$ & $26.2$ & $46.9$ & $35.8$ & $40.2$ & $32.3$ & $30.2$ & $59.3$ \\
    GPT OSS 120B & $39.3$ & $43.0$ & $32.0$ & $30.3$ & $50.1$ & $40.3$ & $40.1$ & $35.8$ & $32.3$ & $53.7$ \\
    Qwen 3.5 9B & $43.9$ & $45.3$ & $44.7$ & $39.4$ & $50.2$ & $35.0$ & $44.8$ & $45.0$ & $38.9$ & $50.5$ \\
    Qwen 3.5 35B & $35.2$ & $41.6$ & $31.6$ & $29.5$ & $29.3$ & $35.2$ & $40.3$ & $48.9$ & $44.4$ & $\mathbf{59.9}$ \\
    Qwen 3.5 122B & $34.7$ & $41.8$ & $44.6$ & $40.5$ & $63.0$ & $55.5$ & $44.8$ & $49.8$ & $44.1$ & $59.4$ \\
    Gemma 4 4B & $32.9$ & $43.9$ & $39.7$ & $39.9$ & $62.8$ & $52.1$ & $44.5$ & $42.6$ & $39.7$ & $58.1$ \\
    Gemma 4 31B & $59.2$ & $44.8$ & $51.3$ & $\mathbf{49.7}$ & $\mathbf{67.1}$ & $55.2$ & $45.7$ & $\mathbf{52.9}$ & $\mathbf{48.6}$ & $59.8$ \\
    \bottomrule
  \end{tabular}
\end{table}

\subsection{Extended Thinking}
\label{app:additional_expt/thinking}

We test whether giving models more scope for internal reasoning helps them identify the facts supporting the Court's conclusions.
For six models, we compare the default configuration used in the main paper, with thinking disabled or set to the minimum supported level, against the maximum available thinking or reasoning mode.

Table~\ref{tab:thinking} shows that extended thinking improves violation prediction and element conclusions for all six models.
{\tt Gemini 3 Flash} reaches $75.7\%$ violation macro-F1 and $61.1\%$ element-conclusion macro-F1.
However, the benefits are less consistent for fact mapping.
List~2 F1 improves for four models but falls for {\tt Qwen 3.5 122B} and {\tt Gemma 4 31B}, and never exceeds $51.4\%$ for any model.
This suggests that limited thinking alone does not explain the difficulty of fact mapping: additional reasoning can help models reach the Court's conclusions better but do not help them identify the supporting facts.

\begin{table}[t]
  \caption{Effect of extended thinking or reasoning (FR condition). Off: default/minimal configuration; On: maximum available thinking or reasoning mode. Metrics: violation macro-F1 (V), avg. element-conclusion macro-F1 over the 6 elements (E), and F1 for List~1 (L1), List~2 (L2) and List~3 (L3).}
  \label{tab:thinking}
  \centering
  \footnotesize
  \setlength{\tabcolsep}{2.5pt}
  \begin{tabular}{lrrrrr@{\hspace{8pt}}rrrrr}
    \toprule
     & \multicolumn{5}{c}{Thinking off} & \multicolumn{5}{c}{Thinking on} \\
    \cmidrule(lr){2-6}\cmidrule(lr){7-11}
    Model & V & E & L1 & L2 & L3 & V & E & L1 & L2 & L3 \\
    \midrule
    \multicolumn{11}{l}{\textit{Closed weights}} \\
    Claude Sonnet 4.6 & $48.4$ & $50.0$ & $51.5$ & $44.4$ & $57.9$ & $53.5$ & $54.9$ & $\mathbf{55.9}$ & $50.5$ & $60.1$ \\
    GPT-5.4 & $52.9$ & $41.8$ & $\mathbf{54.3}$ & $48.8$ & $63.0$ & $68.9$ & $52.6$ & $55.4$ & $\mathbf{51.4}$ & $64.2$ \\
    Gemini 3 Flash & $\mathbf{63.5}$ & $\mathbf{51.8}$ & $51.2$ & $47.2$ & $57.6$ & $\mathbf{75.7}$ & $\mathbf{61.1}$ & $52.5$ & $47.9$ & $54.1$ \\
    \midrule
    \multicolumn{11}{l}{\textit{Open weights}} \\
    GPT OSS 120B & $39.3$ & $43.0$ & $32.0$ & $30.3$ & $50.1$ & $54.3$ & $46.1$ & $41.9$ & $39.5$ & $61.7$ \\
    Qwen 3.5 122B & $34.7$ & $41.8$ & $44.6$ & $40.5$ & $63.0$ & $35.8$ & $42.5$ & $43.3$ & $38.1$ & $58.0$ \\
    Gemma 4 31B & $59.2$ & $44.8$ & $51.3$ & $\mathbf{49.7}$ & $\mathbf{67.1}$ & $62.9$ & $48.0$ & $45.7$ & $42.9$ & $\mathbf{65.4}$ \\
    \bottomrule
  \end{tabular}
\end{table}

\subsection{Few-Shot Prompting}
\label{app:additional_expt/few-shot}
We test whether examples help models understand how to map facts to elements and reach the corresponding conclusions.
We compare zero-shot FR with a prompt containing five demonstrations, i.e., few-shot examples (prompt in App.~\ref{app:prompts/ds_fewshot}).

Table~\ref{tab:fewshot} shows that five-shot prompting improves violation prediction for all ten models, but element conclusions improve for only four.
The examples also help with fact mapping to certain extent: List~2 F1 improves for seven models, with particularly large gains for {\tt Qwen 3.5 35B} ($\textcolor{green!60!black}{\bm{\uparrow}} 56.3 \%$) and {\tt Qwen 3.5 122B} ($\textcolor{green!60!black}{\bm{\uparrow}} 17.5 \%$).
Nevertheless, five-shot List~2 F1 remains at or below $49.6\%$ for all models.
This suggests that the difficulty of fact mapping is not simply due to a lack of examples.
Even after seeing how supporting facts are chosen in other cases, models still struggle to identify them in a new case.

\begin{table}[t]
  \caption{Zero-shot vs. five-shot prompting (FR condition). Metrics: violation \textit{m}F1 (V), avg. element conclusion \textit{m}F1 over the 6 elements (E), and F1 for List~1 (L1), List~2 (L2) and List~3 (L3).}
  \label{tab:fewshot}
  \centering
  \footnotesize
  \setlength{\tabcolsep}{2.5pt}
  \begin{tabular}{lrrrrr@{\hspace{8pt}}rrrrr}
    \toprule
     & \multicolumn{5}{c}{Zero-shot} & \multicolumn{5}{c}{Five-shot} \\
    \cmidrule(lr){2-6}\cmidrule(lr){7-11}
    Model & V & E & L1 & L2 & L3 & V & E & L1 & L2 & L3 \\
    \midrule
    \multicolumn{11}{l}{\textit{Closed weights}} \\
    Claude Sonnet 4.6 & $48.4$ & $50.0$ & $51.5$ & $44.4$ & $57.9$ & $60.2$ & $44.6$ & $52.5$ & $48.2$ & $65.3$ \\
    GPT-5.4 & $52.9$ & $41.8$ & $\mathbf{54.3}$ & $48.8$ & $63.0$ & $62.3$ & $46.6$ & $\mathbf{53.0}$ & $\mathbf{49.6}$ & $65.4$ \\
    Gemini 3 Flash & $\mathbf{63.5}$ & $\mathbf{51.8}$ & $51.2$ & $47.2$ & $57.6$ & $\mathbf{74.2}$ & $\mathbf{48.2}$ & $50.0$ & $49.1$ & $\mathbf{68.8}$ \\
    \midrule
    \multicolumn{11}{l}{\textit{Open weights}} \\
    GPT OSS 20B & $31.3$ & $39.6$ & $27.2$ & $26.2$ & $46.9$ & $38.9$ & $38.3$ & $22.7$ & $22.6$ & $42.1$ \\
    GPT OSS 120B & $39.3$ & $43.0$ & $32.0$ & $30.3$ & $50.1$ & $48.5$ & $41.9$ & $37.9$ & $37.2$ & $60.2$ \\
    Qwen 3.5 9B & $43.9$ & $45.3$ & $44.7$ & $39.4$ & $50.2$ & $48.3$ & $44.6$ & $47.2$ & $42.6$ & $59.7$ \\
    Qwen 3.5 35B & $35.2$ & $41.6$ & $31.6$ & $29.5$ & $29.3$ & $48.6$ & $43.2$ & $48.1$ & $46.1$ & $63.4$ \\
    Qwen 3.5 122B & $34.7$ & $41.8$ & $44.6$ & $40.5$ & $63.0$ & $56.2$ & $45.3$ & $50.5$ & $47.6$ & $64.2$ \\
    Gemma 4 4B & $32.9$ & $43.9$ & $39.7$ & $39.9$ & $62.8$ & $47.0$ & $43.7$ & $38.1$ & $38.5$ & $61.5$ \\
    Gemma 4 31B & $59.2$ & $44.8$ & $51.3$ & $\mathbf{49.7}$ & $\mathbf{67.1}$ & $63.5$ & $46.8$ & $49.6$ & $49.2$ & $67.3$ \\
    \bottomrule
  \end{tabular}
\end{table}

\subsection{Evaluation on Post-Cutoff Judgments}
\label{app:additional_expt/memorization}

We wish to investigate if the models have simply memorized the judgments in our benchmark and reproducing the outputs from memory, which would artificially inflate the scores.
To this end, we collected nine additional Article~11 judgments, of which seven met our sample selection criteria.
These judgments were delivered between 9 July 2025 and 16 July 2026.
We evaluated five models whose disclosed training cutoffs preceded all seven judgments.

Table~\ref{tab:post_cutoff} compares these results with the main benchmark.
Violation macro-F1 is higher on the seven-case set for all five models.
Element-conclusion macro-F1 is also higher, but its ceiling depends on the classes present in each set.
Four elements have only one gold class in these seven cases.
Under our metric conventions (see App.~\ref{app:fullresults/metrics}), each of these elements has a maximum score of $50\%$.
The maximum average across the six elements is therefore $66.7\%$.
{\tt Gemini 3 Flash} reaches this ceiling, correctly predicting every active element conclusion.

Fact-mapping scores are lower on the seven-case set for every model and every list.
List~2 F1 ranges from $21.0\%$ to $34.1\%$, compared with $26.2\%$ to $49.7\%$ for the same models on the main benchmark.

We also evaluated {\tt Claude Sonnet 4.6} and {\tt GPT-5.4} on smaller subsets.
These contained only judgments delivered after their respective disclosed cutoffs of January 2026 and August 2025.
{\tt Claude} correctly predicted all three violation outcomes and all 17 active element conclusions.
{\tt GPT-5.4} correctly predicted all four violation outcomes and 20 of 22 active element conclusions.
It predicted every active element correctly in two of its four cases.

However, we wish to point out, the sample is too small to tell whether these patterns would hold across a larger set of post-cutoff judgments.
We therefore cannot use these results to confirm or rule out memorization as a factor in performance on the main benchmark. We specifically mention this point as a limitation of the current paper in Section~\ref{sec:discussion}.

\begin{table}[t]
  \caption{FR performance on the main benchmark and seven post-cutoff judgments.
  V: violation macro-F1; E: element-conclusion macro-F1 averaged over six elements; L1--L3: fact-mapping F1.
  All scores are percentages.
  On the seven-case set, four elements have only one gold class.
  The maximum E score is therefore $66.7\%$ under our fixed two-class scoring convention.}
  \label{tab:post_cutoff}
  \centering
  \footnotesize
  \setlength{\tabcolsep}{2.5pt}
  \begin{tabular}{lrrrrr@{\hspace{8pt}}rrrrr}
    \toprule
    & \multicolumn{5}{c}{Main benchmark (375 cases)}
    & \multicolumn{5}{c}{Post-cutoff (7 cases)} \\
    \cmidrule(lr){2-6}\cmidrule(lr){7-11}
    Model & V & E & L1 & L2 & L3
          & V & E & L1 & L2 & L3 \\
    \midrule
    Gemini 3 Flash
      & 63.5 & 51.8 & 51.2 & 47.2 & 57.6
      & 100.0 & 66.7 & 33.5 & 32.7 & 27.2 \\
    GPT OSS 20B
      & 31.3 & 39.6 & 27.2 & 26.2 & 46.9
      & 53.3 & 54.9 & 19.6 & 21.4 & 12.1 \\
    GPT OSS 120B
      & 39.3 & 43.0 & 32.0 & 30.3 & 50.1
      & 53.3 & 52.1 & 22.5 & 21.0 & 17.9 \\
    Gemma 4 4B
      & 32.9 & 43.9 & 39.7 & 39.9 & 62.8
      & 65.0 & 55.4 & 24.3 & 23.9 & 20.0 \\
    Gemma 4 31B
      & 59.2 & 44.8 & 51.3 & 49.7 & 67.1
      & 78.8 & 61.8 & 32.8 & 34.1 & 25.4 \\
    \bottomrule
  \end{tabular}
\end{table}

\clearpage
\newpage

\section{Generalizability of Benchmark Construction} \label{app:general}

\subsection{Benchmark Construction with Gemini} \label{app:general/gemini}

As mentioned in Section~\ref{sec:generalizability}, we construct another version of the benchmark with {\tt Gemini 3 Flash} by repeating stages~3-5 of the pipeline (Figure~\ref{fig:pipeline}), but keeping the atomic facts extracted by {\tt Claude Sonnet 4.6}. This enables us to create another set of reference annotations based on a shared set of atomic facts on which the downstream performance can also be compared.

\paragraph{Agreement between reference annotations.}
Table~\ref{tab:gold_agreement} reports overall agreement, with a per-element breakdown in Table~\ref{tab:gold_agreement_elements}. The two reference annotations agree on every violation outcome. Element conclusions also show high agreement overall, although this varies across elements: \textit{Scope} and \textit{Interference} have perfect agreement (AC1: $1.000$) on jointly active cases, while \textit{Positive obligation} has the lowest agreement (AC1: $0.462$).

For fact mapping, AC1 ranges from $0.921-0.969$ for Lists~1--3, whereas Jaccard similarity is $0.340$, $0.314$ and $0.442$ respectively.
These metrics capture different aspects of agreement: AC1 includes agreement on unselected facts, while Jaccard measures overlap among selected facts. This overlap also varies by element: List~2 Jaccard ranges from $0.150$ for \textit{Legitimate aim} to $0.528$ for \textit{Interference}.

\begin{table}[t]
  \caption{Agreement between the Claude-built and the Gemini-built gold standard (both on the same atomic facts), over all 375 cases: Cohen's $\kappa$~\citep{cohen1960coefficient}, Krippendorff's $\alpha$~\citep{krippendorff1970estimating}, Gwet's AC1~\citep{gwet2008computing} and Jaccard similarity. Violation and element conclusions are compared per case; Lists~1--3 are compared per atomic fact (selected or not). Element conclusion is averaged over the four elements for which $\kappa$ and $\alpha$ are defined (Scope and Interference are \textit{yes} in every case in both golds); Lists~1 and 2 are averaged over all six elements. Jaccard applies only to the fact lists and is computed on the selected facts.}
  \label{tab:gold_agreement}
  \centering
  \footnotesize
  \begin{tabular}{lrrrr}
    \toprule
    & $\kappa$ & $\alpha$ & AC1 & Jaccard \\
    \midrule
    Violation & $1.000$ & $1.000$ & $1.000$ & -- \\
    Element conclusion & $0.734$ & $0.737$ & $0.853$ & -- \\
    List 1 & $0.469$ & $0.469$ & $0.934$ & $0.340$ \\
    List 2 & $0.446$ & $0.446$ & $0.969$ & $0.314$ \\
    List 3 & $0.577$ & $0.577$ & $0.921$ & $0.442$ \\
    \bottomrule
  \end{tabular}
\end{table}

\begin{table}[t]
  \caption{Per-element agreement between the Claude-built and the Gemini-built gold standard. Element conclusion: cases for which both golds give a conclusion; Lists~1 and~2: all atomic facts of all cases. Dashes mark undefined values ($\kappa$ and $\alpha$ when both golds are constant, as for Scope and Interference conclusions). Jaccard is computed on the selected facts and pooled over cases.}
  \label{tab:gold_agreement_elements}
  \centering
  \scriptsize
  \setlength{\tabcolsep}{4.0pt}
  \begin{tabular}{lrrr@{}p{10pt}@{}rrrr@{}p{10pt}@{}rrrr}
    \toprule
     & \multicolumn{3}{c}{Element conclusion} & & \multicolumn{4}{c}{List 1} & & \multicolumn{4}{c}{List 2} \\
    \cmidrule(lr){2-4}\cmidrule(lr){6-9}\cmidrule(lr){11-14}
    Element & $\kappa$ & $\alpha$ & AC1 & & $\kappa$ & $\alpha$ & AC1 & Jaccard & & $\kappa$ & $\alpha$ & AC1 & Jaccard \\
    \midrule
    Scope & -- & -- & $1.000$ & & $0.550$ & $0.550$ & $0.903$ & $0.422$ & & $0.603$ & $0.603$ & $0.962$ & $0.451$ \\
    Interference & -- & -- & $1.000$ & & $0.638$ & $0.638$ & $0.937$ & $0.500$ & & $0.677$ & $0.677$ & $0.972$ & $0.528$ \\
    Prescribed by law & $0.920$ & $0.920$ & $0.970$ & & $0.470$ & $0.470$ & $0.943$ & $0.331$ & & $0.440$ & $0.440$ & $0.974$ & $0.292$ \\
    Legitimate aim & $0.691$ & $0.691$ & $0.981$ & & $0.311$ & $0.311$ & $0.969$ & $0.195$ & & $0.254$ & $0.254$ & $0.986$ & $0.150$ \\
    Necessity & $0.961$ & $0.961$ & $0.997$ & & $0.416$ & $0.416$ & $0.864$ & $0.314$ & & $0.437$ & $0.437$ & $0.930$ & $0.307$ \\
    Positive obligation & $0.362$ & $0.376$ & $0.462$ & & $0.428$ & $0.428$ & $0.987$ & $0.277$ & & $0.265$ & $0.265$ & $0.992$ & $0.155$ \\
    \bottomrule
  \end{tabular}
\end{table}

\paragraph{Effect on downstream evaluation.}
Table~\ref{tab:gold_swap_performance} reports FR performance on the {\tt Gemini}-built reference annotations and the difference (in percentage points) w.r.t. the {\tt Claude} reference annotations.
Violation scores do not change since the two references are identical for every case.
Element-conclusion scores increase for every model. Conversely, List~2 and List~3 F1 decrease for every model, by $3.2-9.3$ and $2.7-6.7$ percentage points, respectively. Absolute scores therefore depend on the reference annotations.
Nevertheless, as Figure~\ref{fig:benchmark-alignment} in the main paper shows, the scores remain strongly correlated across references (Pearson's $r$: $0.96-1.00$), and rankings remain largely stable (Spearman's $\rho$: $0.88-1.00$). {\tt Gemini 3 Flash} is the best performing model on both violation and element conclusions, and {\tt Gemma-4-31B} on Lists~2--3, which is the same as observed on {\tt Claude}-built references.
Neither construction model gains a consistent advantage across tasks, although {\tt Gemini} rises from fourth to first on List~1 against its own reference.
Thus, the comparative findings are largely consistent even after changing the benchmark construction model.

\begin{table}[t]
  \caption{Performance of the ten models on the Gemini-built gold standard (FR condition, 375 cases). Violation and element conclusions: macro-F1 (element averaged over the six elements); fact mapping: F1 for Lists~1--3 (Lists~1 and~2 averaged over elements). Each score (except violation) is followed by its change relative to the Claude-built gold standard (Gemini-built minus Claude-built, in percentage points): \rlap{$\textcolor{green!60!black}{\bm{\uparrow}}$}\phantom{$\bm{\downarrow}$} an increase, \rlap{$\textcolor{red!70!black}{\bm{\downarrow}}$}\phantom{$\bm{\downarrow}$} a decrease. The violation gold is identical in the two benchmarks, so violation scores do not change. \textbf{Bold} values show the best model per column on the Gemini-built gold.}
  \label{tab:gold_swap_performance}
  \centering
  \footnotesize
  \setlength{\tabcolsep}{5pt}
  \begin{tabular}{lrr@{\hspace{3pt}}c@{\hspace{1.5pt}}rr@{\hspace{3pt}}c@{\hspace{1.5pt}}rr@{\hspace{3pt}}c@{\hspace{1.5pt}}rr@{\hspace{3pt}}c@{\hspace{1.5pt}}r}
    \toprule
     & \multicolumn{4}{c}{Macro-F1} & \multicolumn{9}{c}{Fact mapping F1} \\
    \cmidrule(lr){2-5}\cmidrule(lr){6-14}
    Model & Viol. & \multicolumn{3}{c}{Elem.} & \multicolumn{3}{c}{L1} & \multicolumn{3}{c}{L2} & \multicolumn{3}{c}{L3} \\
    \midrule
    \multicolumn{14}{l}{\textit{Closed weights}} \\
    Claude Sonnet 4.6 & $48.4$ & $53.0$ & {\scriptsize\rlap{$\textcolor{green!60!black}{\bm{\uparrow}}$}\phantom{$\bm{\downarrow}$}} & {\scriptsize$3.0$} & $51.3$ & {\scriptsize\rlap{$\textcolor{red!70!black}{\bm{\downarrow}}$}\phantom{$\bm{\downarrow}$}} & {\scriptsize$0.2$} & $38.5$ & {\scriptsize\rlap{$\textcolor{red!70!black}{\bm{\downarrow}}$}\phantom{$\bm{\downarrow}$}} & {\scriptsize$5.9$} & $54.2$ & {\scriptsize\rlap{$\textcolor{red!70!black}{\bm{\downarrow}}$}\phantom{$\bm{\downarrow}$}} & {\scriptsize$3.7$} \\
    GPT-5.4 & $52.9$ & $45.4$ & {\scriptsize\rlap{$\textcolor{green!60!black}{\bm{\uparrow}}$}\phantom{$\bm{\downarrow}$}} & {\scriptsize$3.7$} & $51.1$ & {\scriptsize\rlap{$\textcolor{red!70!black}{\bm{\downarrow}}$}\phantom{$\bm{\downarrow}$}} & {\scriptsize$3.2$} & $39.5$ & {\scriptsize\rlap{$\textcolor{red!70!black}{\bm{\downarrow}}$}\phantom{$\bm{\downarrow}$}} & {\scriptsize$9.3$} & $56.3$ & {\scriptsize\rlap{$\textcolor{red!70!black}{\bm{\downarrow}}$}\phantom{$\bm{\downarrow}$}} & {\scriptsize$6.7$} \\
    Gemini 3 Flash & $\mathbf{63.5}$ & $\mathbf{55.2}$ & {\scriptsize\rlap{$\textcolor{green!60!black}{\bm{\uparrow}}$}\phantom{$\bm{\downarrow}$}} & {\scriptsize$3.4$} & $\mathbf{54.0}$ & {\scriptsize\rlap{$\textcolor{green!60!black}{\bm{\uparrow}}$}\phantom{$\bm{\downarrow}$}} & {\scriptsize$2.8$} & $42.4$ & {\scriptsize\rlap{$\textcolor{red!70!black}{\bm{\downarrow}}$}\phantom{$\bm{\downarrow}$}} & {\scriptsize$4.8$} & $54.0$ & {\scriptsize\rlap{$\textcolor{red!70!black}{\bm{\downarrow}}$}\phantom{$\bm{\downarrow}$}} & {\scriptsize$3.5$} \\
    \midrule
    \multicolumn{14}{l}{\textit{Open weights}} \\
    GPT OSS 20B & $31.3$ & $42.9$ & {\scriptsize\rlap{$\textcolor{green!60!black}{\bm{\uparrow}}$}\phantom{$\bm{\downarrow}$}} & {\scriptsize$3.3$} & $27.3$ & {\scriptsize\rlap{$\textcolor{green!60!black}{\bm{\uparrow}}$}\phantom{$\bm{\downarrow}$}} & {\scriptsize$0.2$} & $23.0$ & {\scriptsize\rlap{$\textcolor{red!70!black}{\bm{\downarrow}}$}\phantom{$\bm{\downarrow}$}} & {\scriptsize$3.2$} & $42.3$ & {\scriptsize\rlap{$\textcolor{red!70!black}{\bm{\downarrow}}$}\phantom{$\bm{\downarrow}$}} & {\scriptsize$4.6$} \\
    GPT OSS 120B & $39.3$ & $43.8$ & {\scriptsize\rlap{$\textcolor{green!60!black}{\bm{\uparrow}}$}\phantom{$\bm{\downarrow}$}} & {\scriptsize$0.8$} & $32.9$ & {\scriptsize\rlap{$\textcolor{green!60!black}{\bm{\uparrow}}$}\phantom{$\bm{\downarrow}$}} & {\scriptsize$0.9$} & $26.8$ & {\scriptsize\rlap{$\textcolor{red!70!black}{\bm{\downarrow}}$}\phantom{$\bm{\downarrow}$}} & {\scriptsize$3.5$} & $45.4$ & {\scriptsize\rlap{$\textcolor{red!70!black}{\bm{\downarrow}}$}\phantom{$\bm{\downarrow}$}} & {\scriptsize$4.7$} \\
    Qwen 3.5 9B & $43.9$ & $49.4$ & {\scriptsize\rlap{$\textcolor{green!60!black}{\bm{\uparrow}}$}\phantom{$\bm{\downarrow}$}} & {\scriptsize$4.1$} & $43.3$ & {\scriptsize\rlap{$\textcolor{red!70!black}{\bm{\downarrow}}$}\phantom{$\bm{\downarrow}$}} & {\scriptsize$1.4$} & $33.4$ & {\scriptsize\rlap{$\textcolor{red!70!black}{\bm{\downarrow}}$}\phantom{$\bm{\downarrow}$}} & {\scriptsize$6.0$} & $47.5$ & {\scriptsize\rlap{$\textcolor{red!70!black}{\bm{\downarrow}}$}\phantom{$\bm{\downarrow}$}} & {\scriptsize$2.7$} \\
    Qwen 3.5 35B & $35.2$ & $42.6$ & {\scriptsize\rlap{$\textcolor{green!60!black}{\bm{\uparrow}}$}\phantom{$\bm{\downarrow}$}} & {\scriptsize$1.0$} & $32.8$ & {\scriptsize\rlap{$\textcolor{green!60!black}{\bm{\uparrow}}$}\phantom{$\bm{\downarrow}$}} & {\scriptsize$1.2$} & $25.5$ & {\scriptsize\rlap{$\textcolor{red!70!black}{\bm{\downarrow}}$}\phantom{$\bm{\downarrow}$}} & {\scriptsize$4.0$} & $25.3$ & {\scriptsize\rlap{$\textcolor{red!70!black}{\bm{\downarrow}}$}\phantom{$\bm{\downarrow}$}} & {\scriptsize$4.0$} \\
    Qwen 3.5 122B & $34.7$ & $45.6$ & {\scriptsize\rlap{$\textcolor{green!60!black}{\bm{\uparrow}}$}\phantom{$\bm{\downarrow}$}} & {\scriptsize$3.8$} & $42.5$ & {\scriptsize\rlap{$\textcolor{red!70!black}{\bm{\downarrow}}$}\phantom{$\bm{\downarrow}$}} & {\scriptsize$2.1$} & $32.4$ & {\scriptsize\rlap{$\textcolor{red!70!black}{\bm{\downarrow}}$}\phantom{$\bm{\downarrow}$}} & {\scriptsize$8.1$} & $56.5$ & {\scriptsize\rlap{$\textcolor{red!70!black}{\bm{\downarrow}}$}\phantom{$\bm{\downarrow}$}} & {\scriptsize$6.5$} \\
    Gemma 4 4B & $32.9$ & $46.7$ & {\scriptsize\rlap{$\textcolor{green!60!black}{\bm{\uparrow}}$}\phantom{$\bm{\downarrow}$}} & {\scriptsize$2.8$} & $42.0$ & {\scriptsize\rlap{$\textcolor{green!60!black}{\bm{\uparrow}}$}\phantom{$\bm{\downarrow}$}} & {\scriptsize$2.3$} & $36.1$ & {\scriptsize\rlap{$\textcolor{red!70!black}{\bm{\downarrow}}$}\phantom{$\bm{\downarrow}$}} & {\scriptsize$3.7$} & $57.7$ & {\scriptsize\rlap{$\textcolor{red!70!black}{\bm{\downarrow}}$}\phantom{$\bm{\downarrow}$}} & {\scriptsize$5.1$} \\
    Gemma 4 31B & $59.2$ & $48.9$ & {\scriptsize\rlap{$\textcolor{green!60!black}{\bm{\uparrow}}$}\phantom{$\bm{\downarrow}$}} & {\scriptsize$4.1$} & $53.5$ & {\scriptsize\rlap{$\textcolor{green!60!black}{\bm{\uparrow}}$}\phantom{$\bm{\downarrow}$}} & {\scriptsize$2.2$} & $\mathbf{45.0}$ & {\scriptsize\rlap{$\textcolor{red!70!black}{\bm{\downarrow}}$}\phantom{$\bm{\downarrow}$}} & {\scriptsize$4.8$} & $\mathbf{60.9}$ & {\scriptsize\rlap{$\textcolor{red!70!black}{\bm{\downarrow}}$}\phantom{$\bm{\downarrow}$}} & {\scriptsize$6.2$} \\
    \bottomrule
  \end{tabular}
\end{table}

\subsection{Extension to Article 10}
\label{app:general/art10}

To examine whether the pipeline transfers to another legal provision, we applied it to 26 recent Article~10 ECHR judgments concerning \textit{freedom of expression}, of which 25 passed the sample selection criteria.
We constructed the benchmark using {\tt Claude Sonnet 4.6}, following the procedure in Section~\ref{sec:benchmark}.

\begin{table}[t]
  \caption{Doctrinal elements for Articles~11 (\textit{freedom of assembly}) and ~10 (\textit{freedom of expression}). The six-element structure is the same, but \textit{Scope}, \textit{Interference} and \textit{Legitimate aim} are defined differently.}
  \label{tab:article_comparison}
  \centering
  \footnotesize
  \setlength{\tabcolsep}{7pt}
  \renewcommand{\arraystretch}{1.12}
  \begin{tabular}{
  @{}p{0.18\linewidth}
  p{0.37\linewidth}
  p{0.37\linewidth}@{}
 }
    \toprule
    Element & Article 11: assembly & Article 10: expression \\
    \midrule
    Scope
    & Whether an assembly occurred, the applicant participated or
      attempted to participate, and the assembly was peaceful.
    & Whether the case concerns expression, access to information
      or media activity, and the applicant was involved. \\
    \addlinespace
    Interference
    & Whether a governmental act or omission interfered with
      freedom of assembly.
    & Whether a governmental act or omission interfered with
      freedom of expression, and whether the intervention was
      a formality, condition, restriction or penalty. \\
    \addlinespace
    Prescribed by law
    & Foreseeability, accessibility and safeguards against abuse.
    & Same questions. \\
    \addlinespace
    Legitimate aim
    & National security, public safety, prevention of disorder
      or crime, protection of health or morals, and protection
      of others' rights and freedoms.
    & Shared aims, with additional categories for territorial
      integrity, others' reputation, preventing disclosure of
      confidential information, and maintaining judicial
      authority and impartiality. \\
    \addlinespace
    Necessity
    & Whether the government adequately balanced the intensity
      of the intervention against the weight of the legitimate aim.
    & Same question. \\
    \addlinespace
    Positive obligation
    & Whether the Court identified a positive obligation and
      found that the government violated it.
    & Same questions. \\
    \bottomrule
  \end{tabular}
\end{table}

\paragraph{Adapting the pipeline.}
Articles~10 and~11 share the same broad sequence of legal tests, but differ in the activities they protect and the grounds on which interference may be justified.
Table~\ref{tab:article_comparison} summarizes these differences.
For Article~10, \textit{Scope} covers expression, access to information and media activity, while \textit{Interference} additionally asks whether the intervention constitutes a formality, condition, restriction or penalty.
The \textit{Legitimate Aim} categories also expand to include territorial integrity, reputation, confidential information, and judicial authority and impartiality.
The question the Court asks for \textit{Prescribed by law}, \textit{Necessity} and \textit{Positive obligation} are the same for both articles.
We thus needed to modify only the article-specific wordings and element definitions in the prompts (see all the changes in App.~\ref{app:prompts/art_10}).

\begin{table}[t]
  \caption{Statistics for 25 Article 10 cases. Conclusion statistics use active cases \textit{only}; each element's avg. List~1 and 2 sizes exclude cases where the corresponding list is empty. Avg. List~1 and 2 sizes sum up list sizes across active elements, including empty ones, then average across cases.}
  \label{tab:benchmark_stats_art10}
  \centering
  \footnotesize
  \begin{tabular}{lrrrrr}
    \toprule
     & & \multicolumn{2}{c}{Conclusions} & \multicolumn{2}{c}{Avg facts} \\
    \cmidrule(lr){3-4}\cmidrule(lr){5-6}
    Element & Active & Yes & No & List 1 & List 2 \\
    \midrule
    Scope & $25$ & $25$ ($100\%$) & $0$ ($0\%$) & $13.68$ & $5.20$ \\
    Interference & $25$ & $25$ ($100\%$) & $0$ ($0\%$) & $9.36$ & $4.24$ \\
    Prescribed by law & $25$ & $21$ ($84\%$) & $4$ ($16\%$) & $8.48$ & $4.84$ \\
    Legitimate aim & $23$ & $21$ ($91\%$) & $2$ ($9\%$) & $8.04$ & $3.91$ \\
    Necessity & $22$ & $6$ ($27\%$) & $16$ ($73\%$) & $36.38$ & $18.43$ \\
    Positive obligation & $6$ & $2$ ($33\%$) & $4$ ($67\%$) & $14.50$ & $9.50$ \\
    \midrule
    \multicolumn{6}{l}{\textit{Case-level summary}} \\
    Violated & \multicolumn{5}{l}{$19$ / $25$ ($76.0\%$)} \\
    Not violated & \multicolumn{5}{l}{$6$ / $25$ ($24.0\%$)} \\
    Active elements/case & \multicolumn{5}{l}{Mean $5.04$ (min $4$, max $6$); $72.0\%$ have $5$ active elements} \\
    Atomic facts/case & \multicolumn{5}{l}{Mean $152.52$ (min $63$, max $564$)} \\
    Avg List 1/case & \multicolumn{5}{l}{$70.64$ relevant facts (summed across elements)} \\
    Avg List 2/case & \multicolumn{5}{l}{$34.12$ essential facts (summed across elements)} \\
    Avg List 3/case & \multicolumn{5}{l}{$14.40$ decisive facts} \\
    \bottomrule
  \end{tabular}
\end{table}

\paragraph{Benchmark statistics and evaluation.}
Table~\ref{tab:benchmark_stats_art10} shows the statistics for this Article~10 benchmark.
The Art.~10 sample contains many more atomic facts per case than Art.~11 (152.52 versus 58.75), and a smaller proportion of violation outcomes (76.0\% versus 96.3\%).
\textit{Scope} and \textit{Interference} have affirmative conclusions in every case, but \textit{Necessity} has a less skewed distribution: it concludes affirmatively in 6 of 22 active cases (27\%), compared with 4\% for Art~11.

We also constructed reference annotations using {\tt Gemini 3 Flash}, as done in the study in Section~\ref{sec:generalizability}, App.~\ref{app:general/gemini}.
We again observe high agreement with the {\tt Claude}-built benchmark: Gwet's AC1 is $1.000$ for violation, $0.954$ for element conclusions, and $0.926$, $0.966$ and $0.914$ for Lists~1--3, respectively.
Downstream evaluation of the same ten models in the FR condition against the {\tt Claude}-built reference is reported in Table~\ref{tab:art10} of the main paper.
Models still perform poorly on fact-mapping: List~2 F1 ranges from $19.6-36.0\%$, below the corresponding Article~11 score for every model.
Thus, the models' difficulties in mapping the supporting facts is not limited to Art. 11 only.
This is meant to be a preliminary transfer study, demonstrating the pipeline's applicability to a second provision.

\clearpage
\newpage
\section*{NeurIPS Paper Checklist}

\begin{enumerate}

\item {\bf Claims}
    \item[] Question: Do the main claims made in the abstract and introduction accurately reflect the paper's contributions and scope?
    \item[] Answer: \answerYes{} %
    \item[] Justification: The abstract and introduction (Section~\ref{sec:introduction}) clearly state the contributions and scope of the paper, and these are supported by the discussion on benchmark creation (Section~\ref{sec:benchmark}) and experimental results (Section~\ref{sec:experiments}).

\item {\bf Limitations}
    \item[] Question: Does the paper discuss the limitations of the work performed by the authors?
    \item[] Answer: \answerYes{} %
    \item[] Justification: The limitations of our methodology are explicitly discussed in Section~\ref{sec:discussion} under the paragraph heading \textbf{`Limitations'}.

\item {\bf Theory assumptions and proofs}
    \item[] Question: For each theoretical result, does the paper provide the full set of assumptions and a complete (and correct) proof?
    \item[] Answer: \answerNA{} %
    \item[] Justification: The paper does not contain formal theorems or lemmas requiring proofs.

    \item {\bf Experimental result reproducibility}
    \item[] Question: Does the paper fully disclose all the information needed to reproduce the main experimental results of the paper to the extent that it affects the main claims and/or conclusions of the paper (regardless of whether the code and data are provided or not)?
    \item[] Answer: \answerYes{} %
    \item[] Justification: The entire process of the benchmark creation is described thoroughly in Section~\ref{sec:benchmark}, including all the detailed steps such as sample selection~\ref{sec:benchmark/sample}, the full architecture of the pipeline~\ref{sec:benchmark/architecture} and additional details~\ref{app:benchmark}. For the downstream evaluation experiments, they are thoroughly described in Section~\ref{sec:experiments}, Appendix~\ref{app:fullresults} and Appendix~\ref{app:additional_expt}. The generalizability experiments are described in Section~\ref{sec:experiments} and Appendix~\ref{app:general}. All the prompts are provided in Appendix~\ref{app:prompts}.

\item {\bf Open access to data and code}
    \item[] Question: Does the paper provide open access to the data and code, with sufficient instructions to faithfully reproduce the main experimental results, as described in supplemental material?
    \item[] Answer: \answerYes{} %
    \item[] Justification: We release our dataset in huggingface (\url{https://huggingface.co/datasets/archon159/LegalReasoningBench}), as well as code in github (\url{https://github.com/archon159/LegalReasoningBench}).

\item {\bf Experimental setting/details}
    \item[] Question: Does the paper specify all the training and test details (e.g., data splits, hyperparameters, how they were chosen, type of optimizer) necessary to understand the results?
    \item[] Answer: \answerYes{} %
    \item[] Justification: The implementation details for the benchmark creation are provided in Appendix~\ref{app:benchmark/implementation} and for the downstream evaluation experiments in Appendix~\ref{app:implementation}.

\item {\bf Experiment statistical significance}
    \item[] Question: Does the paper report error bars suitably and correctly defined or other appropriate information about the statistical significance of the experiments?
    \item[] Answer: \answerNo{} %
    \item[] Justification: All downstream evaluation experiments are conducted in zero-shot setting with one single run and temperature set to 0 (Appendix~\ref{app:fullresults}), except the thinking mode enabled setting (Appendix~\ref{app:additional_expt/thinking}). The difference between results of experimental settings is large enough (Table~\ref{tab:main_results}) to draw conclusions without multiple runs or statistical significance testing, since all results are computed over 375 samples. For the human validation results, which are conducted over 20 samples only, we report $95\%$ bootstrapped CIs (Table~\ref{tab:validation}, \ref{tab:validation_overall}). The Art. 10 benchmark construction and downstream evaluation (Section~\ref{sec:generalizability}, Appendix~\ref{app:general/art10}) are done on 25 samples but this is meant to be a preliminary transfer study, and not the main findings of our paper.

\item {\bf Experiments compute resources}
    \item[] Question: For each experiment, does the paper provide sufficient information on the computer resources (type of compute workers, memory, time of execution) needed to reproduce the experiments?
    \item[] Answer: \answerYes{} %
    \item[] Justification: We report the time taken by the pipeline API calls and the API usage cost for benchmark creation, as well as model settings in Appendix~\ref{app:benchmark/implementation}. For downstream evaluation experiments, the model settings and GPU configurations are reported in Appendix~\ref{app:implementation}.

\item {\bf Code of ethics}
    \item[] Question: Does the research conducted in the paper conform, in every respect, with the NeurIPS Code of Ethics \url{https://neurips.cc/public/EthicsGuidelines}?
    \item[] Answer: \answerYes{} %
    \item[] Justification: The research conducted in this paper conforms with the NeurIPS Code of Ethics in every respect.

\item {\bf Broader impacts}
    \item[] Question: Does the paper discuss both potential positive societal impacts and negative societal impacts of the work performed?
    \item[] Answer: \answerYes{} %
    \item[] Justification: The broader impacts of our methodology are explicitly discussed in Section~\ref{sec:discussion} under the paragraph heading \textbf{`Future work and Broader Impacts.'}.

\item {\bf Safeguards}
    \item[] Question: Does the paper describe safeguards that have been put in place for responsible release of data or models that have a high risk for misuse (e.g., pre-trained language models, image generators, or scraped datasets)?
    \item[] Answer: \answerNA{} %
    \item[] Justification: The paper does not release any pre-trained models or image generators. The released dataset is built on publicly available source data and therefore poses no significant risk of misuse.

\item {\bf Licenses for existing assets}
    \item[] Question: Are the creators or original owners of assets (e.g., code, data, models), used in the paper, properly credited and are the license and terms of use explicitly mentioned and properly respected?
    \item[] Answer: \answerYes{} %
    \item[] Justification: All existing assets (datasets, models, and code packages) used in this paper are properly cited, and their licenses and terms of use are respected.

\item {\bf New assets}
    \item[] Question: Are new assets introduced in the paper well documented and is the documentation provided alongside the assets?
    \item[] Answer: \answerYes{} %
    \item[] Justification: All procedures and details of the benchmark creation process are documented in the README files of the released assets (GitHub repository and Hugging Face dataset).

\item {\bf Crowdsourcing and research with human subjects}
    \item[] Question: For crowdsourcing experiments and research with human subjects, does the paper include the full text of instructions given to participants and screenshots, if applicable, as well as details about compensation (if any)?
    \item[] Answer: \answerYes{} %
    \item[] Justification: A sample of 20 cases from the benchmark was validated by three human experts. All details of this exercise, including the payment details, are reported in Appendix~\ref{app:benchmark/validation}. The full guide document provided to the evaluators is provided at \url{https://github.com/archon159/LegalReasoningBench/blob/main/human_validation/Validation_Guide_Upwork.pdf}.

\item {\bf Institutional review board (IRB) approvals or equivalent for research with human subjects}
    \item[] Question: Does the paper describe potential risks incurred by study participants, whether such risks were disclosed to the subjects, and whether Institutional Review Board (IRB) approvals (or an equivalent approval/review based on the requirements of your country or institution) were obtained?
    \item[] Answer: \answerNA{} %
    \item[] Justification: This paper does not include any exercise where the participants can be subjected to risks. The three human experts are legal professionals, who were asked to validate the output of our benchmark creation pipeline against the original judgment documents, which are publicly available (details in Section~\ref{sec:benchmark} and Appendix~\ref{app:benchmark/validation}).

\item {\bf Declaration of LLM usage}
    \item[] Question: Does the paper describe the usage of LLMs if it is an important, original, or non-standard component of the core methods in this research? Note that if the LLM is used only for writing, editing, or formatting purposes and does \emph{not} impact the core methodology, scientific rigor, or originality of the research, declaration is not required.
    \item[] Answer: \answerYes{} %
    \item[] Justification: Yes. LLMs serve two distinct roles in this work. First, LLMs are central to benchmark construction: we use a multi-stage LLM pipeline to extract atomic facts, identify factual groundings, and produce element-level fact mappings from ECtHR judgments. The pipeline architecture, model choices, and prompts are documented in Section~\ref{sec:benchmark}. Second, LLMs are the subjects of evaluation: we benchmark ten LLMs with results reported in Section~\ref{sec:experiments}.

\end{enumerate}

\end{document}